\documentclass[10pt,twocolumn,letterpaper]{article}

 \usepackage{cvpr}              %

\usepackage{pifont}%
\usepackage{multicol}
\usepackage{color, colortbl}

\usepackage{siunitx}
\usepackage{makecell}
\usepackage{multirow}
\usepackage{stfloats}
\usepackage{float}
\usepackage{rotating}
\usepackage{pdflscape}
\usepackage{afterpage}

\newcommand{\cmark}{\textcolor{ForestGreen}{\ding{51}}}%
\newcommand{\xmark}{\textcolor{BrickRed}{\ding{55}}}%
\definecolor{LightGrey}{rgb}{0.9,0.9,0.9}
\definecolor{figred}{rgb}{1.0,0.4,0.4}
\definecolor{figblue}{rgb}{0.2,0.6,1.0}
\definecolor{figpurple}{rgb}{0.7,0.3,1.0}

\definecolor{vqagreen}{rgb}{0.6, 0.8, 0.4}
\definecolor{vqagemini}{rgb}{0.50, 0.46, 0.78}
\definecolor{vqavlama}{rgb}{0.71, 0.74, 0.54}
\definecolor{vqalngva}{rgb}{0.62, 0.06, 0.06}
\definecolor{vqallama}{rgb}{1.0, 0.4, 1.0}

\usepackage[dvipsnames]{xcolor}

\definecolor{recipe}{HTML}{ffadad}
\definecolor{nutrition}{HTML}{fdffb6}
\definecolor{ingredient}{HTML}{ffd6a5}
\definecolor{threed}{HTML}{a0c4ff}
\definecolor{fine}{HTML}{caffbf}
\definecolor{gaze}{HTML}{ffc6ff}
\definecolor{object}{HTML}{bdb2ff}
\definecolor{figbackgroundblue}{HTML}{E0FCF4}
\definecolor{figbackgroundyellow}{HTML}{FFF8E9}
\definecolor{figbackgroundpink}{HTML}{FEECF0}
\definecolor{verb}{HTML}{dbb3ff}
\definecolor{noun}{HTML}{75d7b9
}
\definecolor{hand}{HTML}{9acefe}
\definecolor{why}{HTML}{faad9e}
\definecolor{how}{HTML}{f9a3cd}

\newcommand{\recipe}[1]{\colorbox{recipe}{#1}}
\newcommand{\nutrition}[1]{\colorbox{nutrition}{#1}}
\newcommand{\ingredient}[1]{\colorbox{ingredient}{#1}}
\newcommand{\fine}[1]{\colorbox{fine}{#1}}
\newcommand{\threed}[1]{\colorbox{threed}{#1}}
\newcommand{\gaze}[1]{\colorbox{gaze}{#1}}
\newcommand{\object}[1]{\colorbox{object}{#1}}

\newcommand*{\DName}{HD-EPIC\xspace}

\addtocontents{toc}{\setcounter{tocdepth}{-10}}

\definecolor{cvprblue}{rgb}{0.21,0.49,0.74}
\usepackage[pagebackref,breaklinks,colorlinks,citecolor=cvprblue]{hyperref}
\usepackage{fdsymbol}

\def\paperID{0058} %
\def\confName{CVPR}
\def\confYear{2025}

\title{%
HD-EPIC: A Highly-Detailed Egocentric Video Dataset}

\author{
Toby Perrett$^{*\clubsuit}$$\quad$Ahmad Darkhalil$^{*\clubsuit}$$\quad$Saptarshi Sinha$^{*\clubsuit}$$\quad$Omar Emara$^{*\clubsuit}$$\quad$Sam Pollard$^{*\clubsuit}$\\ Kranti Kumar Parida$^{*\clubsuit}$$\quad$Kaiting Liu,$^{*\spadesuit}$$\quad$Prajwal Gatti$^{*\clubsuit}$$\quad$Siddhant Bansal$^{*\clubsuit}$$\quad$Kevin Flanagan$^{*\clubsuit}$\\Jacob Chalk$^{*\clubsuit}$$\quad$Zhifan Zhu$^{*\clubsuit}$$\quad$Rhodri Guerrier$^{*\clubsuit}$$\quad$Fahd Abdelazim$^{*\clubsuit}$$\quad$Bin Zhu{$^\vardiamondsuit$}\\Davide Moltisanti{$^\varheartsuit$} $\quad$ Michael Wray{$^\clubsuit$} $\quad$ Hazel Doughty{$^\spadesuit$} $\quad$ Dima Damen{$^\clubsuit$}\\
\small{
\noindent $^{\clubsuit}$Uni. of Bristol$\quad^{\spadesuit}$ Leiden Uni.$\quad^{\vardiamondsuit}$Singapore Management Uni. $\quad^{\varheartsuit}$Uni. of Bath $\quad$ $^*$: Equal Contribution}\\
\small{\url{http://hd-epic.github.io}}
}

\begin{document}
\maketitle

\begin{figure*}[b]
\setlength{\fboxsep}{0pt}
    \centering
    \includegraphics[width=\linewidth]{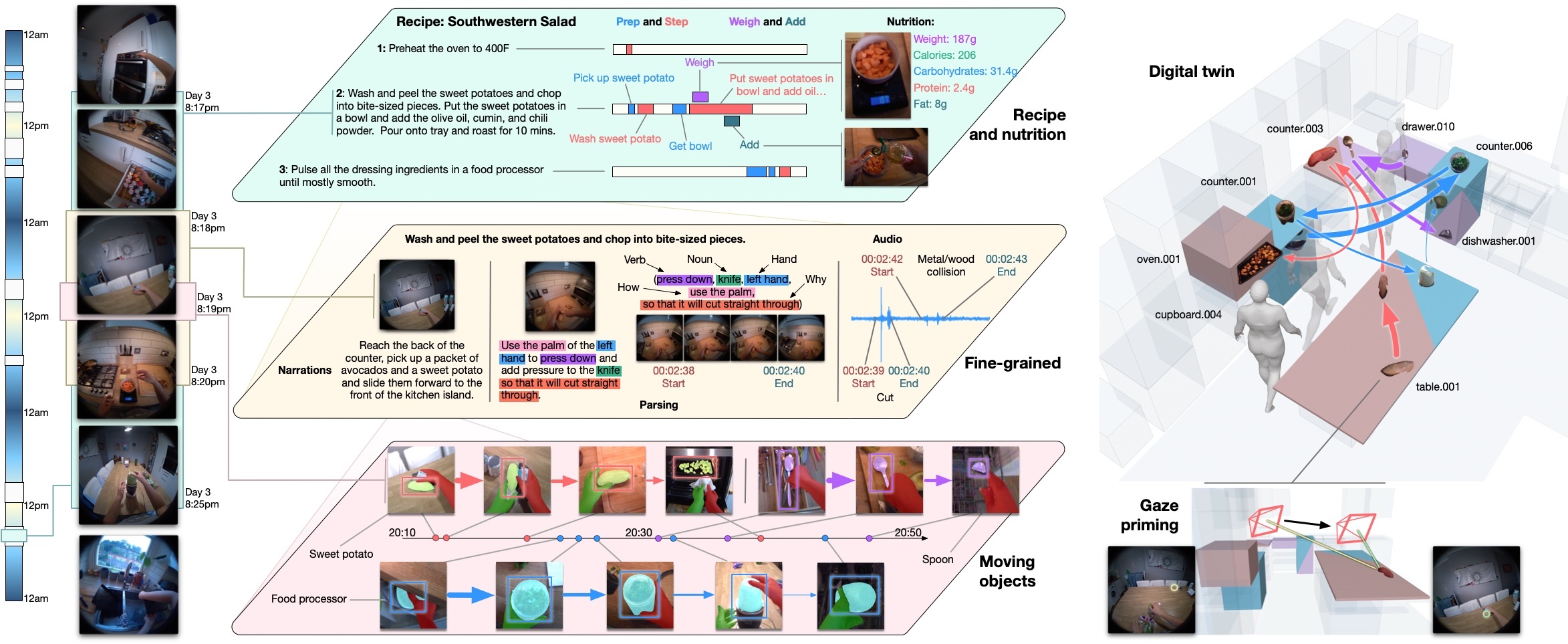}
    \vspace{-21pt}
\captionsetup{type=figure}
\captionof{figure}{\textbf{Annotation Highlights.} We capture multi-day recordings of unscripted activities. \colorbox{figbackgroundblue}{\textbf{Centre-Top}}: Recipes are recorded with steps and their preparation temporally annotated, along with ingredient addition. Ingredients are weighed and nutrition recorded. \colorbox{figbackgroundyellow}{\textbf{Centre-Middle}}: Dense fine-grained narrations detailing what, how, and why are parsed and clustered. Audio events are also annotated. \colorbox{figbackgroundpink}{\textbf{Centre-Bottom}}: Object movements are temporally annotated with bounding boxes and hands and object masks. \textbf{Right-Top}: All annotations are temporally grounded in a 3D digital twin. We show trajectories of 3 (masked) objects: \textcolor{figred}{Sweet potato}, \textcolor{figblue}{Food processor} and \textcolor{figpurple}{Spoon}, highlighting relevant kitchen fixtures. \textbf{Right-Bottom}: Gaze captures when objects are primed (\ie looked at) before being taken/placed.}
    \label{fig:f1}
    \vspace*{-12pt}
\end{figure*}

\begin{abstract}
We present a validation dataset of newly-collected kitchen-based egocentric videos, manually annotated with highly detailed and interconnected ground-truth labels covering: %
recipe steps, fine-grained actions, ingredients with nutritional values, moving objects, and audio annotations.
Importantly, all annotations are grounded in 3D through digital twinning of the scene, fixtures, object locations, and primed with gaze.
Footage is collected from unscripted recordings in diverse home environments, 
making \DName the first dataset collected in-the-wild but with detailed annotations matching those in controlled lab environments.

We show the potential of our highly-detailed annotations through a challenging VQA benchmark of 26K questions assessing the capability to recognise recipes, ingredients, nutrition, fine-grained actions, 3D perception, object motion, and gaze direction.
The powerful long-context Gemini Pro only achieves 37.6\% on this benchmark, showcasing its difficulty and highlighting shortcomings in current VLMs.
We additionally assess action recognition, sound recognition, and long-term video-object segmentation on \DName.

\DName is 41 hours of video in 9 kitchens with digital twins of 413 kitchen fixtures, capturing 69 recipes, 59K fine-grained actions, 51K audio events, 20K object movements and 37K object masks lifted to 3D.
On average, we have 263 annotations per minute of our unscripted videos. %
\end{abstract}

\vspace{-16pt}
\section{Introduction}
\vspace{-0.4em}
\label{sec:Intro}

Detailed understanding of videos, from the brief fine-grained action to the overarching hour-long activity, is effortless for humans but currently out of reach for both foundational and specialised models.
Egocentric videos, in particular, 
introduce additional challenges to general video understanding, including significant camera motion, subtle action motion, objects occluded during manipulations and frequently going out of view.
Understanding such videos requires disentangling the combined signals of head motion, hand interactions and a global understanding of the dynamic scene.
This makes ego videos a great testbed for a comprehensive evaluation of video perception models.

Egocentric vision has recently been fuelled by an influx of datasets~\cite{Damen2022RESCALING,Ego4D2022CVPR,EgoExo4d,HoloAssist2023}.
While large-scale, making them ideal for training, these datasets 
are sparsely annotated, particularly for tasks which link various parts of the long video, or those requiring 3D grounding.
In contrast, richly annotated datasets tend to be synthetic or collected in controlled settings~\cite{pan2023aria,sener2022assembly101,banerjee2024hot3d} which limits their realism. %
We bridge this gap by 
presenting the \textbf{most densely annotated dataset of unscripted recordings}, ideal for comprehensive \textbf{validation} of video-only and video-language models.

We collect new videos, allowing us to capture additional meta-data and to ensure these videos have not already been used to train existing models.
Following EPIC-Kitchens~\cite{damen2018scaling}, participants collect all kitchen activities for three  days.
We thus term our dataset Highly-Detailed EPIC (\DName). 
Fig.~\ref{fig:f1} provides an overview of the multi-tiered annotations, several of which are novel:
\begin{itemize}[label=$\star$]
    \item Recipe steps are temporally annotated, and linked to annotations of all preparatory actions that relate to the step.
    \item Ingredients are weighed in videos %
    and labelled with nutrition. We track dish nutrition as ingredients are added.
    \item Each action has a dense description capturing the what, how, and why of actions along the start and end time.
    \item For each kitchen, we curate a digital twin with labelled fixtures. These are associated with actions (\eg open/close) and the taking/placing of objects.
    \item All moved objects are tracked, with manual masks lifted to 3D bounding boxes.
    \item We associate gaze with object movements, labelling when objects are spotted before take/place actions. %
\end{itemize}

With these dense annotations, we design a challenging Visual Question Answering (VQA) benchmark of 26K questions. 
We purposefully do not use LLMs to generate negatives, instead using similar annotations. We highlight a few novel question types:
\begin{itemize}[label=$\star$]
    \item Recipe nutrition: we question the change in the recipe nutrition as one or more ingredients are added.
    \item Multi-video: we question recipes prepared across recordings, with a VQA that spans multiple long videos.
    \item Object itinerary: we question multi-hop object movements over a long video, relative to kitchen fixtures.
    \item Fixture interactions: we question how many times a particular cupboard/drawer is opened/closed.
    \item Action how/why: we question how/why an action was carried out, using participant-narrated  manners/reasons. %
    \item Anticipation with gaze:  With gaze priming, we query next-object movement, offering evidenced anticipation.
\end{itemize}
Additionally, we report results on action recognition, sound recognition, and long-term video object segmentation.

This paper thus contributes: (i) 41 hours of multi-day unscripted egocentric recordings, (ii) highly-detailed annotations including novel labels (\eg ingredient nutrition, digital twin, gaze prime) and (iii) a challenging VQA benchmark including novel Qs (\eg object itinerary, recipe nutrition changes) along with 3 standard video benchmarks.

\vspace{-0.5em}
\section{Related Work}
\vspace{-0.3em}

With the rise of foundation models \cite{radford2021learning,singh2022flava,alayrac2022flamingo,wang2023internimage,zhang2023video,achiam2023gpt,chen2024internvl,xiao2024florence,dubey2024llama,cheng2024videollama}, there has been a recent influx of %
benchmarks \cite{zhao2024videoniah, kesen2023vilma,mangalam2023egoschema,patraucean2023perception,li2024vitatecs,li2024seed,li2024mvbench,wang2024sok,chen2024rextime,fang2024mmbench,cai2024temporalbench,cores2024tvbench,Video-MME} aiming to test video understanding abilities. 
These benchmarks evaluate diverse  capabilities \eg physics \cite{patraucean2023perception}, counting \cite{fang2024mmbench}, temporal reasoning  \cite{cai2024temporalbench,chen2024rextime} and long video \cite{mangalam2023egoschema,Video-MME,fang2024mmbench}.

A few benchmarks test embodied or egocentric understanding. 
\cite{Ego4D2022CVPR} released a Natural Language Queries (NLQ) benchmark (19.2K queries) centred around episodic memory of objects.
\cite{OpenEQA2023}~collects 1.6K human-made questions and answers on topics such as relative object locations, episodic memory, and spatial reasoning. However, it uses views from the HM3D \cite{ramakrishnan2021habitat} and ScanNet~\cite{dai2017scannet} datasets, so these questions are based on passive views of a static environment. \cite{mangalam2023egoschema, ye2024mm} auto-generate 5K and 7K questions based on Ego4D narrations. Whilst this approach is efficient, it is limited to these short narrations.
\cite{cai2024temporalbench} collects its own annotations for videos from several datasets, including Ego4D. %
Their benchmark is solely focused on temporal questions related to ordering, counting, causality and direction.

To evaluate a wider range of capabilities, a wider range of annotations are required. 
Of particular note are 3D grounding annotations.
Ego4D~\cite{Ego4D2022CVPR} contains some environment scans and static 3D object locations. With SLAM-equipped devices \cite{zhao2024instance} builds a benchmark for 3D object tracking; \cite{pan2023aria} contains an office and living room digital twin; and \cite{EgoExo4d} contains ego- and exo- views of expert tasks.

In contrast to these works which focus on only a few annotation types, we collect the most comprehensive set of annotations in one dataset, including highly detailed narrations, object and hand segmentations, and a comprehensive 3D digital twin of the scene and objects, all from unscripted egocentric footage in participants' homes.

\vspace{-0.5em}
\section{Data Collection}
\vspace{-0.3em}
\label{sec:Data}

\noindent\textbf{Recruitment and Equipment. }
Each participant engaged in a long commitment ($\sim$50 hours) involving  data recording and providing detailed narrations, recipes and nutrition information. %
Data was collected with Project Aria glasses~\cite{Somasundaram2023ProjectAA}---a multi-sensor platform %
with 3 forward cameras (1 RGB and 2 SLAM), 7 microphones and inward cameras for gaze estimation. %
We collected 30 FPS RGB videos at $1408\times1408$ resolution, 60 FPS eye tracking and 30 FPS SLAM.
We supplied participants with multiple devices including scales for nutritional tracking (see Fig.~\ref{fig:backpack}).

\begin{figure*}[t]
    \captionsetup[subfigure]{labelformat=empty,position=top}
    \centering
    \vspace{-1em}
    \subfloat[Recorded over 3 days]{\includegraphics[height=125pt]{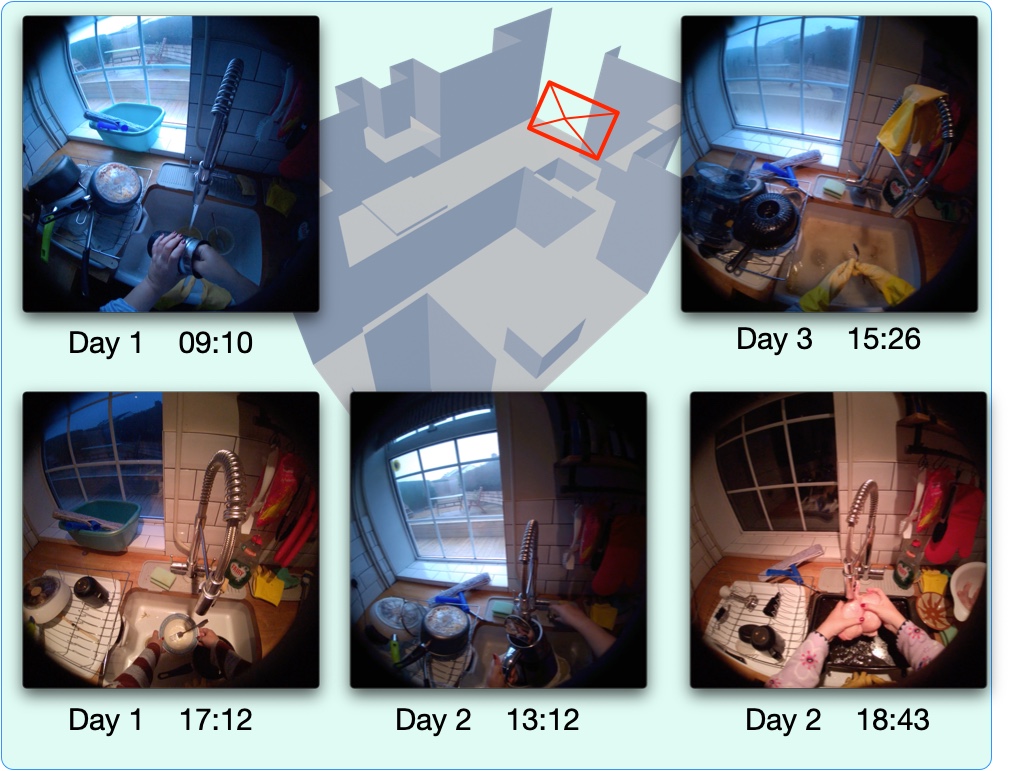} }    
    \subfloat[Objects]{ \includegraphics[height=125pt]{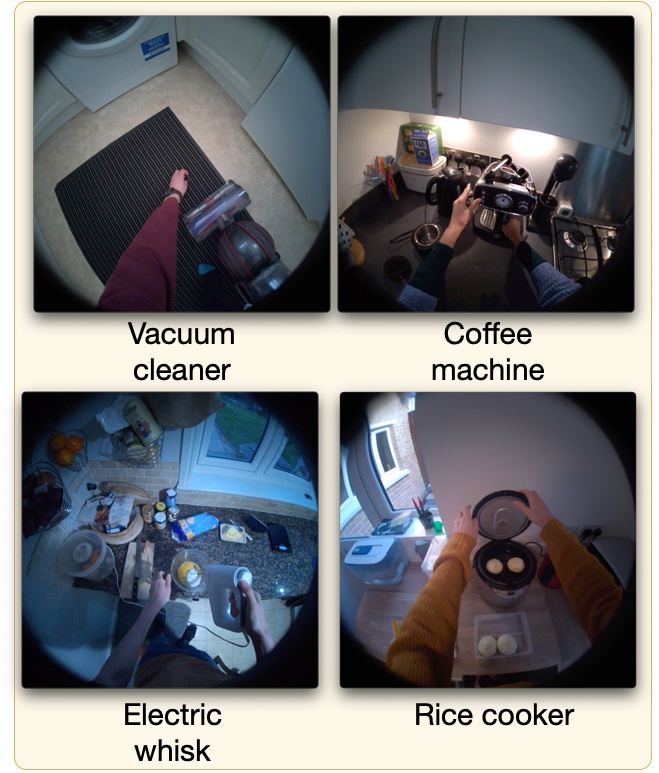} }    
    \subfloat[Activities]{ \includegraphics[height=125pt]{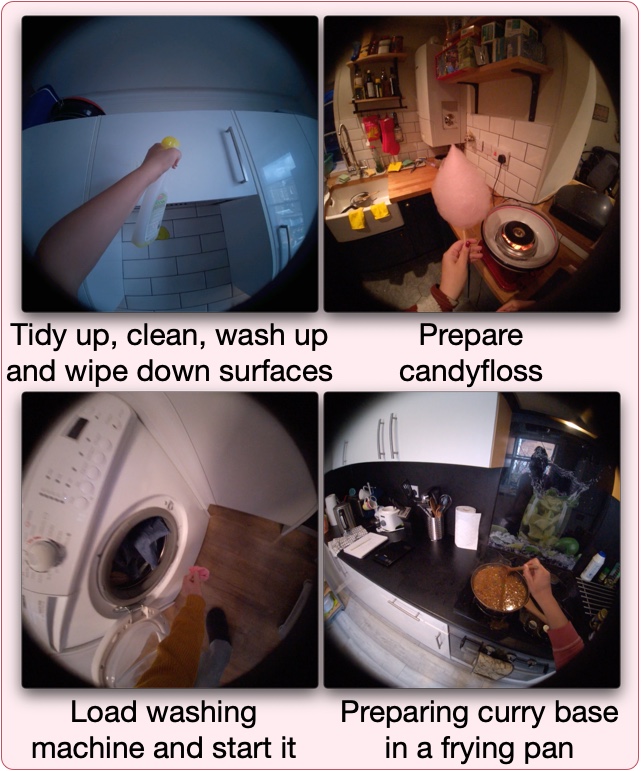} }    
    \subfloat[Recipes]{ \includegraphics[height=125pt]{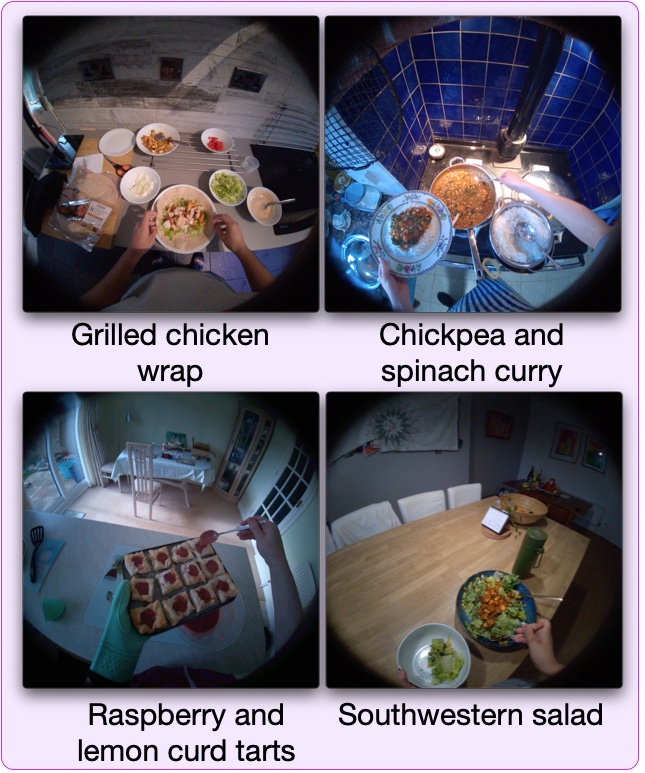} }   
    \vspace{-0.8em}
    \caption{Diversity in \DName, which is filmed over 3 days in-the-wild, resulting in many objects, activities and recipes.}
    \vspace{-1em}
    \label{fig:diversity}
\end{figure*}

\noindent\textbf{Instructions and Collected Data. }
Participants recorded all their daily kitchen activities for at least 3 consecutive days.  
All 9 participants were asked to wear the glasses each time they walked into their kitchen, pressing record upon entering, and stopping the recording when they left the kitchen.
Participants recorded for 3.5 to 7.2 hours (avg. 4.6). 
Overall, we collected 156 videos, with an average length of 15.9 ($\pm$ 14.5) minutes totalling 41.3 hours (4.46M frames).
Fig. \ref{fig:diversity} shows the diversity in the collected data.

\begin{figure}[t]
    \centering
    \includegraphics[width=0.95\linewidth]{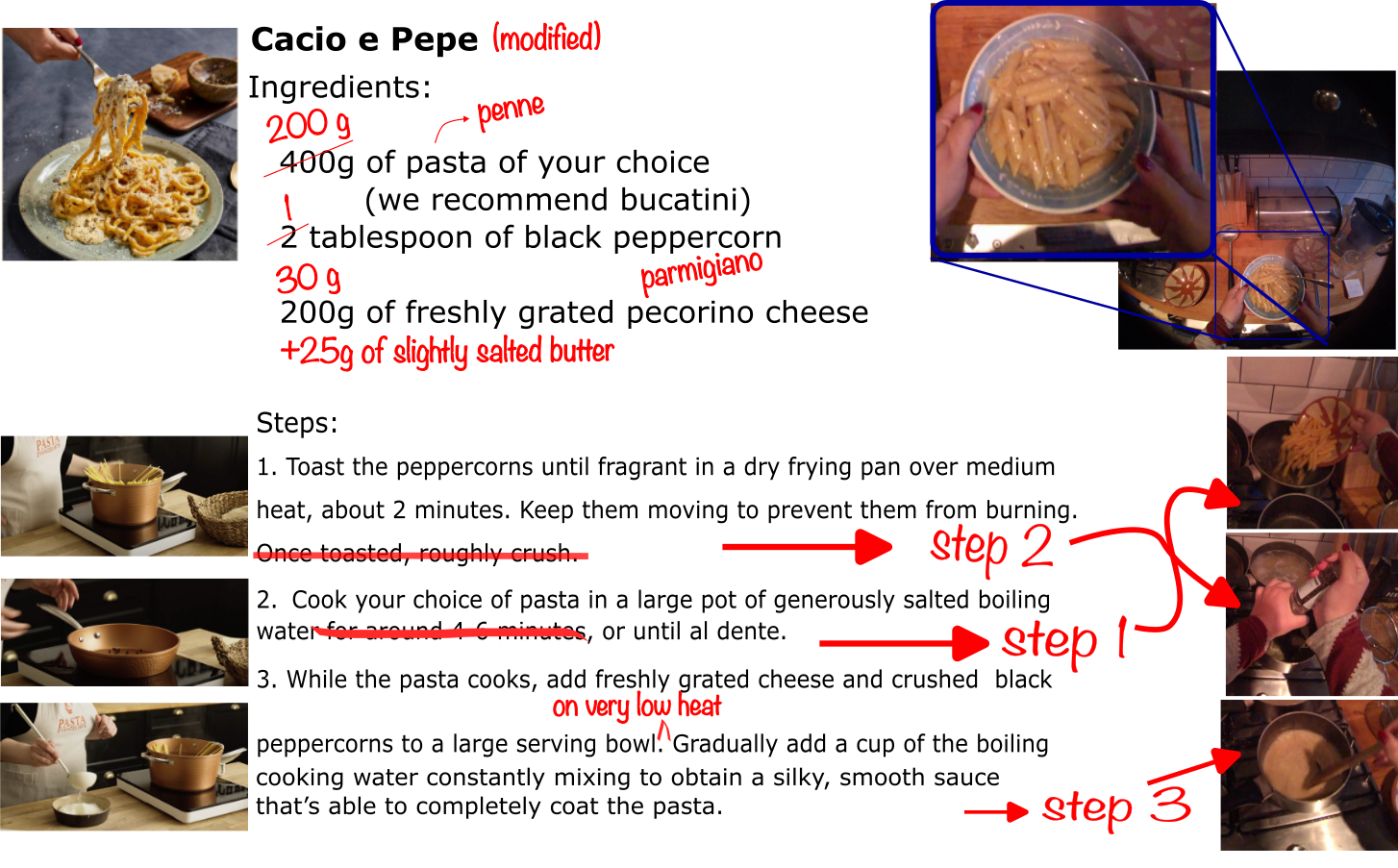}
    \vspace{-1em}
    \caption{Recipe modification in ingredients and steps.}
    \vspace*{-12pt}
    \label{fig:recipe}
\end{figure}

Following data collection, participants provided the recipes they freely prepared, citing the source (\eg website) and any modifications (see Fig.~\ref{fig:recipe}). %
We collected a total of 69 recipes  
covering various cuisines. %
On average, recipes contained 6.6 steps, 8.1 ingredients, and took 4 hours 48 minutes across 2.1 videos from preparation to serving. %
Our longest recipe %
took 2 days and 6 hours to complete.

To track nutrition of recipes, participants weighed and manually logged ingredients with MyFitnessPal~\cite{myfitness}, giving us detailed nutrition information and adding an additional dimension to the dataset. %
In total, participants used 558 ingredients including ingredients high in protein, \eg tuna and kidney beans;  carbohydrates, \eg dates and flour; and fat \eg sour cream and pine nuts. %
Participants prepared both high calorie dishes \eg Lazy Cake %
(4.8K calories)
and low calorie dishes \eg Crispy Cucumber Salad %
(274 calories).

\noindent\textbf{Narrations. }
We follow prior datasets~\cite{damen2018scaling,Damen2022RESCALING,Ego4D2022CVPR}, asking participants to watch their recordings and narrate with a web-based narrator tool~\cite{Ego4D2022CVPR}.
We expand on this by asking participants to describe \textit{what} they are doing, along with \textit{how} and \textit{why}.
This results in a rich set of narrations that are denser, and more detailed than previous datasets (\eg 3.8$\times$ more words/min than Ego4D).
See stats in Supp.~\ref{sec:data_collection_sup}.

\noindent\textbf{Post-Processing---Multi-Video Slam and Gaze. }
We use Aria %
MPS~\cite{ariamps} to process videos obtaining singular multi-day point clouds per kitchen; 1kHz 6DoF camera trajectories; and eye gaze direction.
We post-process VRS files, converting videos to mp4, removing the gaze camera input for anonymity. %
Further details are in Supp.~\ref{sec:data_collection_sup}.

\vspace{-0.5em}
\section{Annotation Pipeline}
\vspace{-0.3em}
\label{sec:annotations}

We collect extensive multi-tiered annotations to achieve the level of detail that distinguishes \DName from other video understanding datasets. Here we detail our pipeline.

\vspace{-0.3em}
\subsection{Annotating Recipe Steps and Ingredients}
\vspace{-0.2em}
\label{sec:annotations-recipe}

Our videos are distinct from short recipe videos found online, which are typically trimmed to only crucial steps, and often edited further or sped up. Videos in \DName include a wider range of recipe-relevant activities, such as fetching or prepping ingredients. %
To comprehensively annotate these videos, we introduce prep and step pairs. 

 The \textit{prep} of a corresponding \textit{step} is defined as all essential actions the participant takes to get ready to execute a given step. For example, the prep of the step `chop tomato', includes retrieving the tomato from storage, washing it, and gathering the knife and chopping board. 
However, if the step is `Add chopped onions and stir', then the chopping of onions is part of the prep for that step. %
This introduces a more fine-grained understanding of all steps, unexplored in prior datasets~\cite{song2024ego4d,lai2023lego,EgoExo4d}.
\begin{figure*}
    \centering
    \vspace{-0.5em}
    \includegraphics[width=1.0\linewidth]{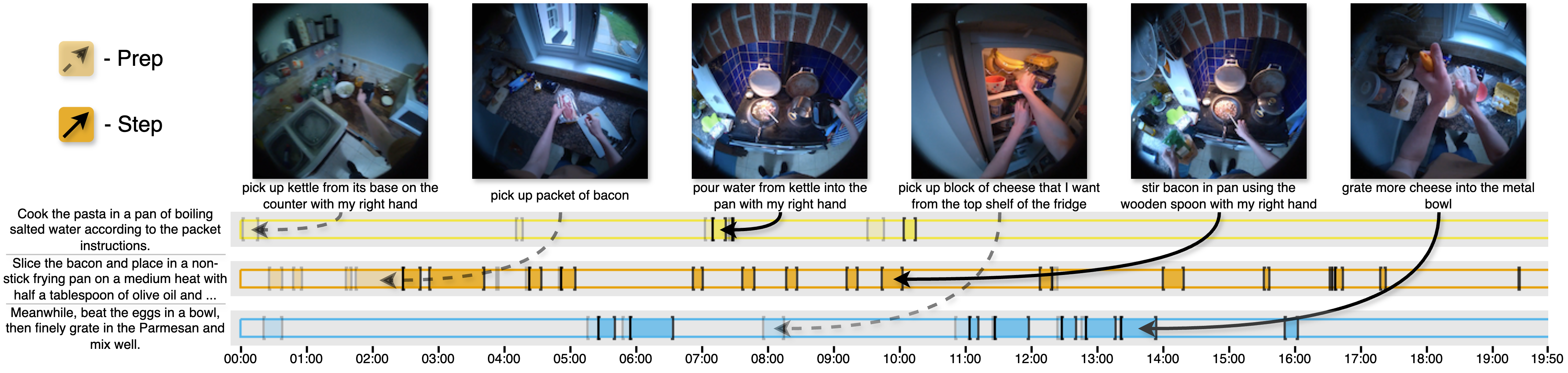}
    \vspace{-2em}
    \caption{For the `Carbonara' recipe, we visualise the \textit{\textcolor{gray}{prep}} and \textit{step} time segments for three consecutive steps (left), along with sample frames with corresponding action narrations (top). 
    The interleaving of different \textit{\textcolor{gray}{preps}}/steps is evident in the annotations.}
\vspace{-0.8em}
    \label{fig:prep-step}
\end{figure*}
Fig.~\ref{fig:prep-step} shows sample prep-step annotations for 3 steps. %
Nearly all steps (93.1\%) have paired prep annotations.
Typically, prep is shorter than a step:
avg. prep is   54.5s ($\pm$95.3s), avg. step is 78.2s ($\pm$100.7s). %
\begin{figure*}[t]
    \centering
    \includegraphics[width=1.0\linewidth]
    {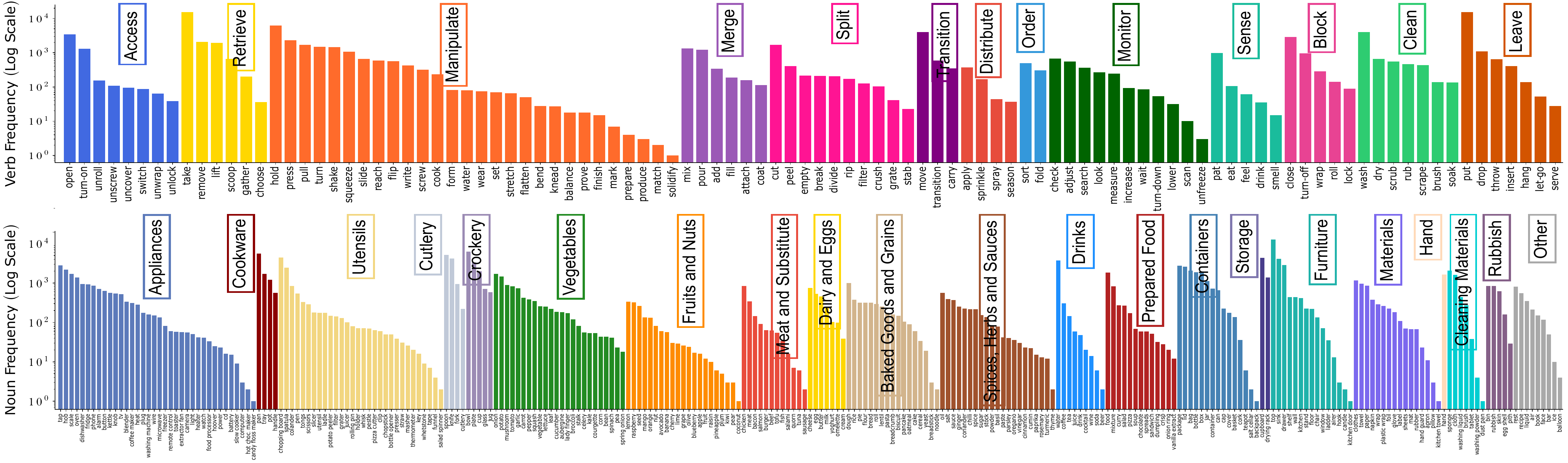}
    \vspace{-2.1em}
    \caption{Frequency of verb clusters (top) and noun clusters (bottom) in narrated sentences by category, shown on a logarithmic scale.}
    \vspace{-12pt}
    \label{fig:verb-noun-dist}
\end{figure*}

\begin{figure}
    \centering
    \vspace{-0.6em}
    \includegraphics[width=1\linewidth]{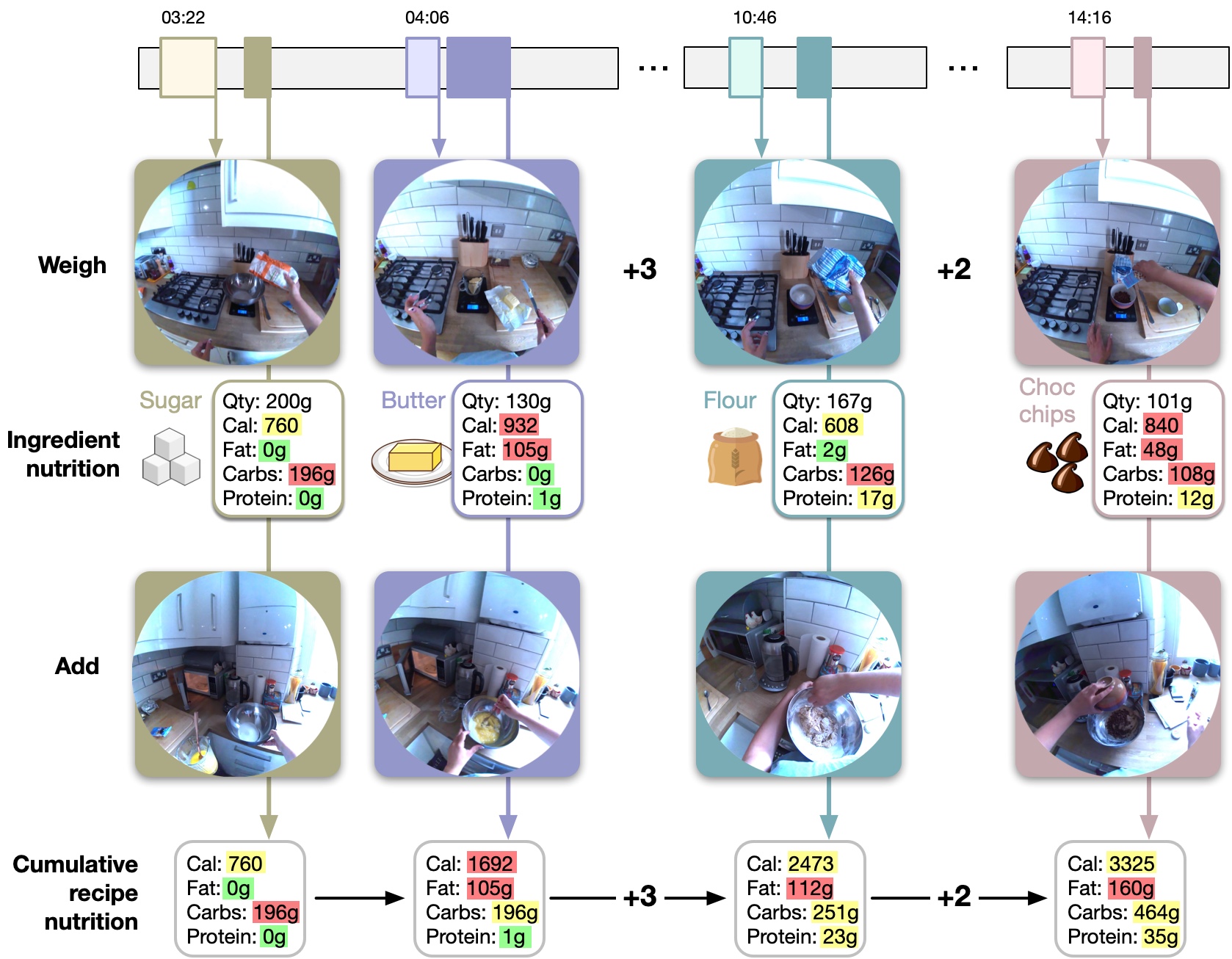} %
    \vspace{-2em}
    \caption{Nutrition is monitored throughout recipes as ingredients are incorporated into dishes. Here we show Banana Bread Chocolate Chip Cookies. We annotate when ingredients are weighed, document their nutrition, and locate their adding time, giving us overall dish nutrition at each stage.}
     \vspace{-12pt}
    \label{fig:nutritionTracker}
\end{figure}

 We also annotate \textit{weighing} and \textit{adding} temporal segments which enables monitoring the nutrition of the full dish as ingredients are incorporated (see
Fig.~\ref{fig:nutritionTracker}).
In total, we annotate 283 in-view weighing sequences (avg. 18.9s) and 501 adding sequences (avg. 31.6s), excluding spices. %
Details of the annotation process are in Supp.~\ref{sec:annotations_pipeline_supp}.

\vspace{-0.3em}
\subsection{Fine-Grained Actions}
\vspace{-0.2em}
\label{sec:annotations-narrations}

\noindent \textbf{Transcription.} We automatically transcribe and manually check and correct 
all audio narrations provided by participants, to obtain detailed action descriptions. %

\noindent \textbf{Action Boundaries.} For all narrations, we label precise start and end times. %
In total, we obtain segments for 59,454 actions, with a mean duration of 2.0s (${\pm}$3.4s).

{
\setlength{\fboxsep}{1pt}
\noindent \textbf{Parsing.}
We parse \colorbox{verb}{\vphantom{Ay}verbs}, 
\colorbox{noun}{\vphantom{Ay}nouns} and \colorbox{hand}{\vphantom{Ay}hands} from open vocabulary narrations so they can be used for closed vocabulary tasks, such as action recognition. We also extract \colorbox{how}{\vphantom{Ay}how} and \colorbox{why}{\vphantom{Ay}why} clauses from 16,004 and 11,540 narrations, respectively. 
For example, ``\colorbox{verb}{\vphantom{Ay}Turn} the \colorbox{noun}{\vphantom{Ay}salt container} \colorbox{how}{clockwise by pushing it} with my \colorbox{hand}{\vphantom{Ay}left hand} \colorbox{why}{\vphantom{Ay}so that the lid} \colorbox{why}{is aligned with the container opening.}''}

\noindent \textbf{Clustering.} 
Fig.~\ref{fig:verb-noun-dist} shows the distribution of clusters (\ie classes) across all videos in \DName, along with hierarchical clusters~\cite{Damen2022RESCALING}. As with prior datasets~\cite{Damen2022RESCALING,Ego4D2022CVPR}, our highly diverse actions and objects are long-tailed.

\noindent\textbf{Sound Annotations. }
We follow~\cite{EPICSOUNDS2023} to collect audio annotations.
These capture start-end times of audio events along with a class name (e.g. `click', `rustle', `metal-plastic collision', `running water').
Overall, we have 50,968 audio annotations from 44 classes. %

Full details of transcription, boundary labelling, parsing, clustering and sound annotations are in Supp.~\ref{sec:annotations_pipeline_supp}.

\vspace{-0.3em}
\subsection{Digital Twins: Scene \& Object Movements}
\vspace{-0.2em}

\noindent \textbf{Scene.}
We create digital copies of participants' kitchens by reconstructing the surfaces and manually curating every fixture (\eg cupboard, drawer), storage space (\eg shelves, hooks) and large appliance (\eg fridge, microwave).
This is distinct from digital twins that rely on known environments with replicas. 
Our digital twin is created in Blender \cite{blender} on top of the multi-video SLAM point clouds from recordings. %

\begin{figure}
    \centering
    \vspace{-1em}
    \includegraphics[width=1.0\linewidth]{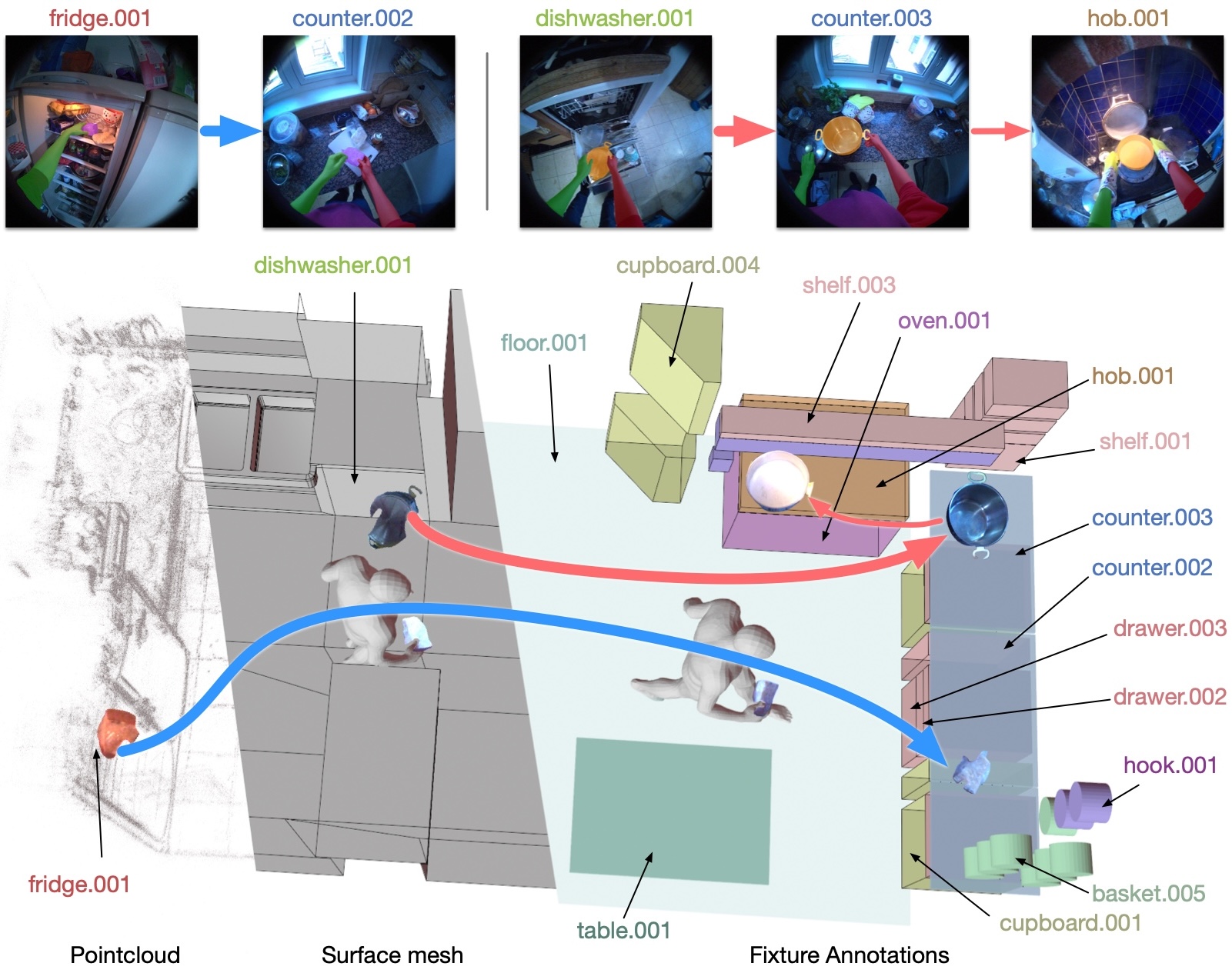}
    \vspace{-2em}
    \caption{Digital Twin: from point cloud (left), to surfaces (middle) and labelled fixtures (right). We show two moved objects (masks on top) at fixtures: \textcolor{figblue}{cheese} and \textcolor{figred}{pan}.
    Body poses from~\cite{yi2024egoallo}.
    }
    \vspace{-10pt}
    \label{fig:3D-twin}
\end{figure}

Each kitchen contains an average of 45.9 labelled fixtures (min 31, max 62), including 14.2 counters/surfaces, 12.2 cupboards, 7.8 drawers and 5.2 appliances (sample in Fig. \ref{fig:3D-twin}). We refer to these annotations as Fixtures $F$.

We then associate narrations which describe scene interactions with $F$. We find actions where a noun indicates a fixture, \eg ``open drawer'', identify the exact ``drawer'' in the digital twin (\eg drawer.001) and update its state. Following studies showing humans fixate up to 1 second before interacting \cite{land1999roles}, we take the fixture $f{\in}F$ with the highest cumulative gaze intersection for the 1s before the narration.

\noindent \textbf{Hand Masks.} We annotate a handful of frames per video for %
both hands.
Frames are selected to cover various actions and kitchen locations. %
We use these to automatically segment, and manually correct a selected subset. %
In total, our dataset contains 
7.7M hand masks: 3.9M right and 3.8M left of which 11K are
manually annotated (details in Supp.~\ref{sec:annotations_pipeline_supp}).

\noindent \textbf{Moving Objects in 2D.} To generate 3D object movement annotations, we first annotate when objects move. %
Annotators %
label a temporal segment each time an object is moved until it is set, along with 2D object bounding boxes at the onset and end of motion. %
For example, if a person moves a cup from a countertop to the sink, one bounding box captures the cup on the countertop and another when in the sink.
Tracks are annotated even for slight shifts/pushes, and thus offer full annotations of all object movements.

Overall, we collected 19.9K object movement tracks and 36.9K bounding boxes.
We label an average of 9.2 objects taken and 9.0 objects placed per minute.
On average, tracks are 9.0s long, the longest is 461.5s. 

\noindent \textbf{Object Masks.} 
Despite progress in segmentation~\cite{kirillov2023segment, ravi2024sam} and available annotations~\cite{VISOR2022,EgoExo4d}, models perform poorly in egocentric video, particularly under occlusions.
We obtain 
pixel-level segmentations from each bounding box by
initialising with iterative SAM2~\cite{ravi2024sam} then manually correcting. 
Annotators corrected 74\% of masks; the IoU between SAM2 and the manual masks is 0.82.

\noindent \textbf{Masks to 3D.} We lift object masks to 3D using dense depth estimates and 2D-to-3D sparse correspondences provided by MPS. %
Given metric depth from~\cite{depthanything}, we identify $S$, the set of pixels within or around the object with 3D correspondences. We then find the linear transformation coefficients:
$\alpha, \beta = \underset{\alpha, \beta}{\mathrm{argmin}} \, \| (\alpha \hat{D}_S + \beta ) - D_S \|^2.
$ where $\hat D_S$ are estimated depth values and $D_S$ are existing depth values, followed by RANSAC to remove outliers.

\noindent \textbf{3D Object Motion. } Objects move 61.4cm ($\pm$84.5cm) on average, 27.6\%  move $\leq$10cm, while 7.6\% move $\geq$2m.

\noindent \textbf{Object-Scene Interactions.} With the 3D object locations, we associate locations with the closest fixture $f{\in}F$, subject to fixture-specific heuristics (\eg objects must be within a counter's x-y plane). 
We manually verify all assignments, correcting any errors. %
On average objects move between 1.8 different fixtures per video (see Supp.~\ref{sec:annotations_pipeline_supp} for stats).

\begin{table*}[t]
\vspace{-1em}
\centering
\setlength{\tabcolsep}{3pt}
\resizebox{\textwidth}{!}{%
\begin{tabular}{@{}lcccccccccccccc@{}}
\toprule
\textbf{Dataset} & \textbf{Val\&Test} & \textbf{Action} & \textbf{Unscripted} & \textbf{Free} & \textbf{Recipe} & \textbf{Nutrition} & \textbf{Gaze} & \textbf{Audio} & \textbf{Object} & \textbf{Hand} & \textbf{3D object} & \textbf{Labelled 3D} & \textbf{Camera} & \textbf{Fully} \\
& \textbf{Hours} & \textbf{Segments} & & \textbf{Setting} & & & & \textbf{Labels} & & & \textbf{over time} & \textbf{environment} & \textbf{pose} & \textbf{annotated} \\
\midrule
HOI4D~\cite{liu2022hoi4d} & 11.4 & \cmark &\xmark &\xmark & \xmark & \xmark & \xmark & \xmark & Mask & Mask &  \cmark & \cmark & \xmark & \cmark \\
Assembly101~\cite{sener2022assembly101} & 66.8 & \cmark &\xmark &\xmark & \xmark & \xmark & \xmark & \xmark & \xmark & 3D pose & \xmark & \xmark & \xmark & \cmark \\
EPIC-KITCHENS-100~\cite{Damen2022RESCALING} & 25.3 & \cmark &\cmark &\cmark & \xmark & \xmark & \xmark & \cmark & Mask & Mask & \cmark & \xmark & \cmark & \xmark \\
Ego4D~\cite{Ego4D2022CVPR} & 288.7 & \cmark &\cmark &\cmark & \xmark & \xmark & \xmark & \xmark & B-Box & B-Box &  \xmark & \xmark & \xmark & \xmark \\
HoloAssist~\cite{HoloAssist2023} & 49.8 & \cmark & \xmark & \xmark & \xmark & \xmark & \cmark & \xmark & \xmark  & 3D pose & \xmark & \xmark & \cmark & \cmark \\ 
Aria Digital Twin~\cite{pan2023aria} & 8.1 & \xmark & \xmark & \xmark & \xmark & \xmark & \cmark & \xmark & Mask & \xmark &  \cmark & \cmark & \cmark  & \cmark \\
Aria Everyday Activities~\cite{lv2024aria} & 7.3 & \xmark & \xmark &\cmark & \xmark & \xmark & \cmark & \xmark & \xmark & \xmark & \xmark & \xmark & \cmark &  \xmark \\
Aria Everyday Objects~\cite{straub2024efm3d} & 0.4 & \xmark &\cmark &\cmark & \xmark & \xmark & \xmark & \xmark & B-Box & \xmark & \cmark & \xmark & \cmark  & \cmark \\
Ego-Exo4D~\cite{EgoExo4d} & 85.1 & \xmark & \cmark & \cmark & \cmark & \xmark & \cmark & \xmark & Mask & 3D pose & \xmark & \xmark & \cmark  & \xmark \\ \midrule
\textbf{\DName} & 41.3 & \cmark & \cmark & \cmark &\cmark & \cmark & \cmark & \cmark & Mask & Mask &  \cmark & \cmark & \cmark  & \cmark \\ \bottomrule
\end{tabular}%
}
\vspace*{-10pt}
\caption{\textbf{Comparison of Egocentric Video Datasets} (see full table~\ref{tab:dataset_comparison_full} in Supp.). 
}
\vspace*{-12pt}
\label{tab:dataset_comparison}
\end{table*}

\begin{figure}
    \centering
    \vspace{-1em}
    \includegraphics[width=1.0\linewidth]{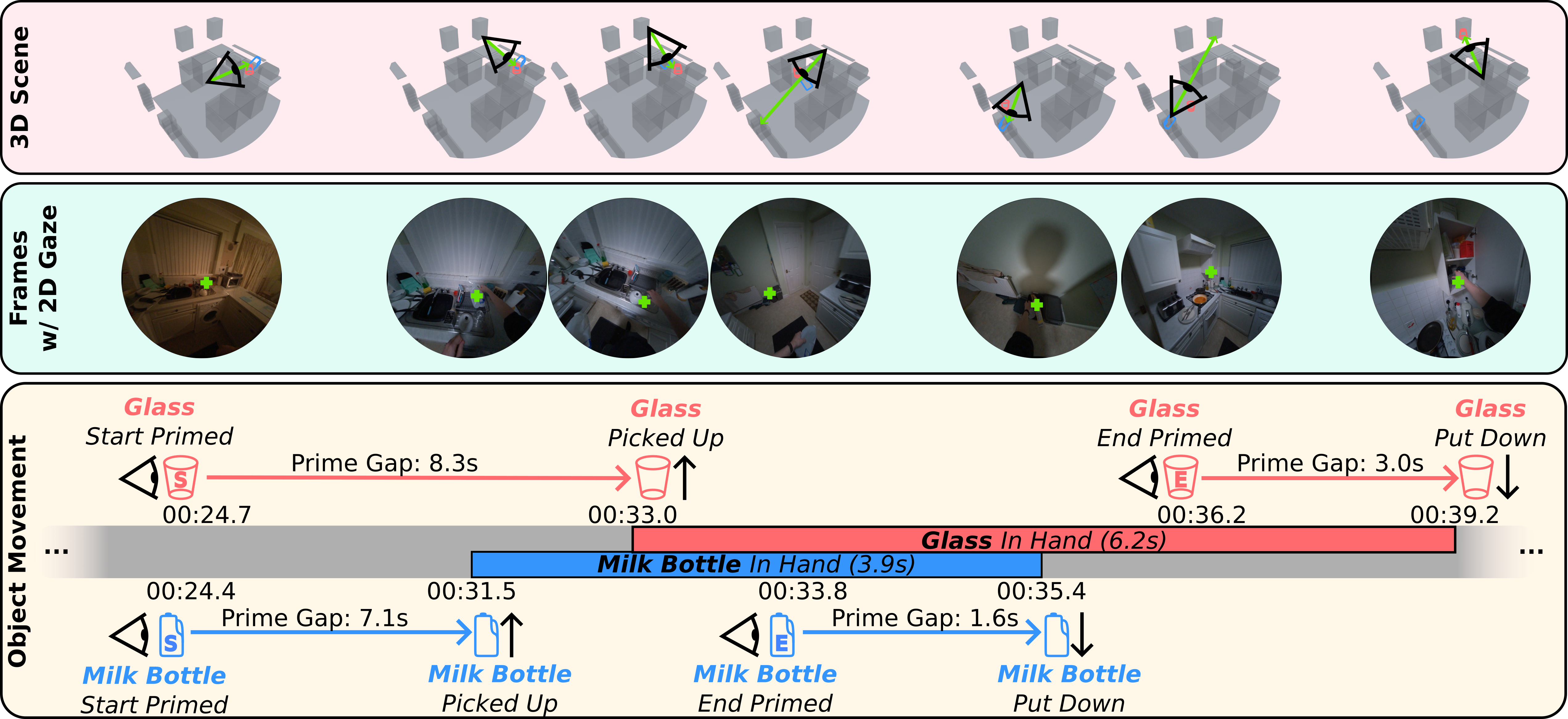}
    \vspace{-2em}
    \caption{\textbf{Priming Object Interaction Through Gaze}. Top: Camera position with projected eye-gaze and object positions in 3D. Middle: 2D gaze location. Bottom: Timeline for priming object movement \eg the glass is primed 8.3s before taking.}
    \vspace{-6pt}
    \label{fig:prime_fig}
\end{figure}

\noindent \textbf{Priming Object Movement.}
The behaviour of gaze when picking up and placing objects is well-studied~\cite{land1999roles,johansson2001eyehand}. %
We combine eye-gaze and 3D object locations, to find when an object is \textit{primed}, \ie the moment in time when the gaze attends to the object's location before picking it up (\textit{pick-up priming}) or when the gaze attends to the future location of an object before it's put down (\textit{put-down priming}).

\begin{figure}
    \centering
    \resizebox{\linewidth}{!}{
        \begin{tabular}{c|cc|cc|c}
        \toprule
        Location  
        & \makecell{Filtered\\(\% Total)} & \makecell{Feasible\\(\% Total)} & \makecell{Primed\\(\% Feasible)} & \makecell{Not Primed\\(\% Feasible)} &  Avg. Time (s) \\
        \midrule
        Start & 
        29.40 & 70.60 & 94.82 & 5.18 & $3.99 (\pm2.94)$ \\
        End & 
        66.92 & 33.08 & 88.46 & 11.54 & $2.62 (\pm2.05)$ \\
        \bottomrule
        \end{tabular}
    }
    \includegraphics[width=1.0\linewidth]{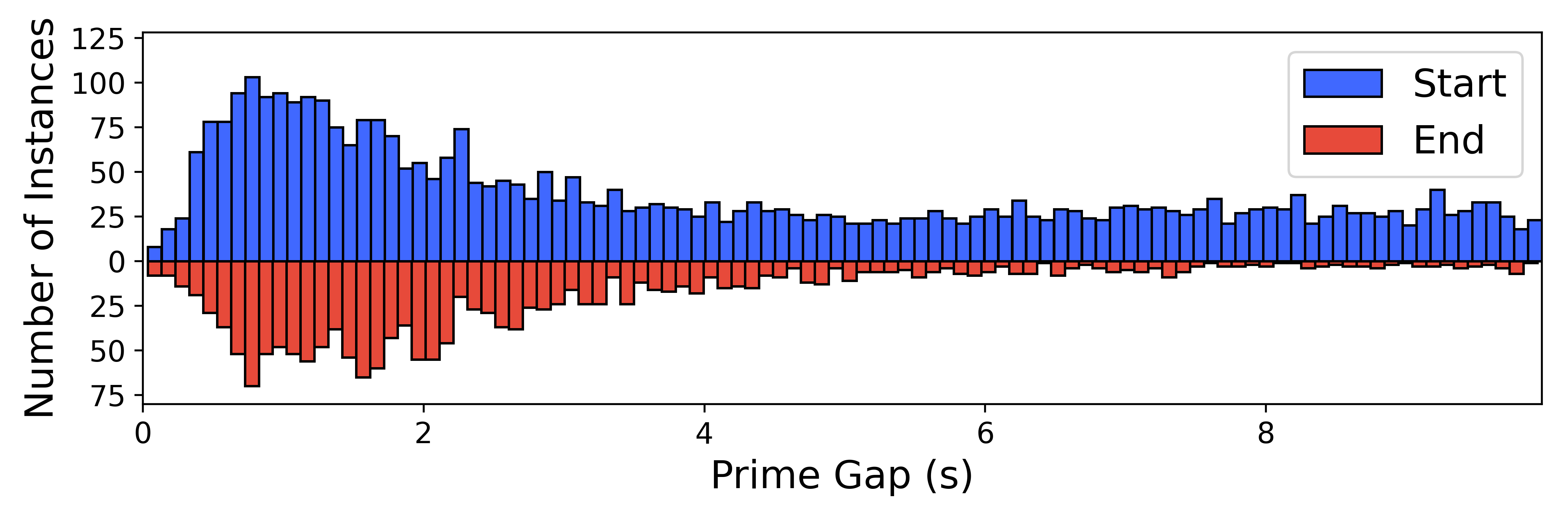}
    \vspace{-2.4em}
    \caption{(Top) Priming Statistics for both start and end locations (Bottom) Histogram showing the difference in time when an object is primed before it is picked up (blue) or placed (red).}
    \vspace{-0.8em}
    \label{fig:prime_stats}
\end{figure}

We calculate the \textit{priming time} for all objects, excluding those taking or placed off screen. 
Additionally, at times, a person is already manipulating an object well before picking it up.
We thus exclude objects with a pick up location already close to the gaze 10s earlier.
 In Fig.~\ref{fig:prime_fig} we show gaze priming for two objects: milk bottle and glass. %
The glass's end location, a cupboard, 
is primed 3s before the glass is put away.
Fig.~\ref{fig:prime_stats} displays priming statistics. %
Of those objects feasible for priming, 94.8\%
are primed, an average of 4.0s before being picked up, compared to 88.5\%
primed an average of 2.6s before being placed.

\noindent \textbf{Long Term Object Tracking.}
\label{sec:long_term_obj_tracks}
We connect object movements and form longer trajectories, \ie object itineraries, to capture sequences of an object's movement.
Our efficient pipeline utilises our lifted 3D locations  
and allows annotating a 1-hour long video in minutes (details in Supp.~\ref{sec:annotations_pipeline_supp}).

\vspace{-0.3em}
\subsection{\DName vs Prior Egocentric Datasets}
\vspace{-0.2em}
Tab.~\ref{tab:dataset_comparison} compares \DName to other egocentric datasets (full table in Supp.).
Compared to the largest dataset with labelled 3D environments (Aria Digital Twin~\cite{pan2023aria}), \DName contains 5x more footage; has more annotations; and importantly was collected in an unscripted manner in the participants' homes. %
In particular, \DName is the first to annotate recipes, nutritional values, detailed action segments, gaze and audio labels on the same set of videos. 
With these diverse and dense annotations, \DName constitutes a true zero-shot benchmark for video understanding.

\vspace{-0.5em}
\section{Benchmarks and Results}
\vspace{-0.3em}
\label{sec:benchmarks}

We show the potential of \DName as a validation dataset with benchmarks on
 general Video Question Answering (VQA) (Sec~\ref{sec:benchmark-vqa}), action and sound recognition (Sec~\ref{sec:benchmark-recognition}) and long-term video object segmentation (Sec~\ref{sec:benchmark-long}).%

\begin{figure}
    \centering
    \vspace{-1em}
    \includegraphics[width=0.8\linewidth]{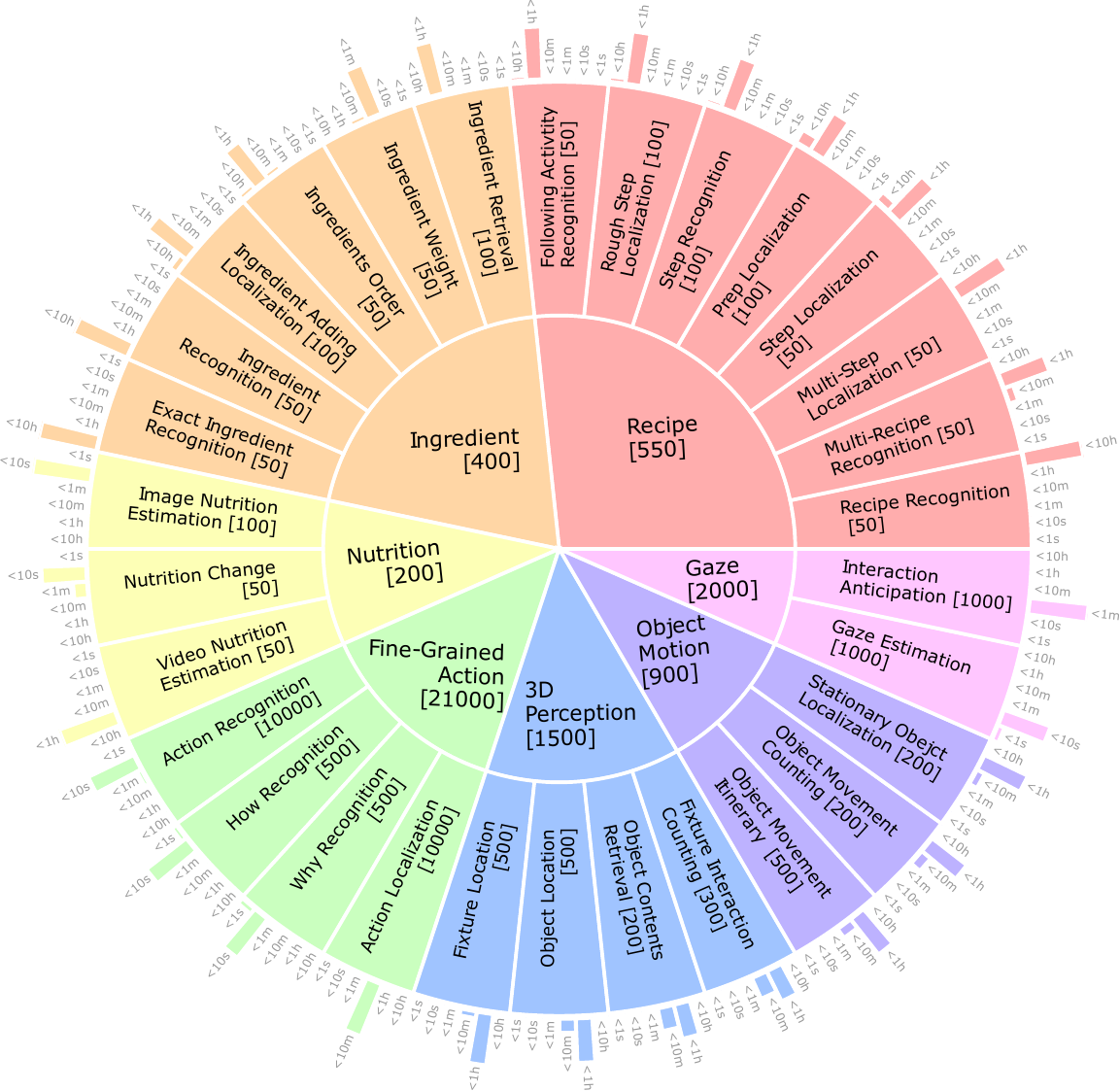}
    \vspace{-1.5em}
    \caption{\textbf{VQA Question Prototypes}. We show our 30 question prototypes by category alongside the number of questions. Outer bars indicate the distribution over input lengths for each question. %
    }
    \label{fig:vqa_prototypes}
    \vspace{-0.5em}
\end{figure}

\vspace{-0.3em}
\subsection{\DName VQA Benchmark and Analysis}
\vspace{-0.2em}
\label{sec:benchmark-vqa}
\setlength{\fboxsep}{2pt}
\noindent\textbf{Benchmark Creation}. We take the dense output of our annotation pipeline and construct a comprehensive VQA benchmark around 7 types of annotations:
\begin{enumerate}
    \item \recipe{Recipe}. Questions on temporally localising, retrieving, or recognising recipes and their steps.
    \item \ingredient{Ingredient}. Questions on the ingredients used, their weight, their adding time and order.
    \item \nutrition{\vphantom{Ay}Nutrition}. Questions on nutrition of ingredients and nutritional changes as ingredients are  added to recipes.
    \item \fine{Fine-grained action}. What, how, and why of actions and their temporal localisation.
    \item \threed{3D perception}. Questions that require the understanding of relative positions of objects in the 3D scene. %
    \item \object{Object motion}. Questions on where, when and how many times objects are moved across long videos.
    \item \gaze{Gaze}. Questions on estimating the fixation on large landmarks and anticipating future object interactions.
\end{enumerate}
For each question type, we define prototypes to sample questions, correct answers, and strong negatives from our annotations. For example, \object{Object Movement Counting} asks ``How many times did the object $<$bbox$>$ seen at $<$time$>$ move in the video?''. This uses long videos, requiring multiple hops to be correctly answered. In contrast, \fine{How Recognition} asks ``What is the best description for how the person carried out the action $<$verb, noun$>$?'' to test a model's ability to capture intricate details of actions.

\begin{figure*}
\vspace{-1em}
    \centering
    \includegraphics[width=1.0\linewidth]{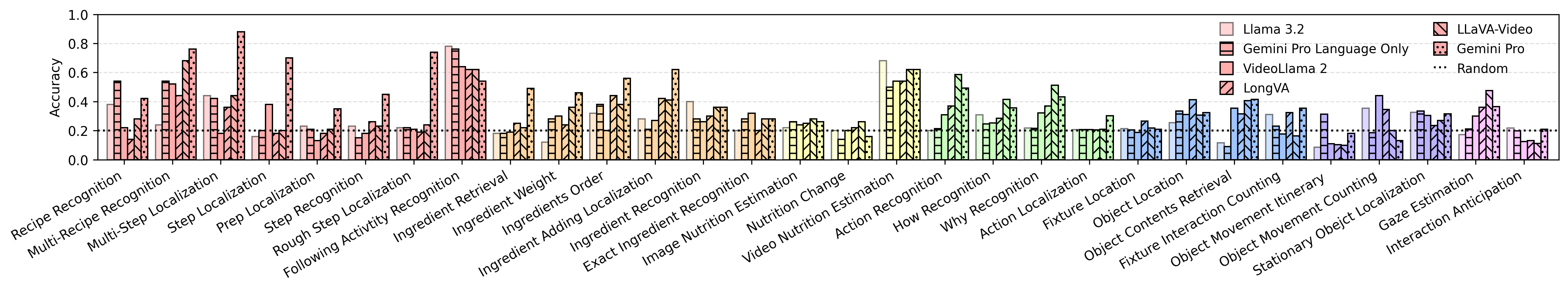}
    \vspace{-2.6em}
    \caption{\textbf{VQA Results per Question Prototype}. Our benchmark contains many challenging questions for current models.} %
    \vspace{-1em}
    \label{fig:results_per_prototype}
\end{figure*}

Each question prototype is 5-way multiple choice. We generate hard negatives for prototypes by sampling within the dataset for difficult answers.
For example, we take 4 different answers of how participants performed the same action. This ensures realistic negatives and challenging questions. %
In total, we have 30 prototypes, and generate 26,650 multiple-choice questions. 
This makes it one of the largest VQA video benchmarks, but keeps it tractable particularly to evaluate closed-source VLMs.
Due to the density of our annotations, we estimate an upper bound of 100,000 possible unique questions with this set of prototypes. 

\begin{figure*}
    \centering
    \vspace{-1em}
    \includegraphics[width=\linewidth]{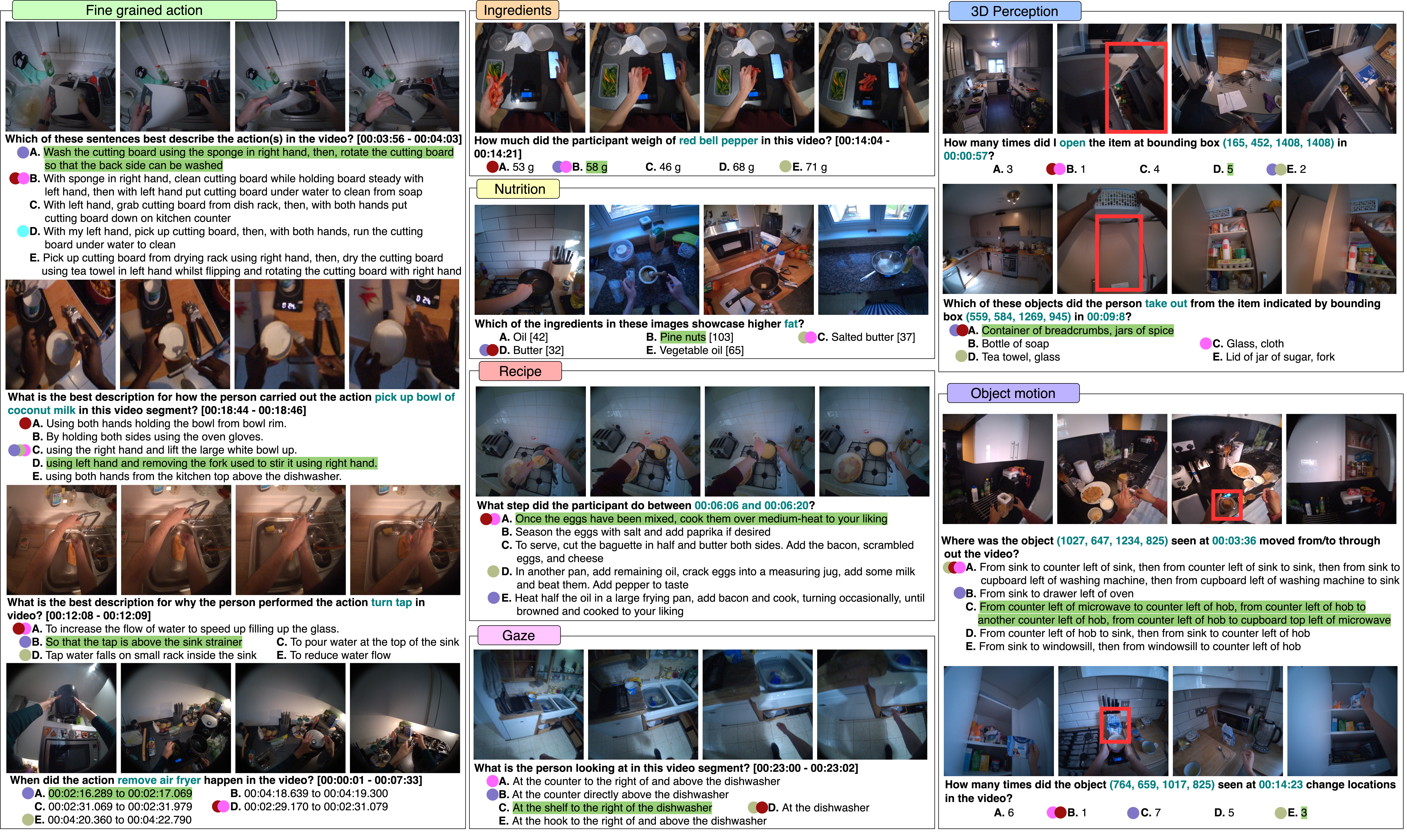}
    \vspace{-2.5em}
    \caption{\textbf{VQA Qualitative Results}. We mark GT answers with a  \colorbox{vqagreen}{green} background, and predictions from different models, \ie, \textcolor{vqallama}{LLaMA 3.2}, \textcolor{vqavlama}{VideoLLaMA 2}, \textcolor{vqalngva}{LongVA}, \textcolor{vqagemini}{Gemini Pro} with coloured dots. Note: Under Nutrition, [fat] values are not provided to the model.
    } 
    \vspace{-2em}
    \label{fig:vqa_qual_res}
\end{figure*}

Fig.~\ref{fig:vqa_prototypes} shows the distribution of 
questions per category alongside the distribution of input lengths which varies from single frames to 7+ hours. Details of each prototype's sampling can be found in Supp.~\ref{sec:benchmark_supp}.
A sample of our questions and answers can be seen in Fig.~\ref{fig:vqa_qual_res}. %

\noindent \textbf{VLM Models.} Due to the size and long-term nature of many question prototypes in our benchmark, we use 5 representative models as baselines (more details in Supp.~\ref{sec:benchmark_supp}):
\begin{itemize}
    \item Llama 3.2 90B~\cite{dubey2024llama}. We use this as a strong open-source (OS) text-only baseline, as LLMs can perform well on visual QA benchmarks \emph{without any visual input}~\cite{xiao2024can}.
    \item VideoLlama 2 7B \cite{cheng2024videollama}. OS short context model.
    \item LongVA \cite{zhang2024long}. Longest context OS model.
    \item LLaVa-Video \cite{zhang2024videoinstructiontuningsynthetic}. OS model trained also on  ego data.
    \item Gemini Pro \cite{team2024gemini}. Closed source, longest context of any model, and state-of-the-art on long-video \cite{Video-MME}.
\end{itemize}

\begin{table}[]
    \centering
    \setlength{\tabcolsep}{3pt}
    \resizebox{\linewidth}{!}
    {\begin{tabular}{lcccccccc}
    \toprule
    Model & \recipe{Recipe} & \ingredient{Ingredient} & \nutrition{\vphantom{Ay}Nutrition} & \fine{\vphantom{Ay}Action} & \threed{\vphantom{Ay}3D} & \object{\vphantom{Ay}Motion} & \gaze{\vphantom{Ay}Gaze} & Avg.\\
 \midrule
\rowcolor{LightGrey} \multicolumn{9}{l}{\textbf{Blind - Language Only}} \\
 Llama 3.2 & 33.5 & 25.0 & 36.7 & 23.3 & 22.3 & 25.5 & 19.5 & 26.5 \\
  Gemini Pro & 38.0 & 26.8 & 30.0 & 22.1 & 21.5 & 27.7 & 20.5 & 26.7\\
\rowcolor{LightGrey} \multicolumn{9}{l}{\textbf{Video-Language}} \\
 VideoLlama 2 & 30.8 & 25.7 & 32.7 & 27.2 & 25.7 & 28.5 & 21.2 & 27.4\\
  LongVA & 29.6 & 30.8 & 33.7 & 30.7 & 32.9 & 22.7 & 24.5 & 29.3\\
  LLaVA-Video & 36.3 & 33.5 & 38.7 & 43.0 & 27.3 & 18.9 & 29.3 & 32.4 \\
  Gemini Pro & 60.5 & 46.2 & 34.7 & 39.6 & 32.5 & 20.8 & 28.7 & 37.6\\
  \color{Gray}{\textit{Sample Human Baseline}} & \color{Gray}{\textit{96.7}} & \color{Gray}{\textit{96.7}} & \color{Gray}{\textit{85.0}} & \color{Gray}{\textit{92.5}} & \color{Gray}{\textit{93.8}} & \color{Gray}{\textit{92.7}} & \color{Gray}{\textit{75.0}} & \color{Gray}{\textit{90.3}}\\
         \bottomrule
    \end{tabular}}
    \vspace{-1em}
    \caption{\textbf{VQA Results per Category} (\% Acc.). Our VQA benchmark  cannot be solved blind or by external knowledge and is a challenge for state-of-the-art video VLM models.}
    \vspace{-1.5em}
    \label{tab:results_main}
\end{table}

\noindent\textbf{VQA Results Per Category and Per Prototype.} Tab. \ref{tab:results_main} provides overall and per-category accuracy averaged over the prototype results shown in Fig.~\ref{fig:results_per_prototype}. %
Both language-only models only achieve 26.5\% and 26.7\%, only 6.7\% above random. %
Open-source video VLMs (VideoLlama, LongVA, LlaVA-Video) perform similarly (27.4\%, 29.3\%, 32.4\%) but have different strengths as shown in Fig.~\ref{fig:results_per_prototype}. For example, Llama better estimates nutrition, while the video is necessary to get above random performance on action recognition and gaze estimation. Gemini achieves the best performance, particularly for \recipe{Recipe} and \ingredient{Ingredient} where external knowledge helps. However, the average performance (37.6\%) and the gap to our sample human baseline (90.3\%) shows the challenge posed by our VQA benchmark.

\begin{figure}
\vspace{-0.3em}
    \centering
    \includegraphics[width=\linewidth]{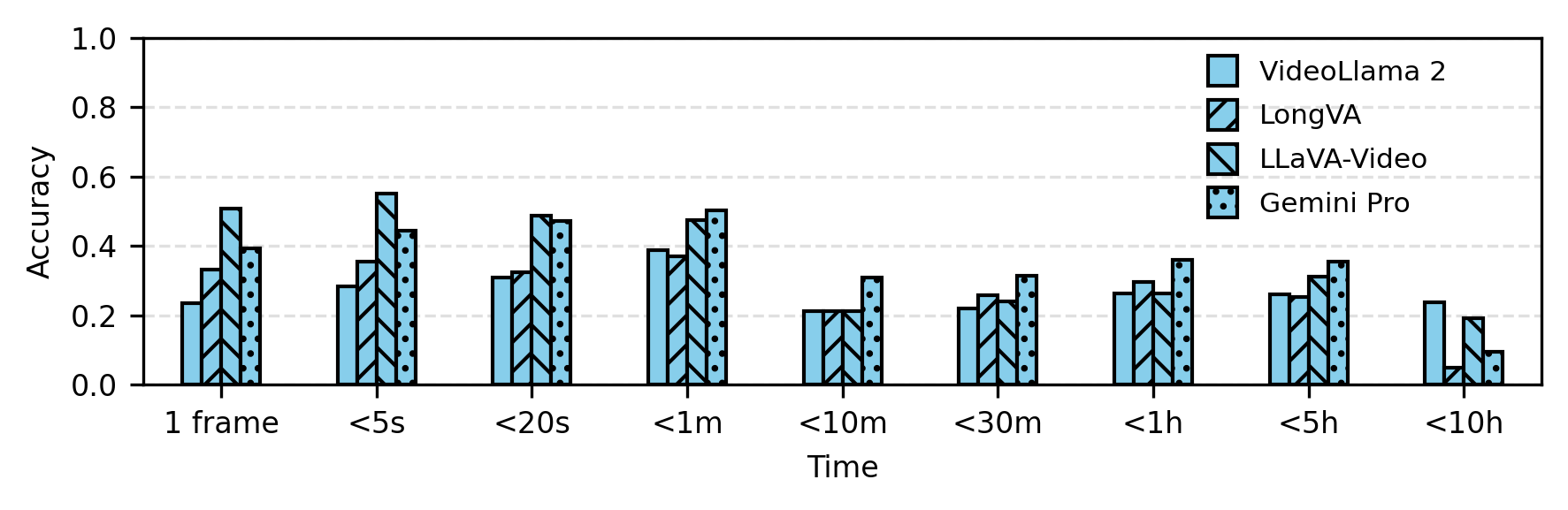}
    \vspace{-2.7em}
    \caption{\textbf{Effect of Input Length}. Models struggle with questions of all video input lengths. s=second, m=minute, 
h=hour.}
    \vspace{-1em}
    \label{fig:result_length}
\end{figure}

\noindent\textbf{Video Length.} Fig.~\ref{fig:result_length} shows
models struggle with all video lengths but are worst with inputs $\geq$1 minute.

{\setlength{\fboxsep}{1pt}
\noindent\textbf{Common Failures.} Fig.~\ref{fig:vqa_qual_res} shows qualitative results. In \recipe{Recipe}, models struggle when steps have common objects or actions. In \ingredient{Ingredient}, models guess weights (readable from the scale by humans) poorly, also causing errors in \nutrition{\vphantom{Ay}Nutrition}. \fine{Fine-grained action} is hard when answers share nouns. %
In \gaze{\vphantom{Ay}Gaze}, models just select recently moved objects. Confusion in \threed{\vphantom{Ay}3D} and \object{Object motion} occurs with directions (right/left) and fixtures (counters/drawers).  }

\begin{table}[t]
    \footnotesize
    \centering
    \setlength{\tabcolsep}{6pt}
    \resizebox{\linewidth}{!}{
    \begin{tabular}{l c c c c | c }
        \toprule
        Model & Modality & Verb & Noun & Action & {\parbox{2cm}{\centering Unseen\\EPIC-100 Action}}\\
        \midrule
                \rowcolor{LightGrey} \multicolumn{5}{l|}{\textbf{EPIC-KITCHENS-100 SOTA}} & \\
        TIM~\cite{Chalk_2024_CVPR} & A+V & 77.1 & 67.2 & 57.5 & 44.6 \\
        \rowcolor{LightGrey} \multicolumn{5}{l|}{\textbf{\DName}} & \\
        Chance & - & 10.9 & 1.8 & 0.0 & - \\
        SlowFast~\cite{Feichtenhofer_2019_ICCV} & V & 29.2 & 10.6 & 5.3 & 29.0 \\
        Omnivore~\cite{girdhar2022omnivore} & V & 19.5 & 17.1 & 8.7 & 28.7 \\
        MotionFormer-HR~\cite{patrick2021keeping} & V & 35.7 & 20.0 & 10.2 & 32.2 \\
        VideoMAE-L~\cite{tong2022videomae} & V & 47.5 & 29.4 & 17.9 & 29.3 \\
        TIM~\cite{Chalk_2024_CVPR} & A+V & \textbf{51.3} & 36.1 & 23.4 & \textbf{44.6} \\
        TIM~\cite{Chalk_2024_CVPR} & V & 51.2 & \textbf{36.5} & \textbf{23.9} & 44.4 \\
        \bottomrule
    \end{tabular}}
    \vspace{-1.2em}
    \caption{\textbf{Action Recognition Benchmark} (\% Acc.). \DName provides a significant challenge for state-of-the-art models.}
    \vspace{-2em}
    \label{tab:action_recognition}
\end{table}

\vspace{-0.3em}
\subsection{Recognition Benchmarks}
\vspace{-0.2em}
\label{sec:benchmark-recognition}

\noindent \textbf{Action Recognition.} We assess 5  action recognition methods~\cite{Feichtenhofer_2019_ICCV,patrick2021keeping,girdhar2022omnivore,tong2022videomae,Chalk_2024_CVPR}, using publicly available checkpoints fine-tuned on EPIC-KITCHENS-100.
Results are shown in Tab.~\ref{tab:action_recognition}. For context we show the results from EPIC-KITCHENS-100 (top row) and on the unseen kitchens subset of EPIC-KITCHENS-100 (last col.).
Best performance on \DName is only 51\% for verbs, 37\%  for nouns and 24\% for actions leaving plenty of room for improvement.

\noindent \textbf{Sound Recognition.} We evaluate 3 audio models~\cite{gong2022ssast,Kazakos2021SlowFastAuditory,Chalk_2024_CVPR}, 
all trained on EPIC-Sounds. 
Tab.~\ref{tab:sound_recognition} %
shows a large gap in performance comparing \DName to EPIC-Sounds for SSAST (-28.4), ASF (-25.9) and TIM (-26.4).
This shows audio is not sufficiently robust to new scenes or devices.

\begin{table}[t]
    \footnotesize
    \centering
    \setlength{\tabcolsep}{6pt}
    \resizebox{0.85\linewidth}{!}{
    \begin{tabular}{lcccccc}
        \toprule
         Model & Modality & Top-1 & Top-5 & mCA & mAP & mAUC \\
         \midrule
                 \rowcolor{LightGrey} \multicolumn{7}{l}{\textbf{EPIC-Sounds SOTA}} \\
         TIM~\cite{Chalk_2024_CVPR} & A+V & 58.3 & 86.0 & 25.8 & 30.6 & 0.879 \\
        \rowcolor{LightGrey} \multicolumn{7}{l}{\textbf{\DName}} \\
         Chance & - & 6.9 & 29.4  & 2.2 & 2.3 & 0.500 \\
         SSAST~\cite{gong2022ssast} & A & 25.1 & 59.8 & 10.8 & 13.5 & 0.748 \\
         TIM~\cite{Chalk_2024_CVPR} & A & 26.9 & 56.9 & 12.4 & 11.4 & 0.689 \\
         ASF~\cite{Kazakos2021SlowFastAuditory} & A & 27.9 & \textbf{64.0} & 11.9 & 14.0 & 0.741 \\
         TIM~\cite{Chalk_2024_CVPR} & A+V & \textbf{31.9} & 61.0 & \textbf{14.4} & \textbf{15.7} & \textbf{0.765} \\
        \bottomrule
    \end{tabular}}
    \vspace{-1em}
    \caption{\textbf{Sound Recognition Benchmark}. Current models struggle on \DName compared to the EPIC-Sounds state-of-the-art.}
    \vspace{-12pt}
    \label{tab:sound_recognition}
\end{table}

\begin{figure}
    \centering
    \resizebox{\linewidth}{!}{
        \begin{tabular}{lccccccccccc}
    \toprule
           & \multicolumn{3}{c}{Total} && \multicolumn{3}{c}{Hands} && \multicolumn{3}{c}{Objects} \\
    \cmidrule{2-4} \cmidrule{6-8} \cmidrule{10-12}
    Model & $\mathcal{J}$ & $\mathcal{F}$ & $\mathcal{J\&F}$ & & $\mathcal{J}$ & $\mathcal{F}$ & $\mathcal{J\&F}$ & & $\mathcal{J}$ & $\mathcal{F}$ & $\mathcal{J\&F}$  \\
    \midrule
    Static & 8.0 & 10.3 & 9.2 && 14.6 & 14.4 & 14.5 && 4.8 & 8.4 & 6.6 \\
    Cutie~\cite{cheng2024putting} & 44.8 & \textbf{52.3} & \textbf{48.6} && 74.8 & 79.5 & 77.2 && \textbf{30.1} & \textbf{39.0} & \textbf{34.6} \\
    SAM2~\cite{ravi2024sam} & \textbf{45.2} & 49.6 & 47.4 && \textbf{87.5} & \textbf{90.8} & \textbf{89.1} && 24.5 & 29.5 & 27.0 \\
    \bottomrule
    \end{tabular}}
    \includegraphics[width=0.9\linewidth]{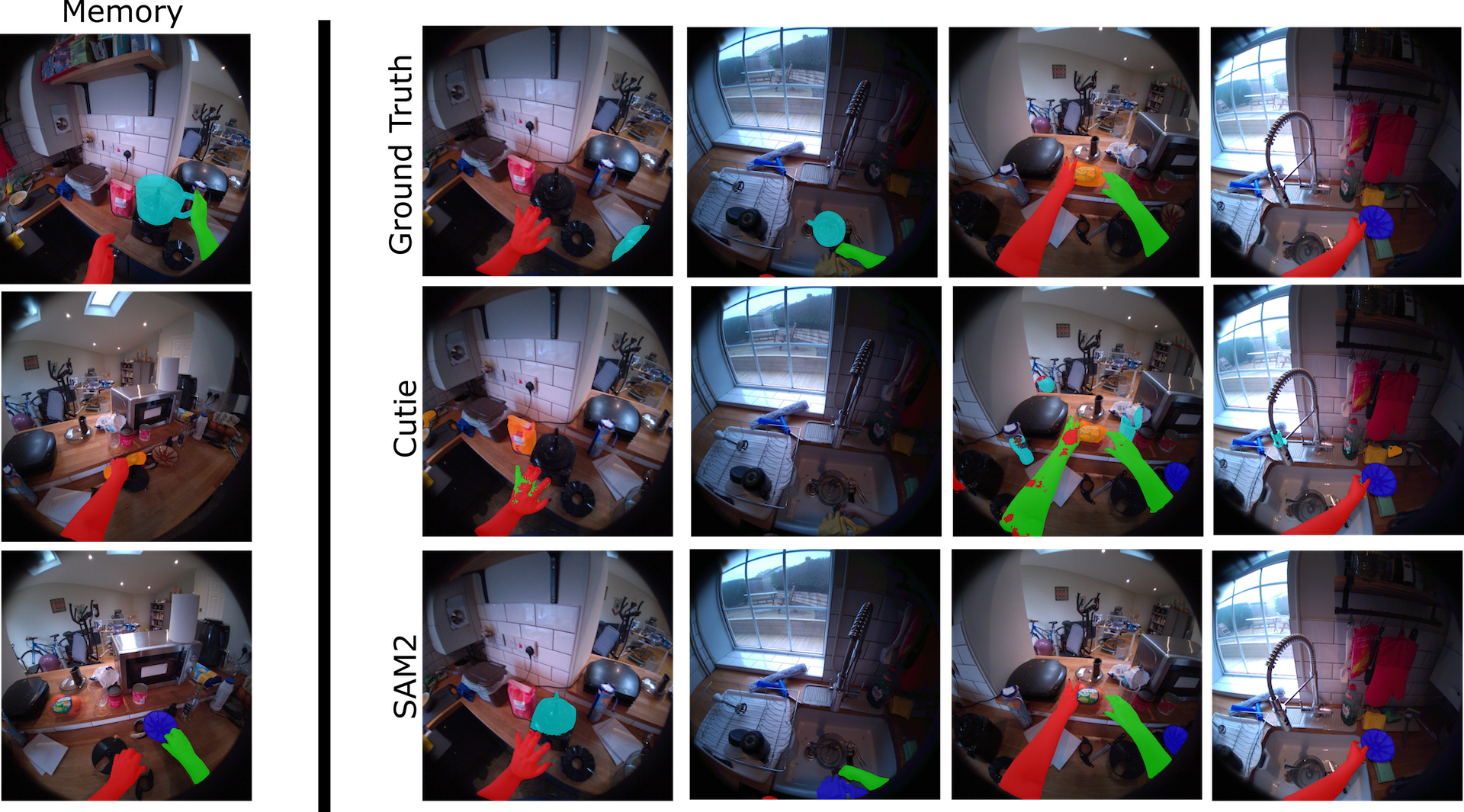}
    \vspace{-0.7em}
    \caption{\textbf{Long-Term VOS}. jaccard index $\mathcal{J}$ \& contour accuracy $\mathcal{F}$ show Cutie and SAM2 struggle with segmenting objects.
    }
    \vspace{-12pt}
    \label{fig:vos_benchmark}
\end{figure}

\vspace{-0.3em}
\subsection{Long-Term VOS Benchmark}
\vspace{-0.2em}
\label{sec:benchmark-long}
We construct a long-term video object segmentation  benchmark using our segmentations and track associations (Sec. \ref{sec:long_term_obj_tracks}). Our benchmark has 1000 sequences, each with 1-5 objects and 2 hand masks (see Supp.~\ref{sec:benchmark_supp} for details). 
While we have a lot more tracks, we keep it comparable to current benchmarks in size.
We evaluate two models~\cite{ravi2024sam,cheng2024putting} with a naive baseline where object masks are kept static. 
Fig.~\ref{fig:vos_benchmark} shows the results. %
SAM2 \cite{ravi2024sam} surpasses Cutie \cite{cheng2024putting} for hands, but does worse on objects. Overall, objects have added challenge in
 diversity in perspective, lighting, location and occlusion.%

\vspace{-0.3em}
\section{Onwards...}
\vspace{-0.2em}
HD-EPIC is available from: \url{http://dx.doi.org/10.5523/bris.3cqb5b81wk2dc2379fx1mrxh47} -- \ie the videos, audio, gaze, blender digital twin, camera pose estimates. 
Annotations are available at: \url{http://hd-epic.github.io} -- \ie object movements, object masks and 3D locations, long-object tracks, and object-action-fixture labels.
We hope \DName will direct future research to a more holistic perception of egocentric videos.

\section*{Acknowledgements}

Research at Bristol is supported by EPSRC Fellowship UMPIRE (EP/T004991/1), EPSRC Program Grant Visual AI (EP/T028572/1) and EPSRC Doctoral Training Program.
O Emara, K Flanagan and F Abdelazim are supported by UKRI CD in Interactive AI (EP/S022937/1). 
The project is also supported by a unrestricted charitable donation from Meta (Aria Project Partnership) to the University of Bristol. 
Gemini Pro results are supported by a research credits grant from Google DeepMind.

Research at Leiden is supported by the Dutch Research Council (NWO) under a Veni grant (VI.Veni.222.160). Research at Singapore is supported by Singapore Ministry of Education (MOE) Academic Research Fund (AcRF)
Tier 1 grant (No. MSS23C018).

We thank Rajan from Elancer and his team, for their huge assistance with temporal and audio annotation.
We thank Srdjan Delic and his team for their assistance with mask annotations.
We thank Damian Steer and the University of Bristol's RDSF team for hosting and maintaining the dataset.
We also thank Owen Tyley for the 3D Digital Twin of the kitchen environments using Blender.

We thank David Fouhey and Evangelos Kazakos for early feedback on the project.
We thank Pierre Moulon, Vijay Baiyya and Cheng Peng from the Aria team for technical assistance in using the MPS code and services.

We acknowledge the usage of EPSRC Tier-2 Jade clusters.
The authors also acknowledge the use of Isambard-AI National AI Research Resource (AIRR). Isambard-AI is operated by the University of Bristol and is funded by the UK Government’s Department for Science, Innovation and Technology (DSIT) via UK Research and Innovation; and the Science and Technology Facilities Council [ST/AIRR/I-A-I/1023].
{
    \small
    \bibliographystyle{ieeenat_fullname}
    \bibliography{main}

\begin{thebibliography}{93}
\providecommand{\natexlab}[1]{#1}
\providecommand{\url}[1]{\texttt{#1}}
\expandafter\ifx\csname urlstyle\endcsname\relax
  \providecommand{\doi}[1]{doi: #1}\else
  \providecommand{\doi}{doi: \begingroup \urlstyle{rm}\Url}\fi

\bibitem[ari()]{ariamps}
Project aria machine perception services.
\newblock
  \url{https://facebookresearch.github.io/projectaria\_tools/docs/ARK/mps}.

\bibitem[ble()]{blender}
\emph{Blender}.
\newblock \url{https://www.blender.org/}.

\bibitem[myf()]{myfitness}
My fitness app.
\newblock \url{https://www.myfitnesspal.com/}.

\bibitem[dut(2022)]{duta20lisa}
\emph{VGG List Annotator (LISA)}, 2022.
\newblock https://www.robots.ox.ac.uk/~vgg/software/lisa/.

\bibitem[Achiam et~al.(2023)Achiam, Adler, Agarwal, Ahmad, Akkaya, Aleman,
  Almeida, Altenschmidt, Altman, Anadkat, et~al.]{achiam2023gpt}
Josh Achiam, Steven Adler, Sandhini Agarwal, Lama Ahmad, Ilge Akkaya,
  Florencia~Leoni Aleman, Diogo Almeida, Janko Altenschmidt, Sam Altman,
  Shyamal Anadkat, et~al.
\newblock {GPT-4} technical report.
\newblock \emph{arXiv preprint arXiv:2303.08774}, 2023.

\bibitem[Alayrac et~al.(2022)Alayrac, Donahue, Luc, Miech, Barr, Hasson, Lenc,
  Mensch, Millican, Reynolds, et~al.]{alayrac2022flamingo}
Jean-Baptiste Alayrac, Jeff Donahue, Pauline Luc, Antoine Miech, Iain Barr,
  Yana Hasson, Karel Lenc, Arthur Mensch, Katherine Millican, Malcolm Reynolds,
  et~al.
\newblock Flamingo: a visual language model for few-shot learning.
\newblock \emph{Advances in Neural Information Processing Systems (NeurIPS)},
  2022.

\bibitem[Banerjee et~al.(2025)Banerjee, Shkodrani, Moulon, Hampali, Zhang,
  Fountain, Miller, Basol, Newcombe, Wang, Engel, and Hodan]{banerjee2024hot3d}
Prithviraj Banerjee, Sindi Shkodrani, Pierre Moulon, Shreyas Hampali, Fan
  Zhang, Jade Fountain, Edward Miller, Selen Basol, Richard Newcombe, Robert
  Wang, Jakob~Julian Engel, and Tomas Hodan.
\newblock {HOT3D: Hand and Object Tracking in 3D from Egocentric Multi-View
  Videos}.
\newblock In \emph{Proceedings of the IEEE/CVF Conference on Computer Vision
  and Pattern Recognition (CVPR)}, 2025.

\bibitem[Cai et~al.(2024)Cai, Tan, Zhang, Zou, Zhang, Yao, Zhu, Gu, Zhong,
  Shang, Dou, Park, Gao, Lee, and Yang]{cai2024temporalbench}
Mu Cai, Reuben Tan, Jianrui Zhang, Bocheng Zou, Kai Zhang, Feng Yao, Fangrui
  Zhu, Jing Gu, Yiwu Zhong, Yuzhang Shang, Yao Dou, Jaden Park, Jianfeng Gao,
  Yong~Jae Lee, and Jianwei Yang.
\newblock {TemporalBench}: Towards fine-grained temporal understanding for
  multimodal video models.
\newblock \emph{arXiv preprint arXiv:2410.10818}, 2024.

\bibitem[Chalk et~al.(2024)Chalk, Huh, Kazakos, Zisserman, and
  Damen]{Chalk_2024_CVPR}
Jacob Chalk, Jaesung Huh, Evangelos Kazakos, Andrew Zisserman, and Dima Damen.
\newblock {TIM}: {A} {T}ime {I}nterval {M}achine for {A}udio-{V}isual {A}ction
  {R}ecognition.
\newblock In \emph{Proceedings of the IEEE/CVF Conference on Computer Vision
  and Pattern Recognition (CVPR)}, 2024.

\bibitem[Chen et~al.(2024{\natexlab{a}})Chen, Liao, Lin, Yu, Chen, and
  Wang]{chen2024rextime}
Jr-Jen Chen, Yu-Chien Liao, Hsi-Che Lin, Yu-Chu Yu, Yen-Chun Chen, and
  Yu-Chiang~Frank Wang.
\newblock {ReXTime}: A benchmark suite for reasoning-across-time in videos.
\newblock In \emph{Advances in Neural Information Processing Systems
  (NeurIPS)}, 2024{\natexlab{a}}.

\bibitem[Chen et~al.(2024{\natexlab{b}})Chen, Wu, Wang, Su, Chen, Xing, Zhong,
  Zhang, Zhu, Lu, et~al.]{chen2024internvl}
Zhe Chen, Jiannan Wu, Wenhai Wang, Weijie Su, Guo Chen, Sen Xing, Muyan Zhong,
  Qinglong Zhang, Xizhou Zhu, Lewei Lu, et~al.
\newblock {InternVL}: Scaling up vision foundation models and aligning for
  generic visual-linguistic tasks.
\newblock In \emph{Proceedings of the IEEE/CVF Conference on Computer Vision
  and Pattern Recognition (CVPR)}, 2024{\natexlab{b}}.

\bibitem[Cheng et~al.(2024{\natexlab{a}})Cheng, Oh, Price, Lee, and
  Schwing]{cheng2024putting}
Ho~Kei Cheng, Seoung~Wug Oh, Brian Price, Joon-Young Lee, and Alexander
  Schwing.
\newblock Putting the object back into video object segmentation.
\newblock In \emph{Proceedings of the IEEE/CVF Conference on Computer Vision
  and Pattern Recognition (CVPR)}, 2024{\natexlab{a}}.

\bibitem[Cheng et~al.(2024{\natexlab{b}})Cheng, Leng, Zhang, Xin, Li, Chen,
  Zhu, Zhang, Luo, Zhao, et~al.]{cheng2024videollama}
Zesen Cheng, Sicong Leng, Hang Zhang, Yifei Xin, Xin Li, Guanzheng Chen,
  Yongxin Zhu, Wenqi Zhang, Ziyang Luo, Deli Zhao, et~al.
\newblock {VideoLLaMA 2: Advancing Spatial-Temporal Modeling and Audio
  Understanding in Video-LLMs}.
\newblock \emph{arXiv preprint arXiv:2406.07476}, 2024{\natexlab{b}}.

\bibitem[Chu et~al.(2023)Chu, Xu, Zhou, Yang, Zhang, Yan, Zhou, and
  Zhou]{chu2023qwen}
Yunfei Chu, Jin Xu, Xiaohuan Zhou, Qian Yang, Shiliang Zhang, Zhijie Yan, Chang
  Zhou, and Jingren Zhou.
\newblock {Qwen-Audio}: Advancing universal audio understanding via unified
  large-scale audio-language models.
\newblock \emph{arXiv preprint arXiv:2311.07919}, 2023.

\bibitem[Cores et~al.(2024)Cores, Dorkenwald, Mucientes, Snoek, and
  Asano]{cores2024tvbench}
Daniel Cores, Michael Dorkenwald, Manuel Mucientes, Cees~GM Snoek, and Yuki~M
  Asano.
\newblock {TVBench}: Redesigning video-language evaluation.
\newblock \emph{arXiv preprint arXiv:2410.07752}, 2024.

\bibitem[Dai et~al.(2017)Dai, Chang, Savva, Halber, Funkhouser, and
  Nie{\ss}ner]{dai2017scannet}
Angela Dai, Angel~X Chang, Manolis Savva, Maciej Halber, Thomas Funkhouser, and
  Matthias Nie{\ss}ner.
\newblock {ScanNet}: Richly-annotated {3D} reconstructions of indoor scenes.
\newblock In \emph{Proceedings of the IEEE/CVF Conference on Computer Vision
  and Pattern Recognition (CVPR)}, 2017.

\bibitem[Damen et~al.(2018)Damen, Doughty, Farinella, Fidler, Furnari, Kazakos,
  Moltisanti, Munro, Perrett, Price, et~al.]{damen2018scaling}
Dima Damen, Hazel Doughty, Giovanni~Maria Farinella, Sanja Fidler, Antonino
  Furnari, Evangelos Kazakos, Davide Moltisanti, Jonathan Munro, Toby Perrett,
  Will Price, et~al.
\newblock Scaling egocentric vision: The {EPIC-KITCHENS} dataset.
\newblock In \emph{Proceedings of the European Conference on Computer Vision
  (ECCV)}, 2018.

\bibitem[Damen et~al.(2022)Damen, Doughty, Farinella, Furnari, Ma, Kazakos,
  Moltisanti, Munro, Perrett, Price, and Wray]{Damen2022RESCALING}
Dima Damen, Hazel Doughty, Giovanni~Maria Farinella, Antonino Furnari, Jian Ma,
  Evangelos Kazakos, Davide Moltisanti, Jonathan Munro, Toby Perrett, Will
  Price, and Michael Wray.
\newblock Rescaling egocentric vision: Collection, pipeline and challenges for
  {EPIC-KITCHENS-100}.
\newblock \emph{International Journal of Computer Vision (IJCV)}, 2022.

\bibitem[Darkhalil et~al.(2022)Darkhalil, Shan, Zhu, Ma, Kar, Higgins, Fidler,
  Fouhey, and Damen]{VISOR2022}
Ahmad Darkhalil, Dandan Shan, Bin Zhu, Jian Ma, Amlan Kar, Richard Higgins,
  Sanja Fidler, David Fouhey, and Dima Damen.
\newblock {EPIC-KITCHENS VISOR} benchmark: Video segmentations and object
  relations.
\newblock In \emph{Proceedings of the Neural Information Processing Systems
  (NeurIPS) Track on Datasets and Benchmarks}, 2022.

\bibitem[De~la Torre et~al.(2009)De~la Torre, Hodgins, Bargteil, Martin, Macey,
  Collado, and Beltran]{de2009guide}
Fernando De~la Torre, Jessica Hodgins, Adam Bargteil, Xavier Martin, Justin
  Macey, Alex Collado, and Pep Beltran.
\newblock {Guide to the Carnegie Mellon University Multimodal activity
  (CMU-MMAC) database}.
\newblock 2009.

\bibitem[Dubey et~al.(2024)Dubey, Jauhri, Pandey, Kadian, Al-Dahle, Letman,
  Mathur, Schelten, Yang, Fan, et~al.]{dubey2024llama}
Abhimanyu Dubey, Abhinav Jauhri, Abhinav Pandey, Abhishek Kadian, Ahmad
  Al-Dahle, Aiesha Letman, Akhil Mathur, Alan Schelten, Amy Yang, Angela Fan,
  et~al.
\newblock {The Llama 3 Herd of Models}.
\newblock \emph{arXiv preprint arXiv:2407.21783}, 2024.

\bibitem[Dutta and Zisserman(2019)]{dutta2019vgg}
Abhishek Dutta and Andrew Zisserman.
\newblock The {VIA} annotation software for images, audio and video.
\newblock In \emph{Proceedings of the 27th ACM International Conference on
  Multimedia (ACMMM)}, 2019.

\bibitem[Fang et~al.(2024)Fang, Mao, Duan, Zhao, Li, Lin, and
  Chen]{fang2024mmbench}
Xinyu Fang, Kangrui Mao, Haodong Duan, Xiangyu Zhao, Yining Li, Dahua Lin, and
  Kai Chen.
\newblock {MMBench-Video}: A long-form multi-shot benchmark for holistic video
  understanding.
\newblock In \emph{Advances in Neural Information Processing Systems
  (NeurIPS)}, 2024.

\bibitem[Feichtenhofer et~al.(2019)Feichtenhofer, Fan, Malik, and
  He]{Feichtenhofer_2019_ICCV}
Christoph Feichtenhofer, Haoqi Fan, Jitendra Malik, and Kaiming He.
\newblock {SlowFast} networks for video recognition.
\newblock In \emph{Proceedings of the IEEE/CVF International Conference on
  Computer Vision (ICCV)}, 2019.

\bibitem[Fu et~al.(2025)Fu, Dai, Luo, Li, Ren, Zhang, Wang, Zhou, Shen, Zhang,
  et~al.]{Video-MME}
Chaoyou Fu, Yuhan Dai, Yondong Luo, Lei Li, Shuhuai Ren, Renrui Zhang, Zihan
  Wang, Chenyu Zhou, Yunhang Shen, Mengdan Zhang, et~al.
\newblock {Video-MME: The First-Ever Comprehensive Evaluation Benchmark of
  Multi-modal LLMs in Video Analysis}.
\newblock In \emph{Proceedings of the IEEE/CVF Conference on Computer Vision
  and Pattern Recognition (CVPR)}, 2025.

\bibitem[Girdhar et~al.(2022)Girdhar, Singh, Ravi, van~der Maaten, Joulin, and
  Misra]{girdhar2022omnivore}
Rohit Girdhar, Mannat Singh, Nikhila Ravi, Laurens van~der Maaten, Armand
  Joulin, and Ishan Misra.
\newblock {Omnivore: A Single Model for Many Visual Modalities}.
\newblock In \emph{Proceedings of the IEEE/CVF Conference on Computer Vision
  and Pattern Recognition (CVPR)}, 2022.

\bibitem[Gong et~al.(2022)Gong, Lai, Chung, and Glass]{gong2022ssast}
Yuan Gong, Cheng-I Lai, Yu-An Chung, and James Glass.
\newblock {SSAST: Self-Supervised Audio Spectrogram Transformer}.
\newblock In \emph{Proceedings of the AAAI Conference on Artificial
  Intelligence}, 2022.

\bibitem[Grauman et~al.(2022)Grauman, Westbury, Byrne, Chavis, Furnari,
  Girdhar, Hamburger, Jiang, Liu, Liu, Martin, Nagarajan, Radosavovic,
  Ramakrishnan, Ryan, Sharma, Wray, Xu, Xu, Zhao, Bansal, Batra, Cartillier,
  Crane, Do, Doulaty, Erapalli, Feichtenhofer, Fragomeni, Fu, Fuegen,
  Gebreselasie, Gonzalez, Hillis, Huang, Huang, Jia, Khoo, Kolar, Kottur,
  Kumar, Landini, Li, Li, Li, Mangalam, Modhugu, Munro, Murrell, Nishiyasu,
  Price, Puentes, Ramazanova, Sari, Somasundaram, Southerland, Sugano, Tao, Vo,
  Wang, Wu, Yagi, Zhu, Arbelaez, Crandall, Damen, Farinella, Ghanem, Ithapu,
  Jawahar, Joo, Kitani, Li, Newcombe, Oliva, Park, Rehg, Sato, Shi, Shou,
  Torralba, Torresani, Yan, and Malik]{Ego4D2022CVPR}
Kristen Grauman, Andrew Westbury, Eugene Byrne, Zachary Chavis, Antonino
  Furnari, Rohit Girdhar, Jackson Hamburger, Hao Jiang, Miao Liu, Xingyu Liu,
  Miguel Martin, Tushar Nagarajan, Ilija Radosavovic, Santhosh~Kumar
  Ramakrishnan, Fiona Ryan, Jayant Sharma, Michael Wray, Mengmeng Xu,
  Eric~Zhongcong Xu, Chen Zhao, Siddhant Bansal, Dhruv Batra, Vincent
  Cartillier, Sean Crane, Tien Do, Morrie Doulaty, Akshay Erapalli, Christoph
  Feichtenhofer, Adriano Fragomeni, Qichen Fu, Christian Fuegen, Abrham
  Gebreselasie, Cristina Gonzalez, James Hillis, Xuhua Huang, Yifei Huang,
  Wenqi Jia, Weslie Khoo, Jachym Kolar, Satwik Kottur, Anurag Kumar, Federico
  Landini, Chao Li, Yanghao Li, Zhenqiang Li, Karttikeya Mangalam, Raghava
  Modhugu, Jonathan Munro, Tullie Murrell, Takumi Nishiyasu, Will Price,
  Paola~Ruiz Puentes, Merey Ramazanova, Leda Sari, Kiran Somasundaram, Audrey
  Southerland, Yusuke Sugano, Ruijie Tao, Minh Vo, Yuchen Wang, Xindi Wu,
  Takuma Yagi, Yunyi Zhu, Pablo Arbelaez, David Crandall, Dima Damen,
  Giovanni~Maria Farinella, Bernard Ghanem, Vamsi~Krishna Ithapu, C.~V.
  Jawahar, Hanbyul Joo, Kris Kitani, Haizhou Li, Richard Newcombe, Aude Oliva,
  Hyun~Soo Park, James~M. Rehg, Yoichi Sato, Jianbo Shi, Mike~Zheng Shou,
  Antonio Torralba, Lorenzo Torresani, Mingfei Yan, and Jitendra Malik.
\newblock {Ego4D}: Around the {W}orld in 3,000 {H}ours of {E}gocentric {V}ideo.
\newblock In \emph{IEEE/CVF Computer Vision and Pattern Recognition (CVPR)},
  2022.

\bibitem[Grauman et~al.(2024)Grauman, Westbury, Torresani, Kitani, Malik,
  Afouras, Ashutosh, Baiyya, Bansal, Boote, Byrne, Chavis, Chen, Cheng, Chu,
  Crane, Dasgupta, Dong, Escobar, Forigua, Gebreselasie, Haresh, Huang, Islam,
  Jain, Khirodkar, Kukreja, Liang, Liu, Majumder, Mao, Martin, Mavroudi,
  Nagarajan, Ragusa, Ramakrishnan, Seminara, Somayazulu, Song, Su, Xue, Zhang,
  Zhang, Castillo, Chen, Fu, Furuta, Gonzalez, Gupta, Hu, Huang, Huang, Khoo,
  Kumar, Kuo, Lakhavani, Liu, Luo, Luo, Meredith, Miller, Oguntola, Pan, Peng,
  Pramanick, Ramazanova, Ryan, Shan, Somasundaram, Song, Southerland, Tateno,
  Wang, Wang, Yagi, Yan, Yang, Yu, Zha, Zhao, Zhao, Zhu, Zhuo, Arbelaez,
  Bertasius, Damen, Engel, Farinella, Furnari, Ghanem, Hoffman, Jawahar,
  Newcombe, Park, Rehg, Sato, Savva, Shi, Shou, and Wray]{EgoExo4d}
Kristen Grauman, Andrew Westbury, Lorenzo Torresani, Kris Kitani, Jitendra
  Malik, Triantafyllos Afouras, Kumar Ashutosh, Vijay Baiyya, Siddhant Bansal,
  Bikram Boote, Eugene Byrne, Zach Chavis, Joya Chen, Feng Cheng, Fu-Jen Chu,
  Sean Crane, Avijit Dasgupta, Jing Dong, Maria Escobar, Cristhian Forigua,
  Abrham Gebreselasie, Sanjay Haresh, Jing Huang, Md~Mohaiminul Islam, Suyog
  Jain, Rawal Khirodkar, Devansh Kukreja, Kevin~J Liang, Jia-Wei Liu, Sagnik
  Majumder, Yongsen Mao, Miguel Martin, Effrosyni Mavroudi, Tushar Nagarajan,
  Francesco Ragusa, Santhosh~Kumar Ramakrishnan, Luigi Seminara, Arjun
  Somayazulu, Yale Song, Shan Su, Zihui Xue, Edward Zhang, Jinxu Zhang, Angela
  Castillo, Changan Chen, Xinzhu Fu, Ryosuke Furuta, Cristina Gonzalez, Prince
  Gupta, Jiabo Hu, Yifei Huang, Yiming Huang, Weslie Khoo, Anush Kumar, Robert
  Kuo, Sach Lakhavani, Miao Liu, Mi Luo, Zhengyi Luo, Brighid Meredith, Austin
  Miller, Oluwatumininu Oguntola, Xiaqing Pan, Penny Peng, Shraman Pramanick,
  Merey Ramazanova, Fiona Ryan, Wei Shan, Kiran Somasundaram, Chenan Song,
  Audrey Southerland, Masatoshi Tateno, Huiyu Wang, Yuchen Wang, Takuma Yagi,
  Mingfei Yan, Xitong Yang, Zecheng Yu, Shengxin~Cindy Zha, Chen Zhao, Ziwei
  Zhao, Zhifan Zhu, Jeff Zhuo, Pablo Arbelaez, Gedas Bertasius, Dima Damen,
  Jakob Engel, Giovanni~Maria Farinella, Antonino Furnari, Bernard Ghanem, Judy
  Hoffman, C.V. Jawahar, Richard Newcombe, Hyun~Soo Park, James~M. Rehg, Yoichi
  Sato, Manolis Savva, Jianbo Shi, Mike~Zheng Shou, and Michael Wray.
\newblock {Ego-Exo4D: Understanding Skilled Human Activity from First- and
  Third-Person Perspectives}.
\newblock In \emph{Proceedings of the IEEE/CVF Conference on Computer Vision
  and Pattern Recognition (CVPR)}, 2024.

\bibitem[Honnibal and Montani(2020)]{honnibal2017spacy}
Matthew Honnibal and Ines Montani.
\newblock {spaCy 2: Natural language understanding with Bloom embeddings,
  convolutional neural networks and incremental parsing}.
\newblock 2020.

\bibitem[Huh et~al.(2023)Huh, Chalk, Kazakos, Damen, and
  Zisserman]{EPICSOUNDS2023}
Jaesung Huh, Jacob Chalk, Evangelos Kazakos, Dima Damen, and Andrew Zisserman.
\newblock {EPIC-SOUNDS}: {A} {L}arge-{S}cale {D}ataset of {A}ctions that
  {S}ound.
\newblock In \emph{IEEE International Conference on Acoustics, Speech, \&
  Signal Processing (ICASSP)}, 2023.

\bibitem[Johansson et~al.(2001)Johansson, Westling, Bäckström, and
  Flanagan]{johansson2001eyehand}
Roland Johansson, Göran Westling, Anders Bäckström, and John Flanagan.
\newblock {Eye–Hand Coordination in Object Manipulation}.
\newblock \emph{The Journal of neuroscience: the official journal of the
  Society for Neuroscience}, 2001.

\bibitem[Kar et~al.(2021)Kar, Kim, Boben, Gao, Li, Ling, Wang, and
  Fidler]{torontoannotsuite}
Amlan Kar, Seung~Wook Kim, Marko Boben, Jun Gao, Tianxing Li, Huan Ling, Zian
  Wang, and Sanja Fidler.
\newblock Toronto annotation suite.
\newblock \url{https://aidemos.cs.toronto.edu/toras}, 2021.

\bibitem[Kazakos et~al.(2021)Kazakos, Nagrani, Zisserman, and
  Damen]{Kazakos2021SlowFastAuditory}
Evangelos Kazakos, Arsha Nagrani, Andrew Zisserman, and Dima Damen.
\newblock {Slow-Fast Auditory Streams For Audio Recognition}.
\newblock In \emph{IEEE International Conference on Acoustics, Speech and
  Signal Processing (ICASSP)}, 2021.

\bibitem[Kesen et~al.(2024)Kesen, Pedrotti, Dogan, Cafagna, Acikgoz,
  Parcalabescu, Calixto, Frank, Gatt, Erdem, and Erdem]{kesen2023vilma}
Ilker Kesen, Andrea Pedrotti, Mustafa Dogan, Michele Cafagna, Emre~Can Acikgoz,
  Letitia Parcalabescu, Iacer Calixto, Anette Frank, Albert Gatt, Aykut Erdem,
  and Erkut Erdem.
\newblock {ViLMA: A Zero-Shot Benchmark for Linguistic and Temporal Grounding
  in Video-Language Models}.
\newblock In \emph{International Conference on Learning Representations
  (ICLR)}, 2024.

\bibitem[Kirillov et~al.(2023)Kirillov, Mintun, Ravi, Mao, Rolland, Gustafson,
  Xiao, Whitehead, Berg, Lo, et~al.]{kirillov2023segment}
Alexander Kirillov, Eric Mintun, Nikhila Ravi, Hanzi Mao, Chloe Rolland, Laura
  Gustafson, Tete Xiao, Spencer Whitehead, Alexander~C Berg, Wan-Yen Lo, et~al.
\newblock Segment anything.
\newblock In \emph{Proceedings of the IEEE/CVF International Conference on
  Computer Vision (ICCV)}, 2023.

\bibitem[Kitaev and Klein(2018)]{kitaev2018constituency}
Nikita Kitaev and Dan Klein.
\newblock Constituency parsing with a self-attentive encoder.
\newblock In \emph{Proceedings of the Annual Meeting of the Association for
  Computational Linguistics (ACL)}, 2018.

\bibitem[Kuchaiev et~al.(2019)Kuchaiev, Li, Nguyen, Hrinchuk, Leary, Ginsburg,
  Kriman, Beliaev, Lavrukhin, Cook, Castonguay, Popova, Huang, and
  Cohen]{kuchaiev2019nemo}
Oleksii Kuchaiev, Jason Li, Huyen Nguyen, Oleksii Hrinchuk, Ryan Leary, Boris
  Ginsburg, Samuel Kriman, Stanislav Beliaev, Vitaly Lavrukhin, Jack Cook,
  Patrice Castonguay, Mariya Popova, Jocelyn Huang, and Jonathan~M. Cohen.
\newblock {NeMo: a toolkit for building AI applications using Neural Modules}.
\newblock \emph{arXiv preprint arXiv:1909.09577}, 2019.

\bibitem[Lai et~al.(2023)Lai, Dai, Chen, Pang, Rehg, and Liu]{lai2023lego}
Bolin Lai, Xiaoliang Dai, Lawrence Chen, Guan Pang, James~M Rehg, and Miao Liu.
\newblock {LEGO: Learning EGOcentric Action Frame Generation via Visual
  Instruction Tuning}.
\newblock In \emph{Proceedings of the European Conference on Computer Vision
  (ECCV)}, 2023.

\bibitem[Land et~al.(1999)Land, Mennie, and Rusted]{land1999roles}
Michael Land, Neil Mennie, and Jennifer Rusted.
\newblock The roles of vision and eye movements in the control of activities of
  daily living.
\newblock \emph{Perception}, 1999.

\bibitem[Lee et~al.(2012)Lee, Ghosh, and Grauman]{lee2012discovering}
Yong~Jae Lee, Joydeep Ghosh, and Kristen Grauman.
\newblock Discovering important people and objects for egocentric video
  summarization.
\newblock In \emph{Proceedings of the IEEE/CVF Conference on Computer Vision
  and Pattern Recognition (CVPR)}, 2012.

\bibitem[Li et~al.(2024{\natexlab{a}})Li, Ge, Ge, Wang, Wang, Zhang, and
  Shan]{li2024seed}
Bohao Li, Yuying Ge, Yixiao Ge, Guangzhi Wang, Rui Wang, Ruimao Zhang, and Ying
  Shan.
\newblock {SEED-Bench: Benchmarking Multimodal Large Language Models}.
\newblock In \emph{Proceedings of the IEEE/CVF Conference on Computer Vision
  and Pattern Recognition (CVPR)}, 2024{\natexlab{a}}.

\bibitem[Li et~al.(2024{\natexlab{b}})Li, Wang, He, Li, Wang, Liu, Wang, Xu,
  Chen, Luo, et~al.]{li2024mvbench}
Kunchang Li, Yali Wang, Yinan He, Yizhuo Li, Yi Wang, Yi Liu, Zun Wang, Jilan
  Xu, Guo Chen, Ping Luo, et~al.
\newblock {MVBench: A Comprehensive Multi-modal Video Understanding Benchmark}.
\newblock In \emph{Proceedings of the IEEE/CVF Conference on Computer Vision
  and Pattern Recognition (CVPR)}, 2024{\natexlab{b}}.

\bibitem[Li et~al.(2024{\natexlab{c}})Li, Li, Ren, Liu, Liu, Gao, Sun, and
  Hou]{li2024vitatecs}
Shicheng Li, Lei Li, Shuhuai Ren, Yuanxin Liu, Yi Liu, Rundong Gao, Xu Sun, and
  Lu Hou.
\newblock {VITATECS: A Diagnostic Dataset for Temporal Concept Understanding of
  Video-Language Models}.
\newblock In \emph{Proceedings of the European Conference on Computer Vision
  (ECCV)}, 2024{\natexlab{c}}.

\bibitem[Li et~al.(2021)Li, Liu, and Rehg]{li2021eye}
Yin Li, Miao Liu, and James~M Rehg.
\newblock In the eye of the beholder: Gaze and actions in first person video.
\newblock \emph{IEEE Transactions on Pattern Analysis and Machine Intelligence
  (TPAMI)}, 2021.

\bibitem[Liu et~al.(2022)Liu, Liu, Jiang, Lyu, Wan, Shen, Liang, Fu, Wang, and
  Yi]{liu2022hoi4d}
Yunze Liu, Yun Liu, Che Jiang, Kangbo Lyu, Weikang Wan, Hao Shen, Boqiang
  Liang, Zhoujie Fu, He Wang, and Li Yi.
\newblock {HOI4D: A 4D Egocentric Dataset for Category-Level Human-Object
  Interaction}.
\newblock In \emph{Proceedings of the IEEE/CVF Conference on Computer Vision
  and Pattern Recognition (CVPR)}, 2022.

\bibitem[Lv et~al.(2024)Lv, Charron, Moulon, Gamino, Peng, Sweeney, Miller,
  Tang, Meissner, Dong, et~al.]{lv2024aria}
Zhaoyang Lv, Nickolas Charron, Pierre Moulon, Alexander Gamino, Cheng Peng,
  Chris Sweeney, Edward Miller, Huixuan Tang, Jeff Meissner, Jing Dong, et~al.
\newblock {Aria Everyday Activities Dataset}.
\newblock \emph{arXiv preprint arXiv:2402.13349}, 2024.

\bibitem[Majumdar et~al.(2024)Majumdar, Ajay, Zhang, Putta, Yenamandra, Henaff,
  Silwal, Mcvay, Maksymets, Arnaud, Yadav, Li, Newman, Sharma, Berges, Zhang,
  Agrawal, Bisk, Batra, Kalakrishnan, Meier, Paxton, Sax, and
  Rajeswaran]{OpenEQA2023}
Arjun Majumdar, Anurag Ajay, Xiaohan Zhang, Pranav Putta, Sriram Yenamandra,
  Mikael Henaff, Sneha Silwal, Paul Mcvay, Oleksandr Maksymets, Sergio Arnaud,
  Karmesh Yadav, Qiyang Li, Ben Newman, Mohit Sharma, Vincent Berges, Shiqi
  Zhang, Pulkit Agrawal, Yonatan Bisk, Dhruv Batra, Mrinal Kalakrishnan,
  Franziska Meier, Chris Paxton, Sasha Sax, and Aravind Rajeswaran.
\newblock {OpenEQA: Embodied Question Answering in the Era of Foundation
  Models}.
\newblock In \emph{Proceedings of the IEEE/CVF Conference on Computer Vision
  and Pattern Recognition (CVPR)}, 2024.

\bibitem[Mangalam et~al.(2023)Mangalam, Akshulakov, and
  Malik]{mangalam2023egoschema}
Karttikeya Mangalam, Raiymbek Akshulakov, and Jitendra Malik.
\newblock {EgoSchema: A Diagnostic Benchmark for Very Long-form Video Language
  Understanding}.
\newblock \emph{Advances in Neural Information Processing Systems (NeurIPS)},
  2023.

\bibitem[Pan et~al.(2023)Pan, Charron, Yang, Peters, Whelan, Kong, Parkhi,
  Newcombe, and Ren]{pan2023aria}
Xiaqing Pan, Nicholas Charron, Yongqian Yang, Scott Peters, Thomas Whelan, Chen
  Kong, Omkar Parkhi, Richard Newcombe, and Yuheng~Carl Ren.
\newblock {Aria Digital Twin: A New Benchmark Dataset for Egocentric 3D Machine
  Perception}.
\newblock In \emph{Proceedings of the IEEE/CVF International Conference on
  Computer Vision (ICCV)}, 2023.

\bibitem[Patrick et~al.(2021)Patrick, Campbell, Asano, Metze, Feichtenhofer,
  Vedaldi, and Henriques]{patrick2021keeping}
Mandela Patrick, Dylan Campbell, Yuki~M. Asano, Ishan Misra~Florian Metze,
  Christoph Feichtenhofer, Andrea Vedaldi, and João~F. Henriques.
\newblock Keeping your eye on the ball: Trajectory attention in video
  transformers.
\newblock In \emph{Advances in Neural Information Processing Systems
  (NeurIPS)}, 2021.

\bibitem[Peddi et~al.(2024)Peddi, Arya, Challa, Pallapothula, Vyas, Gouripeddi,
  Wang, Zhang, Komaragiri, Ragan, Ruozzi, Xiang, and Gogate]{captaincook4d}
Rohith Peddi, Shivvrat Arya, Bharath Challa, Likhitha Pallapothula, Akshay
  Vyas, Bhavya Gouripeddi, Jikai Wang, Qifan Zhang, Vasundhara Komaragiri, Eric
  Ragan, Nicholas Ruozzi, Yu Xiang, and Vibhav Gogate.
\newblock {CaptainCook4D: A Dataset for Understanding Errors in Procedural
  Activities}.
\newblock In \emph{Proceedings of the Neural Information Processing Systems
  (NeurIPS) Track on Datasets and Benchmarks}, 2024.

\bibitem[Perazzi et~al.(2016)Perazzi, Pont-Tuset, McWilliams, Van~Gool, Gross,
  and Sorkine-Hornung]{perazzi2016benchmark}
Federico Perazzi, Jordi Pont-Tuset, Brian McWilliams, Luc Van~Gool, Markus
  Gross, and Alexander Sorkine-Hornung.
\newblock A benchmark dataset and evaluation methodology for video object
  segmentation.
\newblock In \emph{Proceedings of the IEEE/CVF Conference on Computer Vision
  and Pattern Recognition (CVPR)}, 2016.

\bibitem[Perrett et~al.(2024)Perrett, Han, Damen, and
  Zisserman]{perrett2024unique}
Toby Perrett, Tengda Han, Dima Damen, and Andrew Zisserman.
\newblock It's just another day: Unique video captioning by discriminitave
  prompting.
\newblock In \emph{Proceedings of the Asian Conference on Computer Vision
  (ACCV)}, 2024.

\bibitem[Pirsiavash and Ramanan(2012)]{pirsiavash2012detecting}
Hamed Pirsiavash and Deva Ramanan.
\newblock Detecting activities of daily living in first-person camera views.
\newblock In \emph{Proceedings of the IEEE/CVF Conference on Computer Vision
  and Pattern Recognition (CVPR)}, 2012.

\bibitem[Plizzari et~al.(2025)Plizzari, Goel, Perrett, Chalk, Kanazawa, and
  Damen]{Plizzari2024}
Chiara Plizzari, Shubham Goel, Toby Perrett, Jacob Chalk, Angjoo Kanazawa, and
  Dima Damen.
\newblock {Spatial Cognition from Egocentric Video: Out of Sight, Not Out of
  Mind}.
\newblock In \emph{{Proceedings of the IEEE International Conference on 3D
  Vision (3DV)}}, 2025.

\bibitem[Pont-Tuset et~al.(2017)Pont-Tuset, Perazzi, Caelles, Arbel\'aez,
  Sorkine-Hornung, and {Van Gool}]{Pont-Tuset_arXiv_2017}
Jordi Pont-Tuset, Federico Perazzi, Sergi Caelles, Pablo Arbel\'aez, Alexander
  Sorkine-Hornung, and Luc {Van Gool}.
\newblock The 2017 davis challenge on video object segmentation.
\newblock \emph{arXiv:1704.00675}, 2017.

\bibitem[Pătrăucean et~al.(2023)Pătrăucean, Smaira, Gupta, Continente,
  Markeeva, Banarse, Koppula, Heyward, Malinowski, Yang, Doersch, Matejovicova,
  Sulsky, Miech, Frechette, Klimczak, Koster, Zhang, Winkler, Aytar, Osindero,
  Damen, Zisserman, and Carreira]{patraucean2023perception}
Viorica Pătrăucean, Lucas Smaira, Ankush Gupta, Adrià~Recasens Continente,
  Larisa Markeeva, Dylan Banarse, Skanda Koppula, Joseph Heyward, Mateusz
  Malinowski, Yi Yang, Carl Doersch, Tatiana Matejovicova, Yury Sulsky, Antoine
  Miech, Alex Frechette, Hanna Klimczak, Raphael Koster, Junlin Zhang,
  Stephanie Winkler, Yusuf Aytar, Simon Osindero, Dima Damen, Andrew Zisserman,
  and João Carreira.
\newblock Perception test: A diagnostic benchmark for multimodal video models.
\newblock In \emph{Advances in Neural Information Processing Systems
  (NeurIPS)}, 2023.

\bibitem[Radford et~al.(2021)Radford, Kim, Hallacy, Ramesh, Goh, Agarwal,
  Sastry, Askell, Mishkin, Clark, et~al.]{radford2021learning}
Alec Radford, Jong~Wook Kim, Chris Hallacy, Aditya Ramesh, Gabriel Goh,
  Sandhini Agarwal, Girish Sastry, Amanda Askell, Pamela Mishkin, Jack Clark,
  et~al.
\newblock Learning transferable visual models from natural language
  supervision.
\newblock In \emph{International Conference on Machine Learning (ICML)}, 2021.

\bibitem[Radford et~al.(2023)Radford, Kim, Xu, Brockman, McLeavey, and
  Sutskever]{radford2023robust}
Alec Radford, Jong~Wook Kim, Tao Xu, Greg Brockman, Christine McLeavey, and
  Ilya Sutskever.
\newblock Robust speech recognition via large-scale weak supervision.
\newblock In \emph{International Conference on Machine Learning (ICML)}, 2023.

\bibitem[Ragusa et~al.(2020)Ragusa, Furnari, Battiato, Signorello, and
  Farinella]{ragusa2020ego}
Francesco Ragusa, Antonino Furnari, Sebastiano Battiato, Giovanni Signorello,
  and Giovanni~Maria Farinella.
\newblock {EGO-CH: Dataset and fundamental tasks for visitors behavioral
  understanding using egocentric vision}.
\newblock \emph{Pattern Recognition Letters}, 2020.

\bibitem[Ragusa et~al.(2021)Ragusa, Furnari, Livatino, and
  Farinella]{ragusa2021meccano}
Francesco Ragusa, Antonino Furnari, Salvatore Livatino, and Giovanni~Maria
  Farinella.
\newblock {The MECCANO Dataset: Understanding Human-Object Interactions from
  Egocentric Videos in an Industrial-like Domain}.
\newblock In \emph{Proceedings of the IEEE/CVF Winter Conference on
  Applications of Computer Vision (WACV)}, 2021.

\bibitem[Ramakrishnan et~al.(2021)Ramakrishnan, Gokaslan, Wijmans, Maksymets,
  Clegg, Turner, Undersander, Galuba, Westbury, Chang,
  et~al.]{ramakrishnan2021habitat}
Santhosh~K Ramakrishnan, Aaron Gokaslan, Erik Wijmans, Oleksandr Maksymets,
  Alex Clegg, John Turner, Eric Undersander, Wojciech Galuba, Andrew Westbury,
  Angel~X Chang, et~al.
\newblock {Habitat-Matterport 3D Dataset (HM3D): 1000 Large-scale 3D
  Environments for Embodied AI}.
\newblock In \emph{Advances in Neural Information Processing Systems (NeurIPS)
  Datasets and Benchmarks Track}, 2021.

\bibitem[Ravi et~al.(2024)Ravi, Gabeur, Hu, Hu, Ryali, Ma, Khedr, R{\"a}dle,
  Rolland, Gustafson, et~al.]{ravi2024sam}
Nikhila Ravi, Valentin Gabeur, Yuan-Ting Hu, Ronghang Hu, Chaitanya Ryali,
  Tengyu Ma, Haitham Khedr, Roman R{\"a}dle, Chloe Rolland, Laura Gustafson,
  et~al.
\newblock {SAM} 2: Segment anything in images and videos.
\newblock \emph{arXiv preprint arXiv:2408.00714}, 2024.

\bibitem[Sch\"{o}nberger and Frahm(2016)]{schoenberger2016sfm}
Johannes~Lutz Sch\"{o}nberger and Jan-Michael Frahm.
\newblock Structure-from-motion revisited.
\newblock In \emph{Proceedings of the IEEE/CVF Conference on Computer Vision
  and Pattern Recognition (CVPR)}, 2016.

\bibitem[Sch\"{o}nberger et~al.(2016)Sch\"{o}nberger, Zheng, Pollefeys, and
  Frahm]{schoenberger2016mvs}
Johannes~Lutz Sch\"{o}nberger, Enliang Zheng, Marc Pollefeys, and Jan-Michael
  Frahm.
\newblock Pixelwise view selection for unstructured multi-view stereo.
\newblock In \emph{Proceedings of the European Conference on Computer Vision
  (ECCV)}, 2016.

\bibitem[Schoonbeek et~al.(2024)Schoonbeek, Houben, Onvlee, van~der Sommen,
  et~al.]{schoonbeek2024industreal}
Tim~J Schoonbeek, Tim Houben, Hans Onvlee, Fons van~der Sommen, et~al.
\newblock {IndustReal: A Dataset for Procedure Step Recognition Handling
  Execution Errors in Egocentric Videos in an Industrial-Like Setting}.
\newblock In \emph{Proceedings of the IEEE/CVF Winter Conference on
  Applications of Computer Vision (WACV)}, 2024.

\bibitem[Sener et~al.(2022)Sener, Chatterjee, Shelepov, He, Singhania, Wang,
  and Yao]{sener2022assembly101}
Fadime Sener, Dibyadip Chatterjee, Daniel Shelepov, Kun He, Dipika Singhania,
  Robert Wang, and Angela Yao.
\newblock Assembly101: A large-scale multi-view video dataset for understanding
  procedural activities.
\newblock In \emph{Proceedings of the IEEE/CVF Conference on Computer Vision
  and Pattern Recognition (CVPR)}, 2022.

\bibitem[Sigurdsson et~al.(2018)Sigurdsson, Gupta, Schmid, Farhadi, and
  Alahari]{sigurdsson2018actor}
Gunnar~A Sigurdsson, Abhinav Gupta, Cordelia Schmid, Ali Farhadi, and Karteek
  Alahari.
\newblock Actor and observer: Joint modeling of first and third-person videos.
\newblock In \emph{Proceedings of the IEEE/CVF Conference on Computer Vision
  and Pattern Recognition (CVPR)}, 2018.

\bibitem[Singh et~al.(2022)Singh, Hu, Goswami, Couairon, Galuba, Rohrbach, and
  Kiela]{singh2022flava}
Amanpreet Singh, Ronghang Hu, Vedanuj Goswami, Guillaume Couairon, Wojciech
  Galuba, Marcus Rohrbach, and Douwe Kiela.
\newblock {FLAVA: A foundational language and vision alignment model}.
\newblock In \emph{Proceedings of the IEEE/CVF Conference on Computer Vision
  and Pattern Recognition (CVPR)}, 2022.

\bibitem[Singh et~al.(2016)Singh, Fatahalian, and Efros]{singh2016krishnacam}
Krishna~Kumar Singh, Kayvon Fatahalian, and Alexei~A Efros.
\newblock Krishnacam: Using a longitudinal, single-person, egocentric dataset
  for scene understanding tasks.
\newblock In \emph{Proceedings of the IEEE Winter Conference on Applications of
  Computer Vision (WACV)}, 2016.

\bibitem[Somasundaram et~al.(2023)Somasundaram, Dong, Tang, Straub, Yan,
  Goesele, Engel, Nardi, and Newcombe]{Somasundaram2023ProjectAA}
Kiran~K. Somasundaram, Jing Dong, Huixuan Tang, Julian Straub, Mingfei Yan,
  Michael Goesele, Jakob~J. Engel, Renzo~De Nardi, and Richard~A. Newcombe.
\newblock {Project Aria: A New Tool for Egocentric Multi-Modal AI Research}.
\newblock \emph{arXiv preprint arXiv:2308.13561}, 2023.

\bibitem[Song et~al.(2024)Song, Byrne, Nagarajan, Wang, Martin, and
  Torresani]{song2024ego4d}
Yale Song, Eugene Byrne, Tushar Nagarajan, Huiyu Wang, Miguel Martin, and
  Lorenzo Torresani.
\newblock {Ego4D Goal-Step: Toward Hierarchical Understanding of Procedural
  Activities}.
\newblock \emph{Advances in Neural Information Processing Systems (NeurIPS)},
  2024.

\bibitem[Straub et~al.(2024)Straub, DeTone, Shen, Yang, Sweeney, and
  Newcombe]{straub2024efm3d}
Julian Straub, Daniel DeTone, Tianwei Shen, Nan Yang, Chris Sweeney, and
  Richard Newcombe.
\newblock {EFM3D: A Benchmark for Measuring Progress Towards 3D Egocentric
  Foundation Models}.
\newblock \emph{arXiv preprint arXiv:2406.10224}, 2024.

\bibitem[Team et~al.(2024)Team, Georgiev, Lei, Burnell, Bai, Gulati, Tanzer,
  Vincent, Pan, Wang, et~al.]{team2024gemini}
Gemini Team, Petko Georgiev, Ving~Ian Lei, Ryan Burnell, Libin Bai, Anmol
  Gulati, Garrett Tanzer, Damien Vincent, Zhufeng Pan, Shibo Wang, et~al.
\newblock Gemini 1.5: Unlocking multimodal understanding across millions of
  tokens of context.
\newblock \emph{arXiv preprint arXiv:2403.05530}, 2024.

\bibitem[Tong et~al.(2022)Tong, Song, Wang, and Wang]{tong2022videomae}
Zhan Tong, Yibing Song, Jue Wang, and Limin Wang.
\newblock Video{MAE}: Masked autoencoders are data-efficient learners for
  self-supervised video pre-training.
\newblock In \emph{Advances in Neural Information Processing Systems
  (NeurIPS)}, 2022.

\bibitem[Tschernezki et~al.(2023)Tschernezki, Darkhalil, Zhu, Fouhey, Larina,
  Larlus, Damen, and Vedaldi]{EPICFields2023}
Vadim Tschernezki, Ahmad Darkhalil, Zhifan Zhu, David Fouhey, Iro Larina, Diane
  Larlus, Dima Damen, and Andrea Vedaldi.
\newblock {EPIC Fields}: {M}arrying {3D} {G}eometry and {V}ideo
  {U}nderstanding.
\newblock In \emph{Advances in Neural Information Processing Systems
  (NeurIPS)}, 2023.

\bibitem[Wang et~al.(2024)Wang, Wu, Chen, Chen, Guan, Lee, Li, and
  Gan]{wang2024sok}
Andong Wang, Bo Wu, Sunli Chen, Zhenfang Chen, Haotian Guan, Wei-Ning Lee,
  Li~Erran Li, and Chuang Gan.
\newblock {SOK-Bench: A Situated Video Reasoning Benchmark with Aligned
  Open-World Knowledge}.
\newblock In \emph{Proceedings of the IEEE/CVF Conference on Computer Vision
  and Pattern Recognition (CVPR)}, 2024.

\bibitem[Wang et~al.(2023{\natexlab{a}})Wang, Dai, Chen, Huang, Li, Zhu, Hu,
  Lu, Lu, Li, et~al.]{wang2023internimage}
Wenhai Wang, Jifeng Dai, Zhe Chen, Zhenhang Huang, Zhiqi Li, Xizhou Zhu,
  Xiaowei Hu, Tong Lu, Lewei Lu, Hongsheng Li, et~al.
\newblock {InternImage: Exploring Large-Scale Vision Foundation Models with
  Deformable Convolutions}.
\newblock In \emph{Proceedings of the IEEE/CVF Conference on Computer Vision
  and Pattern Recognition (CVPR)}, 2023{\natexlab{a}}.

\bibitem[Wang et~al.(2023{\natexlab{b}})Wang, Kwon, Rad, Pan, Chakraborty,
  Andrist, Bohus, Feniello, Tekin, Frujeri, Joshi, and
  Pollefeys]{HoloAssist2023}
Xin Wang, Taein Kwon, Mahdi Rad, Bowen Pan, Ishani Chakraborty, Sean Andrist,
  Dan Bohus, Ashley Feniello, Bugra Tekin, Felipe~Vieira Frujeri, Neel Joshi,
  and Marc Pollefeys.
\newblock {HoloAssist: an Egocentric Human Interaction Dataset for Interactive
  AI Assistants in the Real World}.
\newblock In \emph{Proceedings of the IEEE/CVF International Conference on
  Computer Vision (ICCV)}, 2023{\natexlab{b}}.

\bibitem[Xiao et~al.(2024{\natexlab{a}})Xiao, Wu, Xu, Dai, Hu, Lu, Zeng, Liu,
  and Yuan]{xiao2024florence}
Bin Xiao, Haiping Wu, Weijian Xu, Xiyang Dai, Houdong Hu, Yumao Lu, Michael
  Zeng, Ce Liu, and Lu Yuan.
\newblock Florence-2: Advancing a unified representation for a variety of
  vision tasks.
\newblock In \emph{Proceedings of the IEEE/CVF Conference on Computer Vision
  and Pattern Recognition (CVPR)}, 2024{\natexlab{a}}.

\bibitem[Xiao et~al.(2024{\natexlab{b}})Xiao, Yao, Li, and Chua]{xiao2024can}
Junbin Xiao, Angela Yao, Yicong Li, and Tat-Seng Chua.
\newblock {Can I trust your answer? Visually grounded video question
  answering}.
\newblock In \emph{Proceedings of the IEEE/CVF Conference on Computer Vision
  and Pattern Recognition (CVPR)}, 2024{\natexlab{b}}.

\bibitem[Xu et~al.(2018)Xu, Yang, Fan, Yue, Liang, Yang, and
  Huang]{xu2018youtube}
Ning Xu, Linjie Yang, Yuchen Fan, Dingcheng Yue, Yuchen Liang, Jianchao Yang,
  and Thomas Huang.
\newblock You{T}ube-{VOS}: A large-scale video object segmentation benchmark.
\newblock \emph{arXiv preprint arXiv:1809.03327}, 2018.

\bibitem[Yang et~al.(2019)Yang, Fan, and Xu]{vos2019}
Linjie Yang, Yuchen Fan, and Ning Xu.
\newblock The 2nd large-scale video object segmentation challenge - video
  object segmentation track, 2019.

\bibitem[Yang et~al.(2024)Yang, Kang, Huang, Xu, Feng, and Zhao]{depthanything}
Lihe Yang, Bingyi Kang, Zilong Huang, Xiaogang Xu, Jiashi Feng, and Hengshuang
  Zhao.
\newblock {Depth Anything: Unleashing the Power of Large-Scale Unlabeled Data}.
\newblock In \emph{Proceedings of the IEEE/CVF Conference on Computer Vision
  and Pattern Recognition (CVPR)}, 2024.

\bibitem[Ye et~al.(2025)Ye, Zhang, Daxberger, Chen, Lin, Li, Zhang, You, Xu,
  Gan, et~al.]{ye2024mm}
Hanrong Ye, Haotian Zhang, Erik Daxberger, Lin Chen, Zongyu Lin, Yanghao Li,
  Bowen Zhang, Haoxuan You, Dan Xu, Zhe Gan, et~al.
\newblock {MM-Ego: Towards Building Egocentric Multimodal LLMs}.
\newblock In \emph{International Conference on Learning Representations
  (ICLR)}, 2025.

\bibitem[Yi et~al.(2025)Yi, Ye, Zheng, M\"uller, Pavlakos, Ma, Malik, and
  Kanazawa]{yi2024egoallo}
Brent Yi, Vickie Ye, Maya Zheng, Lea M\"uller, Georgios Pavlakos, Yi Ma,
  Jitendra Malik, and Angjoo Kanazawa.
\newblock Estimating body and hand motion in an ego-sensed world.
\newblock In \emph{Proceedings of the IEEE/CVF Conference on Computer Vision
  and Pattern Recognition (CVPR)}, 2025.

\bibitem[Zhang et~al.(2023)Zhang, Li, and Bing]{zhang2023video}
Hang Zhang, Xin Li, and Lidong Bing.
\newblock {Video-LLaMA: An Instruction-tuned Audio-Visual Language Model for
  Video Understanding}.
\newblock \emph{Proceedings of the Conference on Empirical Methods in Natural
  Language Processing (EMNLP)}, 2023.

\bibitem[Zhang et~al.(2024{\natexlab{a}})Zhang, Zhang, Li, Zeng, Yang, Zhang,
  Wang, Tan, Li, and Liu]{zhang2024long}
Peiyuan Zhang, Kaichen Zhang, Bo Li, Guangtao Zeng, Jingkang Yang, Yuanhan
  Zhang, Ziyue Wang, Haoran Tan, Chunyuan Li, and Ziwei Liu.
\newblock Long context transfer from language to vision.
\newblock \emph{arXiv preprint arXiv:2406.16852}, 2024{\natexlab{a}}.

\bibitem[Zhang et~al.(2024{\natexlab{b}})Zhang, Wu, Li, Li, Ma, Liu, and
  Li]{zhang2024videoinstructiontuningsynthetic}
Yuanhan Zhang, Jinming Wu, Wei Li, Bo Li, Zejun Ma, Ziwei Liu, and Chunyuan Li.
\newblock Video instruction tuning with synthetic data.
\newblock \emph{arXiv preprint arXiv:2410.02713}, 2024{\natexlab{b}}.

\bibitem[Zhao et~al.(2024)Zhao, Ma, Kong, and Fowlkes]{zhao2024instance}
Yunhan Zhao, Haoyu Ma, Shu Kong, and Charless Fowlkes.
\newblock {Instance tracking in 3D scenes from egocentric videos}.
\newblock In \emph{Proceedings of the IEEE/CVF Conference on Computer Vision
  and Pattern Recognition (CVPR)}, 2024.

\bibitem[Zhao et~al.(2025)Zhao, Lu, Huo, Du, Yue, Guo, Wang, Chen, and
  Liu]{zhao2024videoniah}
Zijia Zhao, Haoyu Lu, Yuqi Huo, Yifan Du, Tongtian Yue, Longteng Guo, Bingning
  Wang, Weipeng Chen, and Jing Liu.
\newblock {Needle In A Video Haystack: A Scalable Synthetic Framework for
  Benchmarking Video MLLMs}.
\newblock In \emph{International Conference on Learning Representations
  (ICLR)}, 2025.

\bibitem[Zhou et~al.(2018)Zhou, Xu, and Corso]{zhou2018towards}
Luowei Zhou, Chenliang Xu, and Jason Corso.
\newblock Towards automatic learning of procedures from web instructional
  videos.
\newblock In \emph{Proceedings of the AAAI Conference on Artificial
  Intelligence}, 2018.

\end{thebibliography}
}

 \clearpage
\setcounter{page}{1}
\maketitlesupplementary

\appendix

\addtocontents{toc}{\setcounter{tocdepth}{2}} 

\renewcommand{\thefigure}{A\arabic{figure}}
\renewcommand{\thetable}{A\arabic{table}}
\setcounter{figure}{0}
\setcounter{table}{0}

{
\hypersetup{linkcolor=black}
\tableofcontents
}

\section{Video Showcase}
We share a video showcasing our annotations on our webpage: \url{http://hd-epic.github.io} containing three parts.

\noindent\textbf{start $\rightarrow$ 01:16. Annotations overview.} This is a walk-through of Fig.~\ref{fig:f1}, showing our annotation hierarchy and how annotations are linked to the 3D Digital Twin.

\noindent\textbf{01:16 $\rightarrow$ 04:24. Annotation showcase on one sequence from one kitchen.} 
For one video, we show a walk through of the various annotations on this same video.

\noindent\textbf{04:25 $\rightarrow$ end. Annotation examples from other kitchens.}
First are examples of our highly-detailed narrations, followed by examples of hand and object segmentations. We then present audio annotations with waveforms. Next, examples of object-fixture assignment are shown in the Digital Twin. Finally, examples of ingredient nutrition, running total nutrition, and total nutrition for whole recipes are shown.

\section{Data Collection}
\label{sec:data_collection_sup}

In this section, we provide more details of: the recruitment and equipment used in Sec.~\ref{subsec:recruitment_equipment_sup}; instructions and collected data in Sec.~\ref{subsec:instructions_collected_data_sup}; narrations in Sec.~\ref{subsec:narrations_supp}; and Post-Processing in Sec.~\ref{subsec:post_processing_supp}.

\subsection{Recruitment and Equipment}
\label{subsec:recruitment_equipment_sup}
\begin{figure}
    \centering
    \includegraphics[width=0.9\linewidth]{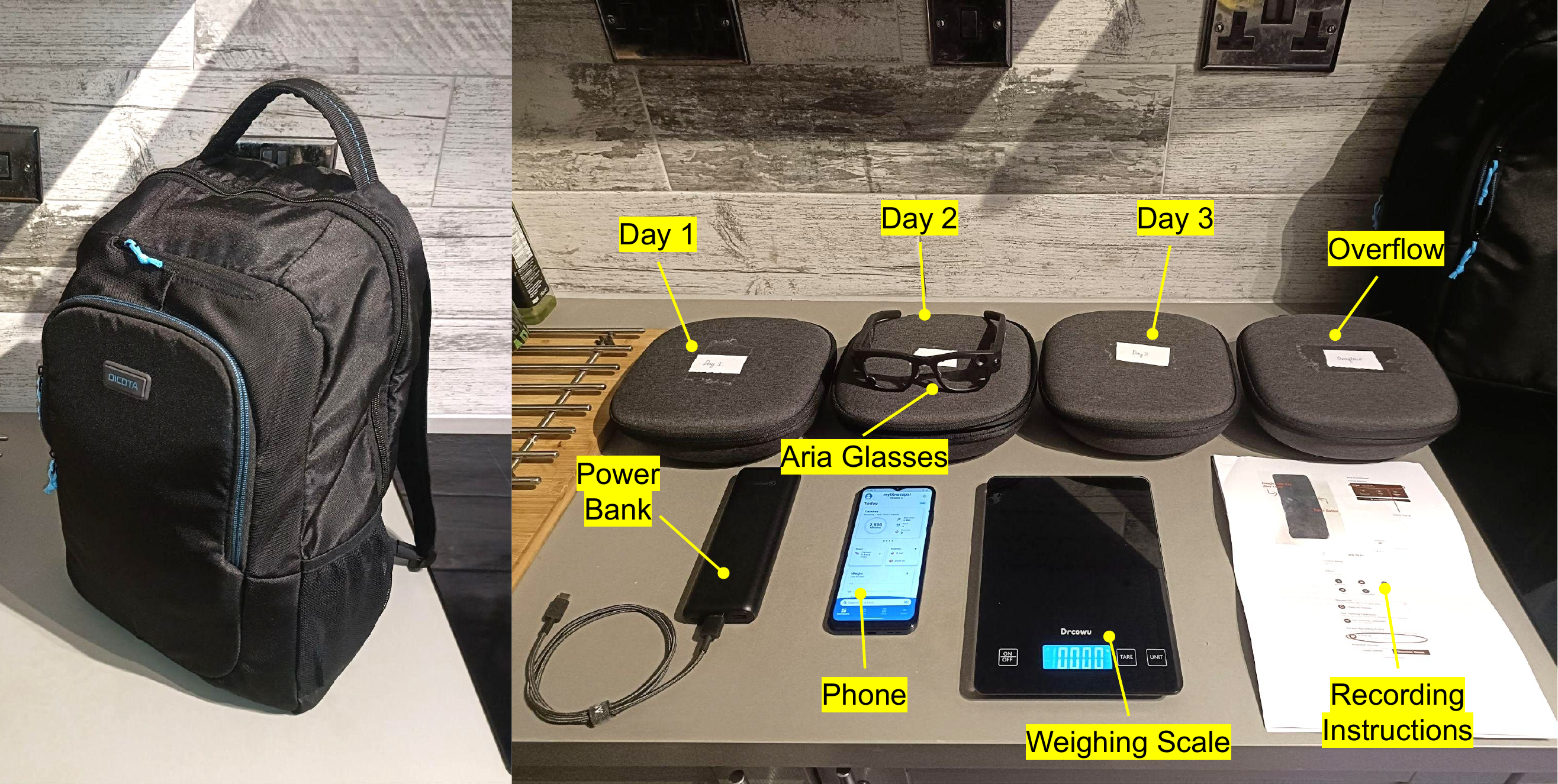}
    \caption{Participant P04 unpacking the data recording equipment.}
    \label{fig:backpack}
\end{figure}

Each participant engaged in a long-time commitment (avg. 50 hours): data collection, reviewing footage, and providing detailed narrations including activities/recipe/nutrition information. %
Participants signed a consent form and a data release form after which they were rewarded monetarily. %
In Fig.~\ref{fig:backpack}, we show the devices provided to participants for the data collection, packaged tightly in one backpack that the participants took home.
Data collection was carried out using Project Aria devices~\cite{Somasundaram2023ProjectAA}---a multi-sensor platform, in the form factor of a pair of glasses\footnote{Participants who require glasses and cannot use lenses were excluded}, with 3 front facing cameras (1 RGB and 2 SLAM), 7 microphones and inward facing cameras for gaze estimation\footnote{We used Profile 28---$1408\times1408$ resolution 30 FPS RGB footage, 60 FPS eye tracking, 30 FPS SLAM}.
Due to the storage limits, four devices were provided to each participant. %
We supplied a smartphone with the Aria Mobile Companion app, with all four devices already registered, as well as the MyFitnessPal app~\cite{myfitness} installed. 
Due to battery life limits, devices were sometimes wired to a pocket-carried power bank whilst recording long sessions.
Participants also received recording instructions and a set of digital weighing scales for nutritional tracking.
The scales' display is clear to read, to enable OCR readings.

\subsection{Instructions and Collected Data}
\label{subsec:instructions_collected_data_sup}
Participants were recruited to record their daily kitchen activities for at least three consecutive days. 
All recordings were unscripted: %
participants were asked to wear the glasses every time they walked into their kitchen, pressing the recording button upon entering, and stopping the recording when they left the kitchen.
Videos were recorded between January and July 2024. The number of hours collected per participant ranged from 3.5 to 7.2, with an average of 4.6. 
We collected a total of 156 videos, with an average length of 15.9 ($\pm 14.5$) mins. %
In total, we collected 41.3 hours (4.46M frames) of footage from 9 participants.

Following data collection, participants provided the recipe they freely prepared. They cited the source of the recipe (e.g.\ online website or cookbook) and any modifications to ingredients or steps. %
We collected a total of 69 recipes from the 9 participants. 
Our recipes cover various cuisines including: Masala Dosa (Indian), Cacio e Pepe (Italian), Sfesiha (Lebanese), and Banana Bread Chocolate Chip Cookies (American) with an average (max) of 6.6 (18) steps and 8.1 (24) ingredients per recipe.
Additionally, participants provided time segments (as start-end times) covering each step of their recipes.
Recipes typically expand over several videos, on average taking 4 hours and 48 minutes to complete across 2.1 videos from starting preparation to conclusion\footnote{In contrast, Ego-Exo4D recipes are 8.5 mins on average.}.
Our longest captured recipe (P03\textunderscore R03) took 2 days and 6 hours to complete.

We found that participants would interleave the preparation of multiple dishes, often preparing several courses simultaneously (e.g.\ salad or dessert while preparing a main).
Whilst natural in daily routines, this has not been captured in previous recordings which consist of edited online videos~\cite{zhou2018towards} or short recipes captured under controlled settings~\cite{EgoExo4d}.

To track the nutrition values of prepared meals, we instructed participants to weigh ingredients using the digital scales and log nutrition values via the provided MyFitnessPal app. %
By combining the weight and ingredient information, detailed nutrition information associated with recipes were collected, adding an interesting dimension where we can estimate the weighed ingredients from visual data. %
While spices are often included as ingredients, we do not capture their nutrition values as they rarely alter these values and are mainly included to improve the dish's taste.
In total, participants used 558 ingredients including:
high protein ingredients such as tuna and kidney beans, high carb such as dates and flour, and high fat such as sour cream and pine nuts. %
Participants prepared both high calorific meals such as Lazy Cake Recipe (P01\textunderscore R05, 4.8K calories) and low-calorie meals including Crispy Cucumber Salad (P08\textunderscore R09, 274 calories).

\subsection{Narrations} 
\label{subsec:narrations_supp}
\begin{figure*}
    \centering
\includegraphics[width=\linewidth]{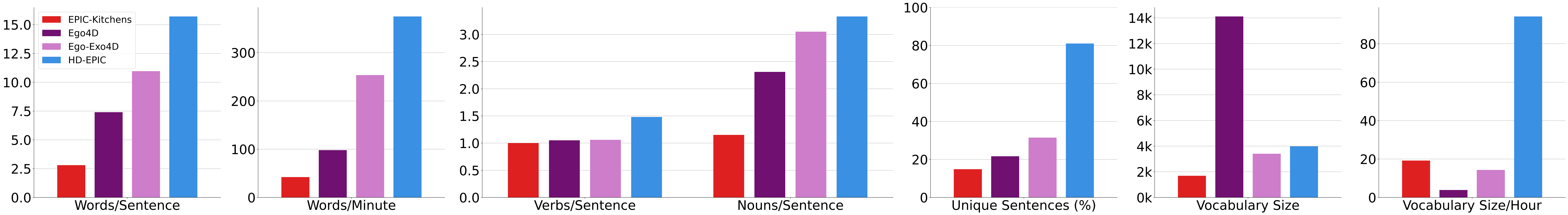}
    \caption{Dataset Statistics: Comparison with EPIC-Kitchens, Ego4D and Ego-Exo4D.}
    \label{fig:narr_diff}
\end{figure*}
We collect sentences detailing the actions carried out by participants in each video.
Following~\cite{Damen2022RESCALING}, we instruct the participants to record spoken narrations of all actions throughout the videos and re-purpose the web-based narrator tool used in~\cite{EgoExo4d} for expert commentary.
Using the tool, participants pause the video and record a narration whenever they observe an action and provide detailed narrations which include at least one verb and noun, and information about \emph{what} they did \emph{why} and \emph{how} they did it when relevant. %

This results in a rich set of narrations which are denser and more detailed than previous egocentric datasets.
Fig.~\ref{fig:narr_diff} shows that HD-EPIC narrations are much denser with a higher words per sentence and words per minute than Ego-Exo4D, the next densest dataset, demonstrating a significant increase in detail.  %
We also see increases in the numbers of nouns and verbs per sentence as sentences describe actions in more detail, even common actions are described uniquely---e.g.\ `pick plate from a pile of plates using the right hand' and `pick plate from a lower shelf using the left hand'. 
In total, 81\% of all narrations in \DName are unique sentences, compared to only 31\% in Ego-Exo4D, 22\% in Ego4D, and 15\% in EPIC-KITCHENS.
Non-unique narrations have recently been identified as a concern in recent works~\cite{perrett2024unique}.

\subsection{Post-Processing: Multi-Video SLAM and Gaze}
\label{subsec:post_processing_supp}

We make use of the Multi-Video SLAM API provided by the Aria Machine Perception Services (MPS) \cite{ariamps} to reconstruct a single point cloud from all recordings obtained in a kitchen, across multiple days. It is important to highlight that no additional scanning is used to reconstruct the scene---it is simply reconstructed from the unscripted recordings. %
We obtain the 3D point clouds and 2D tracked points along with camera trajectories consisting of 1kHz frequency and 6DoF, estimated using odometry, for all videos. Out of the $156$ videos, $3$ videos could not be successfully processed via SLAM API. %
For these videos, we run COLMAP~\cite{schoenberger2016mvs, schoenberger2016sfm} following EPIC Fields~\cite{EPICFields2023} pre-processing to obtain camera poses and manually align them to the corresponding kitchen obtained from MPS.

We also use the MPS services to obtain yaw and pitch angle for eye gaze and obtain a unit vector along the gaze direction by anchoring it to the central pupil frame.
The depth of the gaze vector is estimated by projecting it into the world frame and obtaining its first intersection with any surface on the 3D space. We then use the device calibration and project the ray to obtain 2D location on the image frame. After gaze is calculated, we remove the gaze camera input from all VRS files for anonymity.

We also provide additional information on the anonymisation process for videos.
To ensure the privacy of the participants, faces were anonymised in the rare case where they are visible due to reflections or accidental capture.
Participants were asked to view their videos and notify us of cases where faces or identifiable information was present, we also anonymised videos which we found to need anonymisation.
We then used black boxes to mask these faces/identifiable information to avoid potential regeneration. %
In total 8 videos were marked for anonymisation (6\% of our 156 videos)---we redacted a total of 2774 frames. Correspondingly, we manually marked and anonymised these frames in the two gray-scaled SLAM cameras stored in the VRS files.

We separately post-process the VRS files to convert them to mp4. For the audio, we select channels 5 and 6 to get a binaural audio out of the 7 audio channels in the input recording. We select these channels as they are closer to the ears and remove other audio channels to mitigate breathing noises.  We then extract the frames from VRS matching the audio timestamp. Finally, we obtain a video with $30$ FPS and binaural audio with $48$ kHz sampling rate.
These videos are used in all follow-up annotations and benchmarking.

\begin{table*}[t]
\centering
\setlength{\tabcolsep}{3pt}
\resizebox{\textwidth}{!}{%
\begin{tabular}{@{}lcccccccccccccc@{}}
\toprule
\textbf{Dataset} & \textbf{Val\&Test} & \textbf{Action} & \textbf{Unscripted} & \textbf{Free} & \textbf{Recipe} & \textbf{Nutrition} & \textbf{Gaze} & \textbf{Audio} & \textbf{Object} & \textbf{Hand} & \textbf{3D object} & \textbf{Labelled 3D} & \textbf{Camera} & \textbf{Fully} \\
& \textbf{Hours} & \textbf{Segments} & & \textbf{Setting} & & & & \textbf{Labels} & & & \textbf{over time} & \textbf{environment} & \textbf{pose} & \textbf{annotated} \\ \midrule
CMU-MMAC~\cite{de2009guide} & 16.6 & \cmark & \xmark & \xmark & \xmark & \xmark & \xmark & \xmark & \xmark & \xmark  & \xmark & \xmark & \xmark & \xmark \\
ADL~\cite{pirsiavash2012detecting} & 5.8 & \cmark & \cmark & \cmark & \xmark & \xmark & \xmark & \xmark & B-Box & \xmark & \xmark & \xmark & \xmark & \cmark \\
UTE~\cite{lee2012discovering} & 16.9 & \xmark &\cmark &\cmark & \xmark & \xmark & \xmark & \xmark & Polygon & \xmark & \xmark & \xmark & \xmark & \cmark \\
KrishnaCam~\cite{singh2016krishnacam} & 14.0 & \xmark &\cmark &\cmark & \xmark & \xmark & \xmark & \xmark & \xmark & \xmark & \xmark & \xmark & \xmark & \cmark \\
Charades-EGO~\cite{sigurdsson2018actor} & 6.7$^1$ & \cmark & \xmark & \cmark & \xmark & \xmark & \xmark & \xmark & \xmark & \xmark & \xmark & \xmark & \xmark & \cmark \\
EGO-CH~\cite{ragusa2020ego} & 26.5 & \cmark &\cmark &\xmark & \xmark & \xmark & \xmark & \xmark & B-Box & \xmark & \xmark & \xmark & \xmark  & \xmark \\
MECCANO~\cite{ragusa2021meccano} & 3.0 & \cmark & \xmark &\xmark & \xmark & \xmark & \cmark & \xmark & B-Box & B-Box & \xmark & \xmark & \xmark  & \cmark \\
EGTEA Gaze+~\cite{li2021eye} & 19.1 & \cmark &\xmark &\xmark & \xmark & \xmark & \cmark & \xmark & \xmark & Mask &  \xmark & \xmark & \xmark & \cmark \\
HOI4D~\cite{liu2022hoi4d} & 11.4 & \cmark &\xmark &\xmark & \xmark & \xmark & \xmark & \xmark & Mask & Mask &  \cmark & \cmark & \xmark  & \cmark \\
Assembly101~\cite{sener2022assembly101} & 66.8$^1$ & \cmark &\xmark &\xmark & \xmark & \xmark & \xmark & \xmark & \xmark & 3D pose & \xmark & \xmark & \xmark & \cmark \\
EPIC-KITCHENS-100~\cite{Damen2022RESCALING} & 25.3 & \cmark &\cmark &\cmark & \xmark & \xmark & \xmark & \cmark & Mask & Mask & \cmark & \xmark & \cmark  & \xmark \\
Ego4D~\cite{Ego4D2022CVPR} & 288.7$^2$ & \cmark &\cmark &\cmark & \xmark & \xmark & \xmark & \xmark & B-Box & B-Box &  \xmark & \xmark & \xmark  & \xmark \\
HoloAssist~\cite{HoloAssist2023} & 49.8 & \cmark & \xmark & \xmark & \xmark & \xmark & \cmark & \xmark & \xmark  & 3D pose & \xmark & \xmark & \cmark & \cmark \\ 
Aria Digital Twin~\cite{pan2023aria} & 8.1 & \xmark & \xmark & \xmark & \xmark & \xmark & \cmark & \xmark & Mask & \xmark &  \cmark & \cmark & \cmark  & \cmark \\
Aria Everyday Activities~\cite{lv2024aria} & 7.3 & \xmark & \xmark &\cmark & \xmark & \xmark & \cmark & \xmark & \xmark & \xmark & \xmark & \xmark & \cmark  & \xmark \\
Aria Everyday Objects~\cite{straub2024efm3d} & 0.4 & \xmark &\cmark &\cmark & \xmark & \xmark & \xmark & \xmark & B-Box & \xmark & \cmark & \xmark & \cmark  & \cmark \\
IndustReal~\cite{schoonbeek2024industreal} & 3.5 & \cmark &\xmark &\xmark & \xmark & \xmark & \cmark & \xmark & \xmark & 3D pose & \xmark & \xmark & \xmark  & \cmark \\ 
IT3DEgo~\cite{zhao2024instance} & 4.6 & \xmark & \cmark & \cmark & \xmark & \xmark  & \xmark & \xmark & B-Box & \xmark & \cmark & \xmark & \cmark  & \cmark \\ CaptainCook4D~\cite{captaincook4d} & 42.5 & \cmark$^3$ & \xmark & \xmark & \cmark$^4$ & \xmark & \xmark & \xmark & \xmark & 3D pose & \xmark & \xmark & \cmark & \xmark \\ 
Ego-Exo4D~\cite{EgoExo4d} & 85.1 & \xmark & \cmark & \cmark & \cmark$^3$ & \xmark & \cmark & \xmark & Mask & 3D pose & \xmark & \xmark & \cmark  & \xmark \\ \midrule
\textbf{\DName} & 41.3 & \cmark & \cmark & \cmark &\cmark & \cmark & \cmark & \cmark & Mask & Mask &  \cmark & \cmark & \cmark  & \cmark \\ \bottomrule
\end{tabular}%
}
\caption{Comparing egocentric video datasets. 
$^1$Subset of egocentric videos. $^2$Episodic Memory, Hands+Object and Forecasting benchmarks.
$^3$Only key steps annotated.  
}
\label{tab:dataset_comparison_full}
\end{table*}

\section{Annotation Pipeline}
\label{sec:annotations_pipeline_supp}

\subsection{Annotating Recipe Steps and Ingredients}
\label{sec:annotations-recipe_supp}

\begin{figure*}
    \centering
    \includegraphics[width=1.0\linewidth]{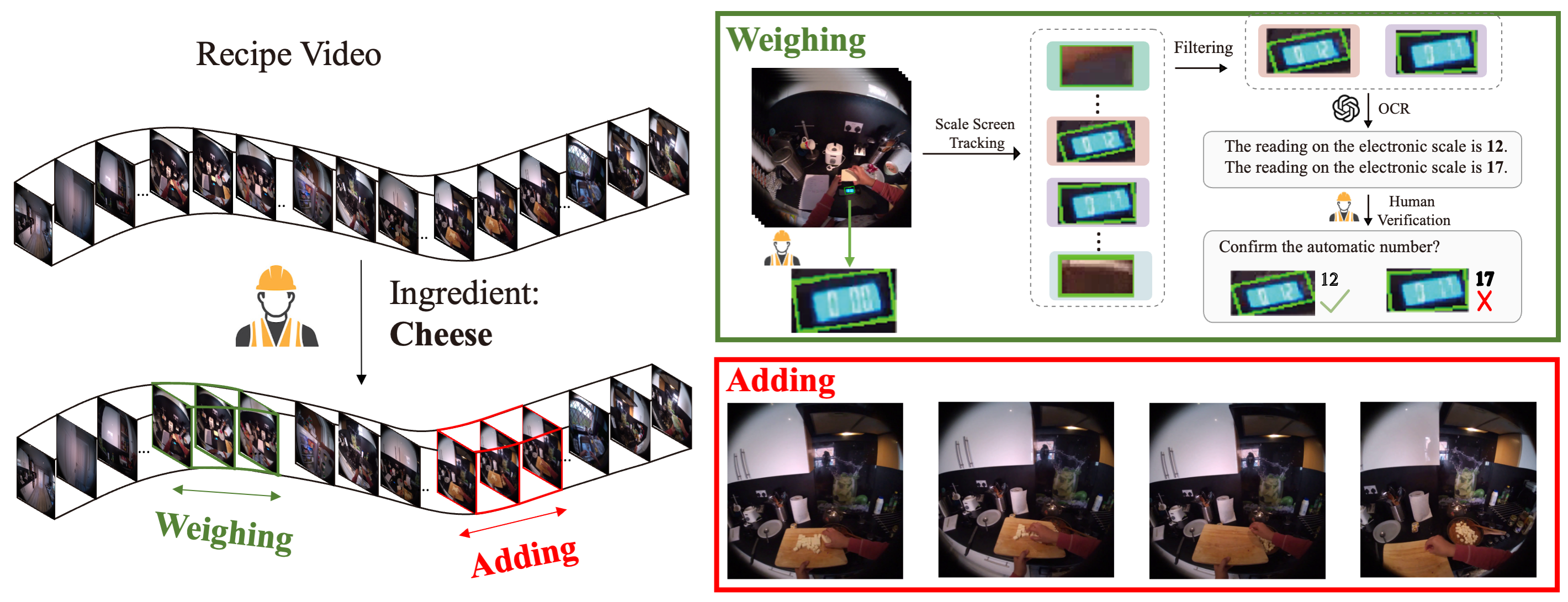}
    \caption{Per ingredient, we temporally label weighing (if present) and adding to recipe. Weighing is OCRed and manually verified.}
    \label{fig:nutrition-anno}
\end{figure*}

We match the annotated start-end times to recipes steps to automatically compile `recipe videos',
which we use for annotating recipe steps and ingredients.
We further link recipes, steps, and ingredients by annotating time segments where ingredients are added to the recipe.
For each ingredient, we temporally annotate the weighing and adding actions. The weighing segments allow tracking the amount of each ingredient from zero to the total used in the dish whereas %
the adding segments mark when an ingredient is incorporated in the recipe. When weighing is performed with the scales, we obtain automatic OCR readings of the scales which are manually verified and checked to ensure the full amount matches the quantity in the nutrition information provided by the partcipant.
Fig.~\ref{fig:nutrition-anno} visualises the annotations per ingredient.

\subsection{Fine-Grained Actions}
\label{sec:fine_grained_details_supp}

\subsubsection{Transcriptions}
The audio narrations, provided by the participants %
are first converted into timestamped transcriptions.
We use an ensemble of audio transcription models, namely Whisper~\cite{radford2023robust}, Qwen-Audio~\cite{chu2023qwen} and Nemo~\cite{kuchaiev2019nemo}, to improve the quality of transcriptions.
We find that each of these models makes a different set of errors in understanding the diverse spoken accent of participants, so we opt for the consensus of models along with manual checks.
For each narration, we select a transcription from one of these models based on a number of checks. Transcriptions with spelling errors or words not in a whitelisted kitchen vocabulary, which we populate during transcription cleaning, are excluded. If there is more than one unique transcription, we select the highest model confidence. 
If all transcriptions are excluded by the spelling or whitelist checks, but multiple models have produced the same transcription, then we select this transcription. Otherwise the Whisper transcription is selected as a default.
We then manually check all candidate transcriptions using dedicated human checkers, who listen to the audio and verify or correct the transcription.
We then fix typos and 
obtain a set of cleaned detailed transcriptions.

\begin{figure*}
    \centering
    \includegraphics[width=\linewidth]{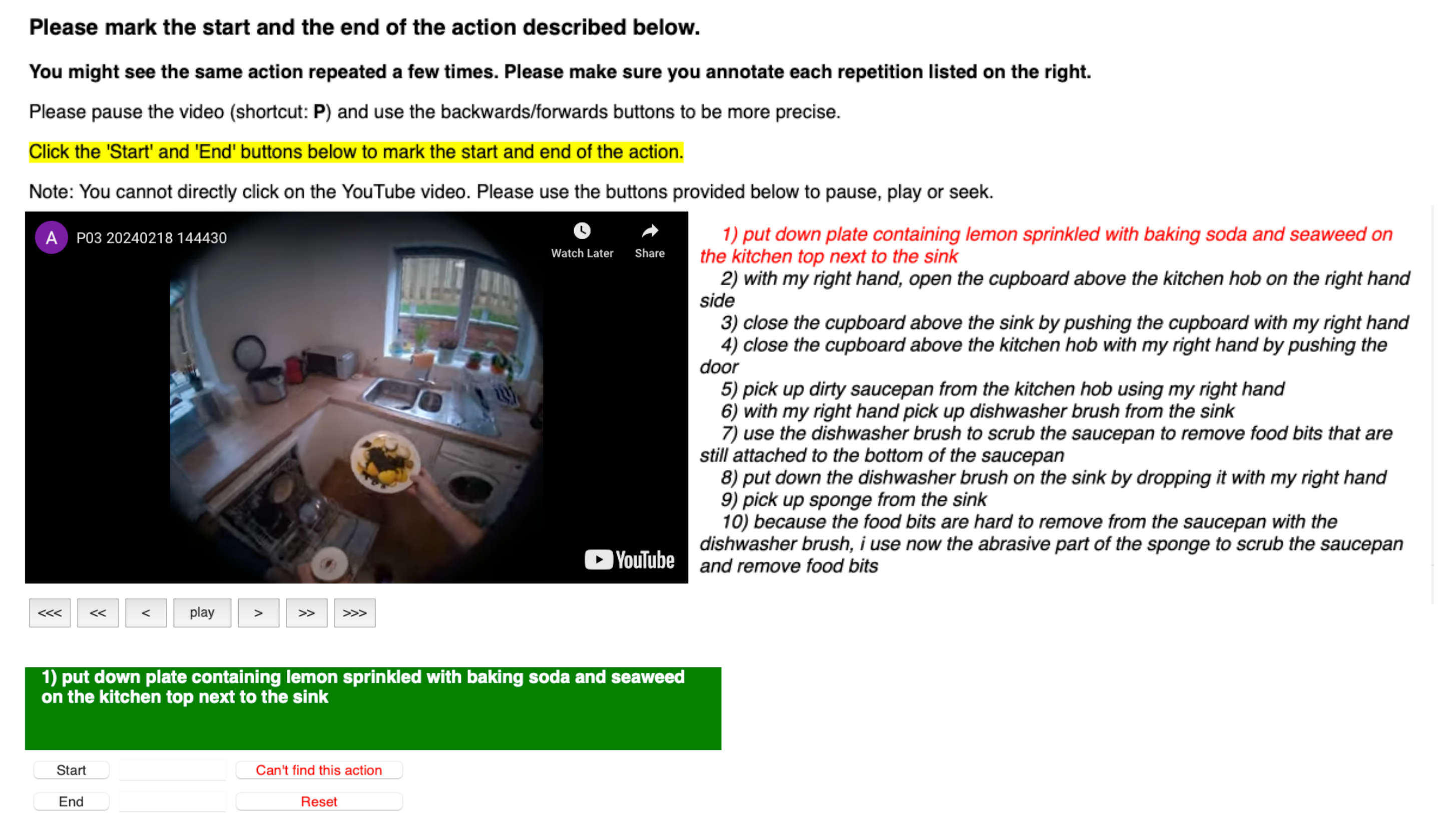}
    \caption{Interface for annotating action boundaries}
    \label{fig:amt_interface}
\end{figure*}

\subsubsection{Action Annotations}

For each narration, we label the precise time segment bounded by when the action in the narration starts and concludes, using AMT workers.
We divide the video into HITs of 10 consecutive narrations. Multiple annotators ($\mathcal{K}_a \ge 3$) are tasked to annotate both start and end times of the segment containing the narrated action, primed by the timestamp, denoted by $A_i=[t_{s_i}, t_{e_i}]$. %
The option to skip an action is also provided to the annotators if they are unable to find it in the video; accounting for rare cases when the action occurs off-camera.
The interface used to annotate action boundaries is shown in \cref{fig:amt_interface}.
We ensure we have sufficient inter-annotator agreements between at least three annotators, adding annotators until satisfied.

We select the best annotation by measuring inter-annotator agreement. For narration $i$, let $A_{i}(j)$ represent the annotation from annotator $j$. The agreement score for annotator $j$ on narration $i$ is $\alpha_i(j) = \frac{1}{\mathcal{K}_a} \sum_{k=1}^{\mathcal{K}_a} IOU(A_{i}(j), A_{i}(k))$, where $IOU$ is temporal intersection-over-union. The average agreement score across the HIT is $\alpha(j) = \frac{1}{N_H} \sum_{i=1}^{N_H} \alpha_i(j)$, with $N_H$ actions per HIT. We select the best annotator $\hat{j}$ as the one with the highest average agreement on their three lowest-scoring actions and similarly find $\hat{k}$ as the second-best annotator. 

The ground-truth action segment $A_i$ is then found as:
\begin{equation}
A_i = 
\begin{cases} 
\text{Union}(A_i(\hat{j}), A_i(\hat{k})), & \text{if } IoU(A_i(\hat{j}), A_i(\hat{k})) > 0.5 \\
A_i(\hat{j}), & \text{otherwise}
\end{cases}
\end{equation} where we choose to merge annotations from the two best annotators for an action segment if they have a strong agreement, which helps to avoid overly tight segments~\cite{damen2018scaling}.

To ensure annotation quality, we require minimum annotator agreement $\min(\alpha(j)) \geq 0.3$ for each HIT. Failed HITs are re-attempted with up to 3 new annotators, then manually annotated if still below threshold. We also perform quality checks on randomly sampled action segments.

\begin{table*}[t]
\centering
\renewcommand{\arraystretch}{1.5}
\fontsize{6.5pt}{8.4pt}\selectfont

\begin{tabular}{p{4cm} p{1cm} p{1.5cm} p{1cm} p{1.5cm} p{2.9cm} p{2.9cm}} %
\hline
\textbf{Narration} & \textbf{Verbs} & \textbf{Nouns} & \textbf{Hands} & \textbf{Main Action} & \textbf{How} & \textbf{Why} \\ \hline

Transferring the plate stack to my right hand so that I can turn off the tap with my left hand. 
& transfer & plate stack, tap & right hand, left hand & transfer plate stack & - & so that I can turn off the tap with left hand \\ \hline

Push the bag of potatoes along the bottom shelf of the fridge so that I can close the fridge. 
& push & bag of potatoes, bottom shelf of fridge, fridge & - & push bag of potatoes & - & so that I can close the fridge \\ \hline

Pick up the two chopped lemon pieces using both hands so as to weigh them.
& pick up & two chopped lemon pieces & both hands & pick up two chopped lemon pieces & using both hands & to weigh them \\ \hline

Brush bit of onion skin off the chopping board using the knife in my right hand because I don't want to accidentally eat it.
& brush & bit of onion skin, chopping board, knife & right hand & brush bit of onion skin & using the knife in my right hand & because I don't want to accidentally eat it \\ \hline

Shake the foil wrapping to make sure it is empty.
& shake & foil wrapping & - & shake foil wrapping & - & to make sure it is empty \\ \hline

Turn on the gas burner by twisting the dial in the right hand and pressing down the ignition switch in the left hand.
& turn on, twist, press & gas burner, dial, ignition switch & right hand, left hand & turn on gas burner & by twisting the dial in the right hand and pressing down the ignition switch in the left hand & - \\ \hline

I open the washing up liquid bottle by flicking the bottle lid up with my right hand.
& open, flick up & washing up liquid bottle, bottle lid & right hand & open washing up liquid bottle & by flicking the bottle lid up with right hand & - \\ \hline

Wipe down the countertop using the wet kitchen roll in the right hand, pushing the food and sugar into the food bin that's held next to the countertop using the left hand.
& wipe down, push & countertop, wet kitchen roll, food, sugar, food bin & right hand, left hand & wipe down countertop & using the wet kitchen roll in the right hand, pushing the food and sugar into the food bin held next to the countertop using the left hand & - \\ \hline

Put the knife in the drawer by putting it vertically in a slot.
& put & knife, drawer, slot & - & put knife & by putting it vertically in a slot & - \\ \hline

Open the semolina flour bag by unrolling the top part of the plastic bag.
& open, unroll & semolina flour bag, top part of plastic bag & - & open semolina flour bag & by unrolling the top part of the plastic bag & - \\ \hline

\end{tabular}
\caption{Parsing Examples}
\label{tab:parsingexamples}
\end{table*}

\subsubsection{Parsing and Clustering}
We next parse the open vocabulary narrations so they can be used for tasks that expect a closed vocabulary, such as action recognition. 
Additionally, we wish to identify the parts of the narration where the participant is detailing the manner in which the action was carried out and the goal achieved by carrying out the action.
We achieve these by parsing these open vocabulary narrations as follows.
As a result of the fine-grained nature of the collected narrations, it was expected that LLMs with their impressive performance in NLP tasks would provide better results compared to smaller language models. To that end different open source LLMs were tested for the task of extracting the pairs of nouns and verbs and identifying the main action. Initial testing showed mixed results with hallucinations being a major issue. Even though a combination of different prompting techniques such as few-shot prompting improved the results, given the number of narrations, the results were too unpredictable. 

Each narration transcription is parsed into nouns, verbs, and any noted hands (Left/Right/Both) using spaCy's Part-of-Speech tagging \cite{honnibal2017spacy}.
To better identify compound nouns, we supplement spaCy's noun chunks
with Constituency Parsing \cite{kitaev2018constituency}, 
and use pattern matching to filter out irrelevant parts, such as articles.
For pronouns (e.g.\ `it's'), we use the co-reference resolution within the spaCy pipeline along with heuristics to replace these with corresponding nouns, selecting from the directly preceding narration. %
We also identify the primary verb-noun pairs, as the `ROOT' verb in spaCy with its corresponding noun, to form the main action.
To enhance accuracy, we also apply heuristics to de-prioritize some verbs (e.g.\ ``use something") from being considered the main action. We show some parsing examples in Table \ref{tab:parsingexamples}.

\begin{figure*}[t]
    \centering
    \includegraphics[width=\linewidth]{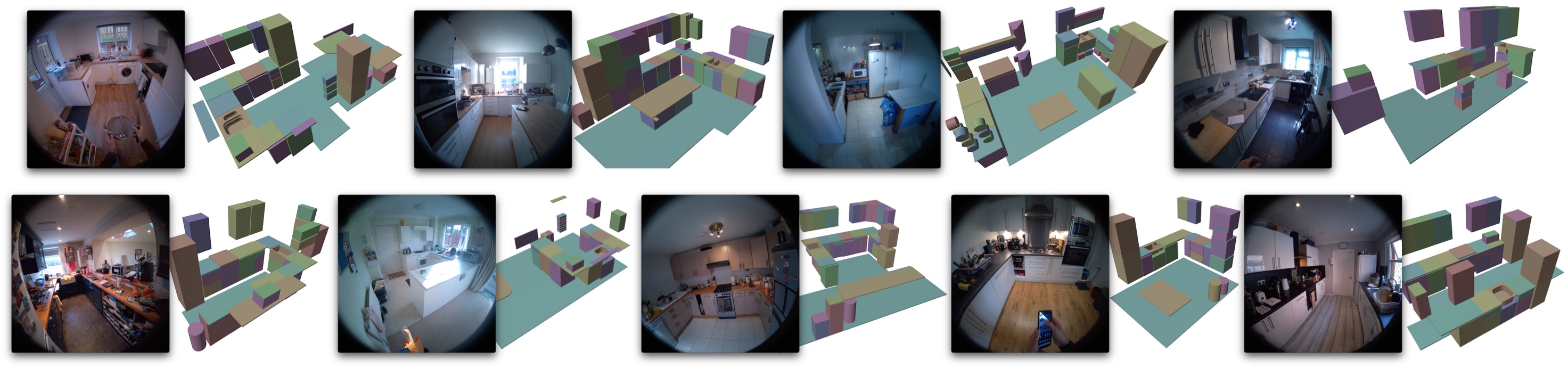}
    \caption{All kitchens with their fixture annotations (coloured randomly).}
    \label{fig:all_blenders}
\end{figure*}
\begin{figure}
    \centering
    \includegraphics[width=1.0\linewidth]{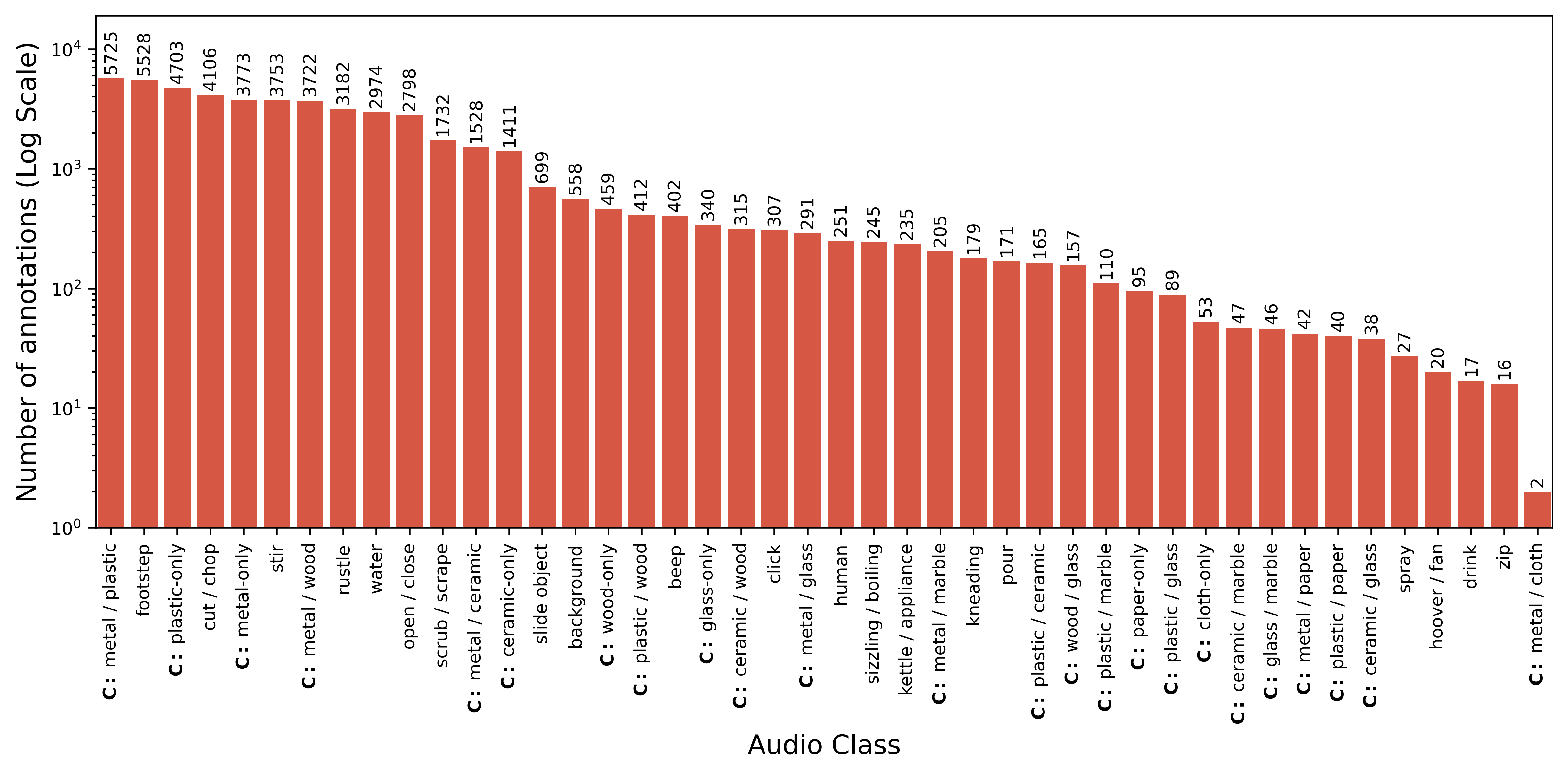}
    \caption{Distribution of all audio-classes in \DName Sounds 
    }
    \label{fig:DEPIC-Sounds_dist}
\end{figure}

Similar to the POS tagging, clustering nouns and verbs using LLMs was too noisy and inconsistent. The clusters from~\cite{Damen2022RESCALING} were used in addition to a combination of heuristics, word embeddings, and manual assignment to 
assign any new verbs or nouns to these clusters.
We assigned $1,172$ new verbs (such as \textit{brew}, \textit{simmer}, \textit{burn}) and $16,812$ new nouns (\eg \textit{mortar}, \textit{nutmeg}, \textit{glue}) to the clusters.
We additionally added 9 new verb clusters and 3 new noun clusters which did not fit in the clusters from~\cite{Damen2022RESCALING}.

\subsubsection{Sound Annotations}
\label{sec:annotations-audio_supp}

We follow a procedure similar to~\cite{EPICSOUNDS2023} to collect audio annotations. 
First, we only use a stereo 2-channel audio, extracted from the 5$^{th}$ and 7$^{th}$ microphones. %
These offer the closest audio input to the two ears and in our experience decreases the presence of breathing sounds in the audio signal, thus better allowing annotators to focus on the sound event.
We divide the untrimmed audio, for one video, into chunks based on a silence threshold to provide manageable project sizes to the annotators, whilst avoiding splitting an ongoing sound.

We use a modified version of the VIA Interface~\cite{dutta2019vgg}. %
Annotators were tasked with listening to the audio and labelling the start and end times of any distinct audio event they heard.
They selected labels from the classes from~\cite{EPICSOUNDS2023}, though annotators could still provide a free-form text description. 
We post-process the annotations, 
typo-correcting free-form text descriptions and grouping into the classes from~\cite{EPICSOUNDS2023}. 
We did not find a sufficient number of new free-form text descriptions to warrant additional classes, hence the classes match that of~\cite{EPICSOUNDS2023}. 

To reduce label noise, 
we used a customised version of the LISA~\cite{duta20lisa} interface. 
Here, the annotators were provided the audio-visual stream of each sound annotation 
and asked to signify if the current label was correct. 
If the annotator deemed the current label incorrect, we asked for a proposed correction. 

We show the distribution of audio-classes in Fig.~\ref{fig:DEPIC-Sounds_dist}. %

\subsection{Digital Twin}
\label{sec:digital_twin_supp}

\noindent{\textbf{Environments.}}
Fig. \ref{fig:all_blenders} shows all kitchens used, along with their annotated fixtures.

\begin{figure}[t]
    \centering
    \includegraphics[width=\linewidth]{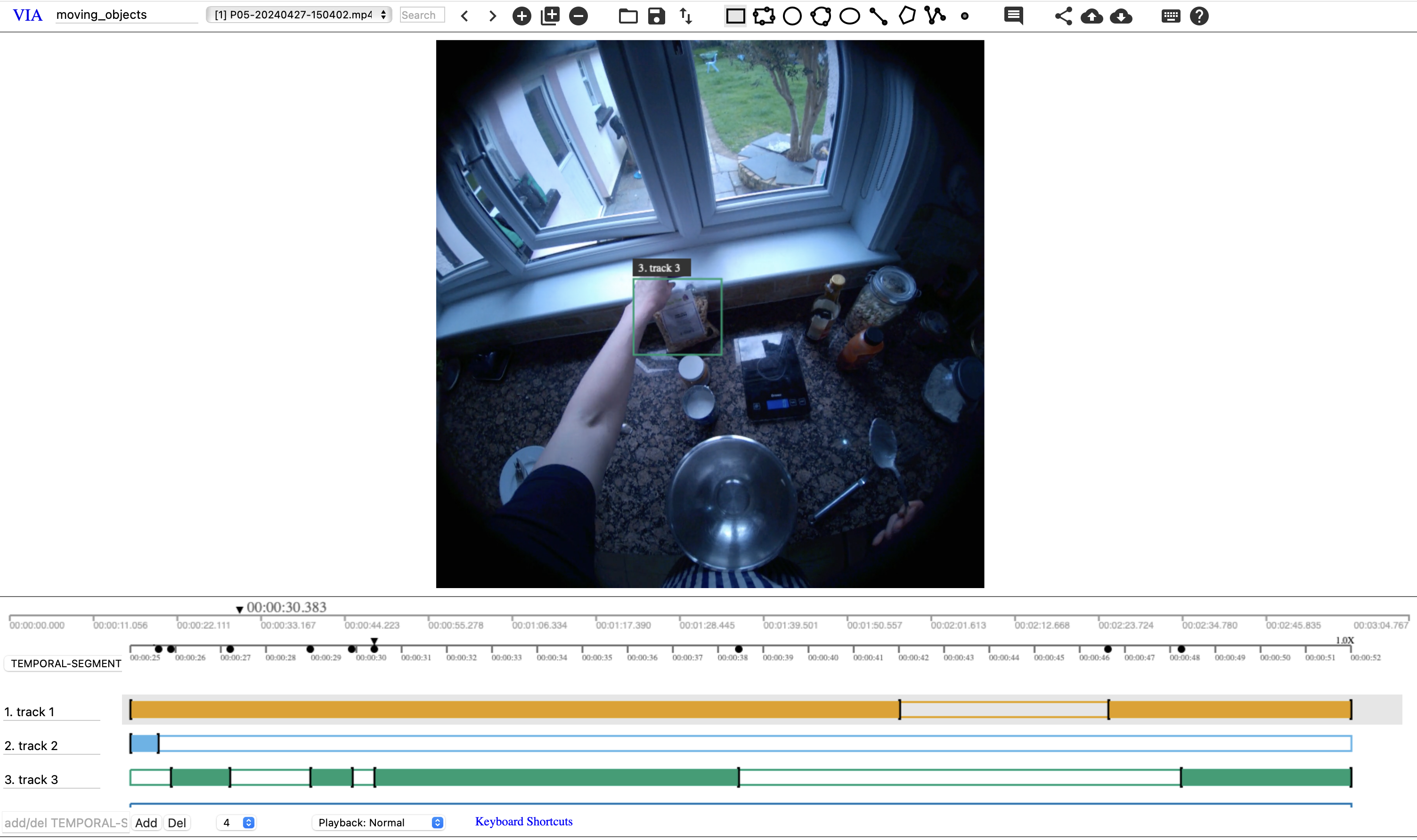}
    \caption{The VIA interface used for annotations object movements in 2D.}
    \label{fig:via_interface}
\end{figure}

\noindent{\textbf{Moving objects in 2D.}} For annotating the object moving segments, we used the VIA \cite{dutta2019vgg} interface as shown in \cref{fig:via_interface}. In addition to the start and end timestamps of the object motion, we ask the annotators to provide bounding boxes around the object at the onset and end of the motion.

\begin{figure}
    \centering
    \includegraphics[width=\linewidth]{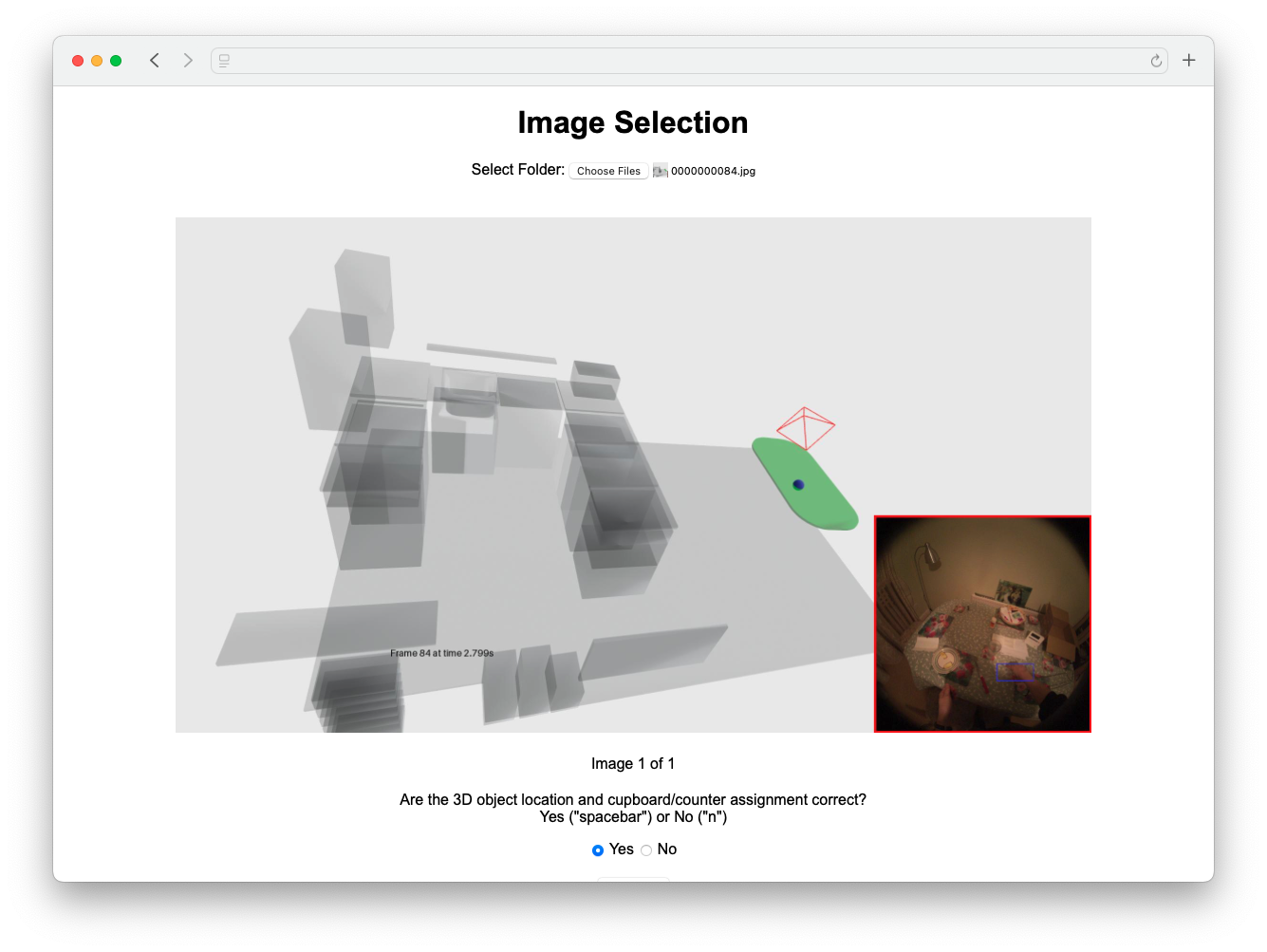}
    \caption{Interface for reviewing object/fixture assignments.}
    \label{fig:blender_review}
\end{figure}

\noindent{\textbf{Object/fixture review.}}
Fig. \ref{fig:blender_review} shows the interface for manually reviewing object/fixture assignments. Reviewers were asked to verify whether the 3D location and assigned fixture look correct. This process was repeated - after each iteration corrections to the Fixture annotations (\eg including missing bins, hooks \etc) were made in Blender, as well as adding heuristics to the assignment process.

\begin{figure}
    \includegraphics[height=0.4\linewidth]{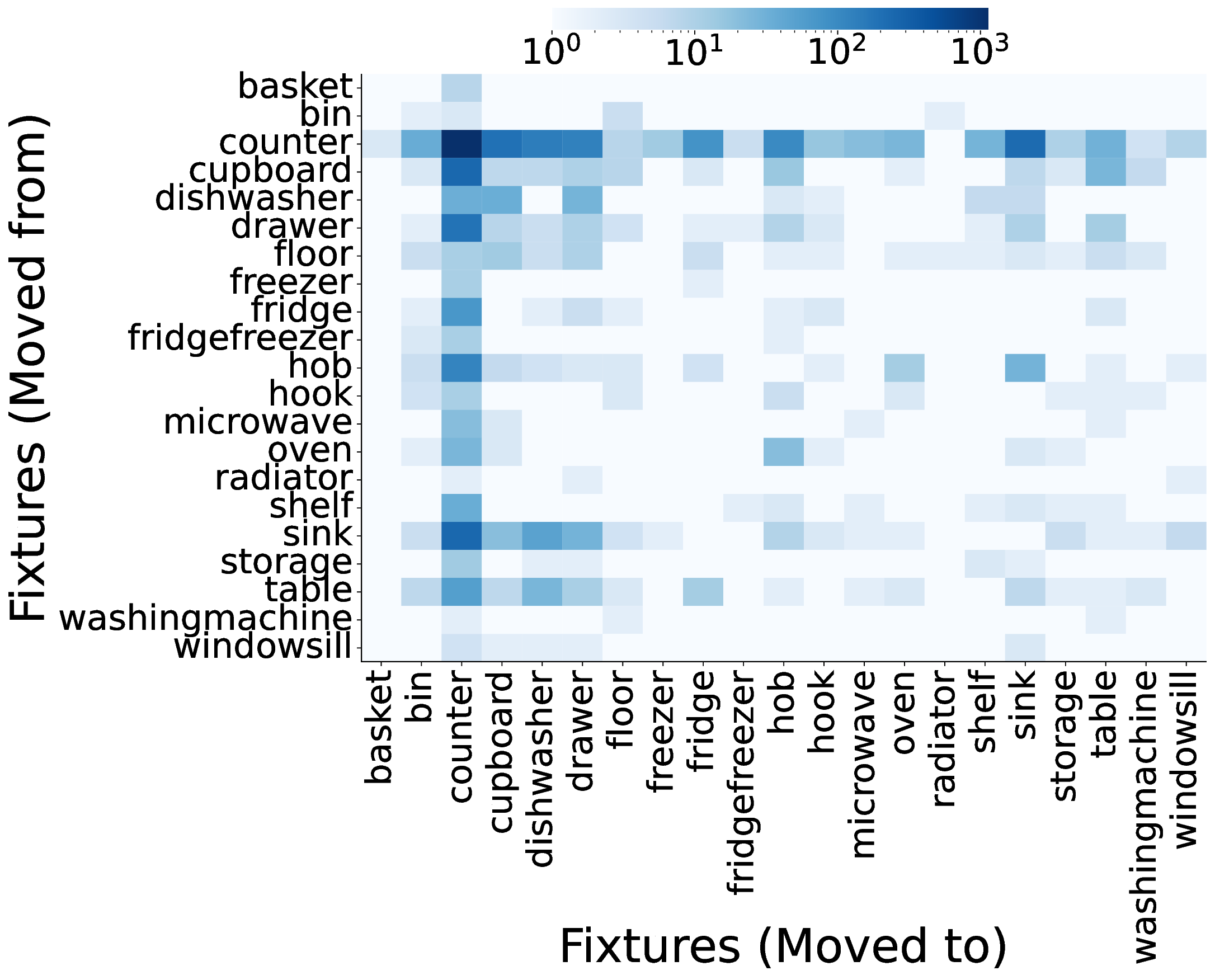}    
    \includegraphics[height=0.4\linewidth]{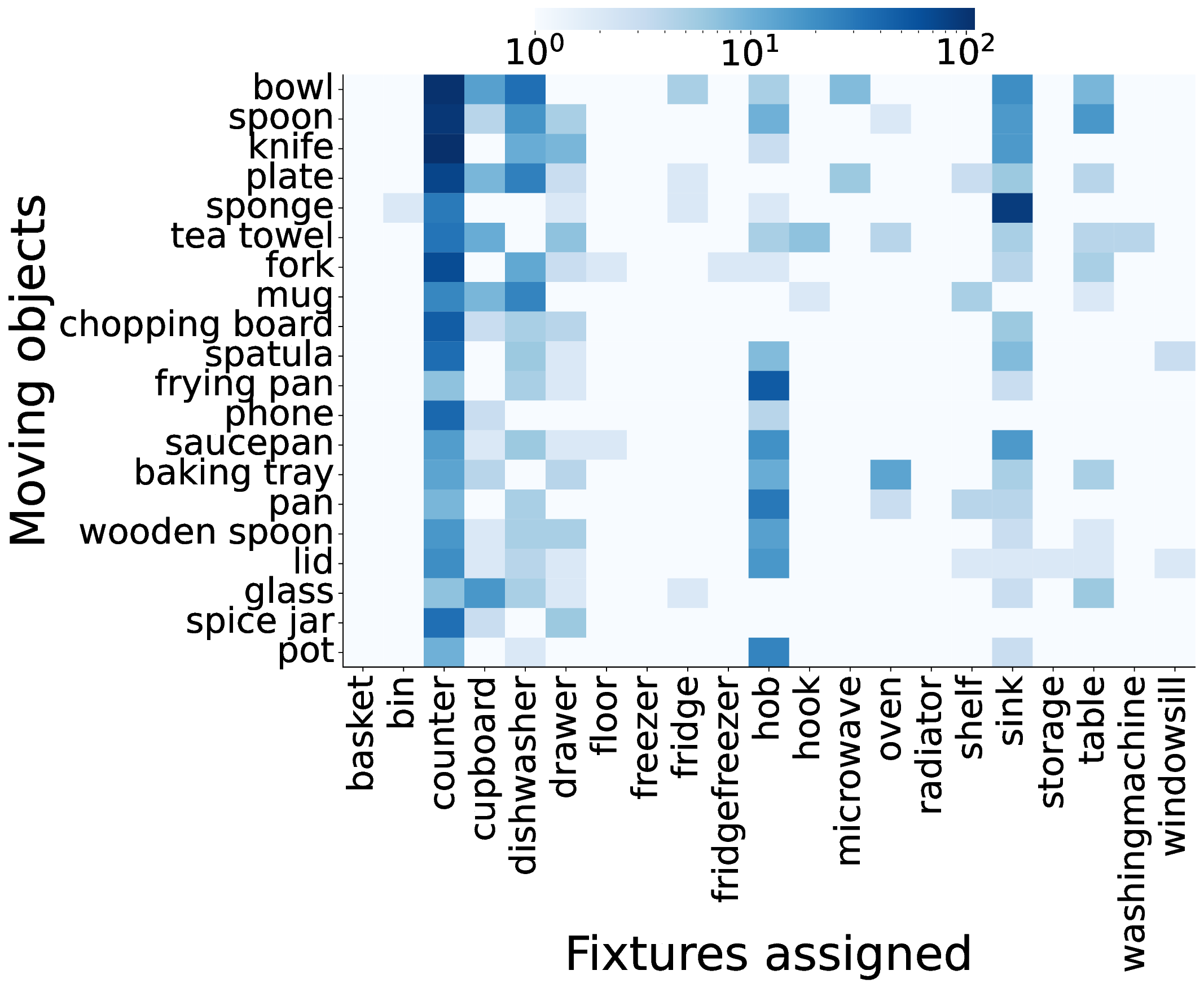}  
    \caption{Left: Common  source$\rightarrow$destination  object movements. Right: Common take/put locations of the top-20 moving objects.}
    \label{fig:fixture_transitions}
\end{figure}

\noindent{\textbf{Fixture transitions.}}
Frequent transitions are cross-fixtures (\ie excluding objects picked up and placed down on the same fixture), sink$\rightarrow$counter and counter$\rightarrow$cupboard.
Fig. \ref{fig:fixture_transitions} shows common transitions and locations where the 20 most common objects are placed on/in, normalised per object. Again, counters are the most used fixtures.

\begin{figure*}
    \centering
    \includegraphics[width=\textwidth]{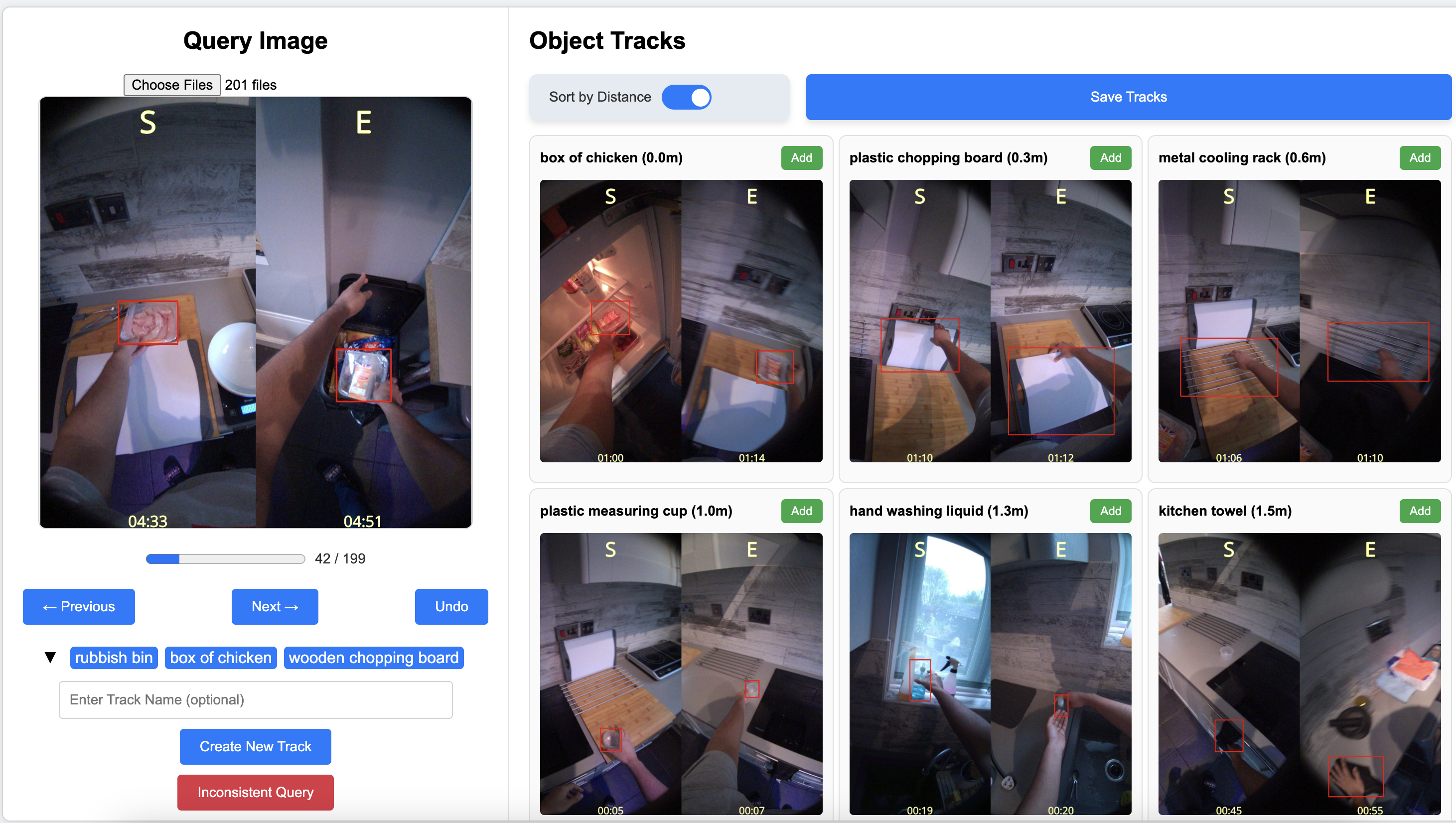}
    \caption{\textbf{Object associations interface}. For each query track, a set of possible object names extracted from the narrations is suggested, we provide these names as an additional annotations. Importantly, all previous tracks are sorted by their distance. It is trivial for the annotator to confirm that the track on the left, matches the box of chicken on the right, which has been positioned in exactly the same location (0.0m). By clicking `Add', this track is corerctly associated connecting the object movement over time.}
    \label{fig:object_associations}
\end{figure*}
\noindent{\textbf{Long Term Object Tracking.}}
We link object movements over time to provide long term tracking for all objects, \ie object itineraries. These capture sequences of an object's movement throughout the video.
Fig. \ref{fig:object_associations} shows the object associations interface used to connect objects throughout the video. It utilises our lifted 3D locations to speed up the annotation process. For each track, the annotation interface re-orders objects based on the comparison between the query track's initial 3D location (pick up) and previous tracks' final 3D location (put down).
This approach was used in~\cite{Plizzari2024} for egocentric tracking, and we use it as an initial step in our interface. 
Around 12\% of the objects moved at least 5 times in the video making it suitable for long-term tracking. On average, an object moves 2.4 ($\pm$ 2.6) times in HD-EPIC.

\noindent \textbf{Hand and Object Segmentations}
To improve performance, we first manually annotate a small set of images from each video (5-10 frames) for both left and right hands.
To cover the diverse locations and lighting conditions, we use the camera pose estimates %
to identify clusters of hotspots and select a set of images that represent hands in action at various locations.
This is sent to dedicated annotators who manually segment these.
We follow the same approach in~\cite{VISOR2022}, where hands and arms are segmented jointly to avoid the ad hoc decision of the hand boundary. We use the TORAS~\cite{torontoannotsuite} annotation tool for all our manual segmentation.
We manually annotate 
768 frames across 156 videos, which accounts for roughly $0.02\%$ of the total frames.

For each video, we utilise SAM2~\cite{ravi2024sam} to predict 2D hand masks for every frame. The manually annotated frames are placed in the conditional memory (roughly 5 frames per video) whilst the previous 6 predictions are used in the non-conditional memory in order to improve temporal consistency.
We notice that SAM2, like most Video Object Segmentation (VOS) models, suffers from identifying the hand late when it enters the scene. In order to address this, we run the aforementioned process both forwards and backwards through the video.

\noindent \textbf{Correcting Noisy Hand Segmentations.} 
We devise a simple, non-learned procedure to automatically detect noisy masks based on the rate of change of the mask's area over consecutive frames. 
A major change in size often indicates noise.
If the previous frame has no mask and the current mask is large, then it is likely that a frame has missed a mask or under-segmented.
We flag frames for noise correction.
Additionally, if the previous mask is reasonable but the current mask is much larger or smaller, then this is another indication of errors. We use a dynamic threshold which catches when masks expand or contract too quickly based on the area of the previous mask.

Once we have flagged noisy frames, we sample from these frames for manual annotations. 
In order to propagate the new knowledge from the manual correction phase, we re-run the SAM2 procedure within a small window of frames around the original error, with the newly corrected frame in the memory (alongside the original 5 ground-truth images). The idea here is to correct frames around the original noise with the new manually annotated frame as a conditioning frame. In total, this correction procedure produces roughly 7.5k right hand masks and roughly 7k left hand masks for manual correction ($0.19\%$ of total hand masks). As this is still a rather large amount of annotation to conduct, we further reduce this by uniform temporal sampling, achieving an average of roughly 20 hand side masks per video to be corrected. In total, we manually clean and annotate a further 5163 right hand masks and 4332 left hand masks ($0.12\%$ of total hand masks).

\subsection{Density of Our Annotations}
On average, we have 263 annotations per minute of the dataset. Here in \cref{tab:annotations_per_min}, we show the breakdown of this number. Note that this excludes the hand masks that we generate at 30 fps.
\begin{table}[]
\resizebox{\linewidth}{!}{
\begin{tabular}{c|cc}
\toprule
Annotation Type                                                                                           & Total annotations & Annotations/min \\ \midrule
Narrations                                                                                                & 59,454             & 24.0            \\
\begin{tabular}[c]{@{}c@{}}Parsing (Verbs + Nouns \\ + Hands + How + Why)\end{tabular}                    &      303,968             & 122.7           \\
Recipes (Preps + Steps) & 4,052 & 1.6 \\
Sound                                                                                                     & 50,968             & 20.6            \\
Action boundaries                                                                                         & 59,454             & 24.0            \\
\begin{tabular}[c]{@{}c@{}}Object Motion (Pick up + Put down \\ + Fixtures + Bboxes + Masks)\end{tabular} &     153,480              & 62.0          \\
Object Itinerary                                                                                          &  4,881                 & 2.0            \\
Object Priming (Starts + Ends)                                                                            &    18,264               & 7.4            \\ 
\midrule
Total & & 263.2\\
\bottomrule
\end{tabular}}
\caption{HD-EPIC annotations per minute}\label{tab:annotations_per_min}
\end{table}
\section{Benchmark and Results}
\label{sec:benchmark_supp}

We first detail how each of the question prototypes in the VQA benchmark are formed in Sec.~\ref{sec:vqa_bench_supp}. We then provide additional experiments for the VQA demonstrating an ablation of the inputs and the prediction bias of models in Sec.~\ref{sec:vqa_experiment_supp}. Sec.~\ref{sec:vqa_bench_models_supp} gives additional details of the models used in the VQA benchmark. Finally, we provide additional details of the long-term VOS benchmark and recognition benchmark in Sec.~\ref{sec:long_term_vos_supp} and Sec.~\ref{sec:benchmark-recognition_supp}. 

\subsection{VQA Benchmark Details}
\label{sec:vqa_bench_supp}

In this section, we detail how all question prototypes are formed and sampled.

\subsubsection{Recipes}
These questions test the ability to understand, retrieve, localize, and recognize recipes and their steps.

\noindent\textbf{Recipe Recognition. } 
To test a model's ability to recognise entire recipes from long-term video, we formulated the question prototype ``Which of these recipes were carried out by the participant?''
As input, the model receives every video recorded by a given participant, concatenated chronologically. 
The positive answer is the name of one of the recipes prepared in these videos. 
To sample difficult negative answers, we generated recipe names similar to the positive recipe via ChatGPT~\cite{achiam2023gpt}, with the prompt ``You are a professional chef. Your goal is to identify 4 similar recipes for each line of the text file. These recipes must technically be different dishes, they should not be different names for the same meal. The recipes can make use of the same ingredients or new ones.''
The results were manually edited to ensure validity, which allowed us to create more difficult negatives than were otherwise present in the dataset.

\noindent\textbf{Multi-Recipe Recognition. }
Extending this, the question prototype ``Which of these recipes were carried out in this video?'' tests a model's ability to recognise every recipe occurring within a video. 
As input, the model now receives an entire video containing one or more recipes. 
The positive answer is the names of all recipes occurring in the video, with the negatives being generated in the same way as in  \textbf{Recipe Recognition} above.

\noindent\textbf{Multi-Step Localisation. }
With the question prototype ``In this video, when did the participant perform each of the following \texttt{$<$recipe name$>$} recipe steps: ``\texttt{$<$recipe step a$>$}'', ``\texttt{$<$recipe step b$>$}'', ``\texttt{$<$recipe step c$>$}''?'', we then test a model's ability to localise a subset of time segments of $3$ steps of a recipe in a video.
One video is supplied as input and in the question text we supply the text for 3 recipe steps selected in a random order.
The positive answer is a list of the largest segment start and end times for the $3$ steps from a given recipe that occur within the video.
To generate negative answers, we first collect all the time segments for recipe steps that don't correspond to the recipe in the question text (if any exist).
Then we concatenate these with all the time segments for recipe prep occurring in the video.
Finally, for each negative answer we randomly sample $3$ segments.

\noindent\textbf{Step Localisation. }
Similarly, the question prototype ``When did the participant perform step \texttt{$<$recipe step$>$} from recipe \texttt{$<$recipe name$>$}?'' assesses a model's ability to localise all major segments corresponding to a recipe step.
Again, the input is one video which must contain  $\geq5$ unique recipe steps.
The positive answer is the list of all annotated start and end times for a randomly chosen recipe step.
In the question text, the name of the recipe and the recipe step text is provided.
To form negative answers, we randomly sample other recipe steps from the same video and use the start and end times of the corresponding time segments.

\noindent\textbf{Prep Localisation. }
Likewise, the question prototype ``When did the participant perform prep for \texttt{$<$recipe step$>$} from recipe \texttt{$<$recipe name$>$}?'' assesses a model's ability to localise all major segments corresponding to prep for a given recipe step.
Again, the input is one video which must contain  $\geq5$ unique recipe steps.
The positive answer is the list of all annotated start and end times for the prep relating to a randomly chosen recipe step.
In the question text, the name of the recipe and the recipe step text is provided.
To form negative answers, we randomly sample other recipe steps from the same video and use the start and end times of the corresponding prep time segments.

\noindent\textbf{Step Recognition. }
As recipes are comprised of many high-level steps, the question prototype ``What step did the participant do between \texttt{$<$TIME HH:SS:MM video 1$>$} and \texttt{$<$TIME HH:SS:MM video 1$>$} in this video?'' tests a model's ability to recognise these long activities.
One video is supplied as input, which must contain at least $5$ unique recipe steps.
The positive answer is randomly sampled from the steps occurring in the video.
In the question text, the provided start time is the start of longest segment relating to the positive step and the end time is the end of the longest segment.
To form negatives, we randomly sample recipe steps from the same video as the positive answer.

\noindent\textbf{Rough Step Localisation. }
Where the prior question prototypes consider the localisation of all segments of a recipe step/prep, ``Which of these time segments belongs to the \texttt{$<$recipe name$>$} recipe step \texttt{$<$recipe step$>$} in this video?'' tests whether a model can localise recipe steps from a subset of the possible time segments.
Again, the input is one video with $\geq5$ unique recipe steps.
Here, the positive answer is the start and end time of the longest time segment relating to the recipe step.
The negative answers are formed by randomly sampling recipe steps from the same video and using their corresponding start and end times of the longest segment.

\noindent\textbf{Following Activity Recognition. }
Finally, we test a model's ability to recognise high-level activities that follow recipe steps with the question prototype ``Which high-level activity did the participant do while completing recipe step \texttt{$<$recipe step$>$} in this video?''
As input, the model receives one video and the recipe step text in the question.
The positive answer is determined by finding the high-level activity that occurs while the given recipe step is concluding.
Negative answers are sampled from other activities occurring in the video that do not overlap with the end time of the given step.

All of these question prototypes were sampled uniformly at random from the space of possible positive answers. However, as some simpler recipes occur more frequently than others, these were excluded from the recipe recognition and step localisation tasks.

\subsubsection{Ingredients}
The ingredient questions assess a model's ability to understand and identify the ingredients used, their exact amounts and in what order they are added.

\noindent\textbf{Ingredient Retrieval.} 
We test a model's ability to recognise added ingredients, with the question prototype ``Between \texttt{$<$TIME HH:SS:MM video 1$>$} and \texttt{$<$TIME HH:SS:MM video 1$>$}, which of these ingredients were added to the dish being prepared?''
One video is used as input, with the start and end time of an ingredient-adding segment supplied in the question text.
Instead of using the exact annotated times, we pad the start and end times by adding $5$ seconds to either side of the segment.
If the new times overlap with the adding of a new ingredient, we instead add the smallest possible time that avoids the introduction of a new ingredient to the segment.
The positive answer is randomly selected from the ingredients added in the video. %
We generate negatives by randomly selecting from ingredients that are added in the video and do not overlap with the positive answer.

\noindent\textbf{Ingredient Weight. }
This question tests the ability to identify ingredient weight with the prototype ``How much did the participant weigh of \texttt{$<$ingredient$>$} in this video?''. The input is a video segment annotated as the weighing sequence for a given ingredient and the positive is the correct weight of this ingredient. It should be noted that humans can trivially perform this task by reading the scale while current models struggle. %
The negatives are generated by sampling a random multiplier between (0.5, 5) for the positive answer. %

\noindent\textbf{Ingredient Order. } This question tests the ability to identify the order in which ingredients are added with the prototype “What is the order of ingredients added to the dish in this video?”. The input is a video segment where at least three ingredients were added to the dish and the positive answer is the list of the added ingredients in the correct order. For recipes with more than five ingredients, a subset of five consecutive ingredients was sampled randomly. The negatives are random permutations of the positive order. %

\noindent\textbf{Ingredient Adding Localisation. } We assess the ability to temporally localise when ingredients are added with ``When was ingredient \texttt{$<$ingredient$>$} added to recipe \texttt{$<$recipe name$>$}?''. The input is an untrimmed video and positive is the start and end time of the annotated adding segment. Negative answers are the start and end time of  \textit{action} segments from the input video where the narration mentions the ingredient. %
Action segments overlapping with the adding time segment are discarded. 
If less than 4 action segments include the ingredient, we use action segments containing similar ingredients. Similarity is calculated using cosine distance $>0.5$ of spaCy~\cite{honnibal2017spacy} (model en\_core\_web\_lg) word embeddings. %

\noindent\textbf{Ingredient Recognition}
This tests the ability to identify the ingredients used with ``Which of these ingredients is used in \texttt{$<$recipe name$>$}?''. The input is all videos of one participant, with one of the completed recipes selected as \texttt{$<$recipe name$>$}. Only recipes with at least four ingredients were considered. %
The positive answer is obtained by randomly sampling a single ingredient from the ingredients used. The negatives were generated using an LLM which is asked to provide potential ingredients for a given recipe that do not include the ingredients used in the dish. This was done to ensure more challenging negatives than those sampled from other recipes in the dataset. 
We also created a negative version of this question: ``Which of these ingredients is not used in \texttt{$<$recipe name$>$}?'' was also created. Here, the input video selection is the same and positive and negative selection strategies are swapped. %

\noindent\textbf{Exact Ingredient Recognition}
This question tests the ability of a model to identify the exact quantity of an ingredient with ``What was the exact quantity of \texttt{$<$ingredient$>$} used in \texttt{$<$recipe name$>$}?''. For this question, the input is all videos relating to a randomly selected recipe. The positive answer is randomly selected from the used ingredients while excluding ingredients used in the \textbf{Ingredient Weight} question. The negatives are generated by sampling a random multiplier between (0.5, 5) applied to the positive answer. %

To select final questions for our benchmark we uniformly sample over participants.
\subsubsection{Nutrition}
The nutrition questions assess a model's ability to understand the nutritional values of ingredients used in the videos.

\noindent\textbf{Image Nutrition Estimation. } Our first question is ``Which of the ingredients in these images showcase higher \texttt{$<$nutrition$>$?}'', where \texttt{$<$nutrition$>$} is one of ``calories'', ``fat'', ``carbs'' or ``protein''. The input is 5 video frames showing different ingredients. To obtain these we manually label a frame clearly showing the ingredient. We search this frame within the %
ingredient-adding sequence as this allows a clear view of the correct quantity of the ingredient. In total we annotate 200 frames/ingredients, from which we generate 834 questions.
The positive is a randomly selected ingredient. To sample the 4 negatives and their corresponding frame, we first calculate the value-to-amount ratio of each ingredient, \eg if we have 30g of protein in a 100g ingredient, the ratio is 0.3. %
For a nutritional value we form negatives by finding ingredients that have a \textit{higher} value-to-amount ratio but a \textit{lower} nutritional value compared to the positive answer. This strategy samples hard negatives that are not answerable by text alone. %
To ensure a reasonable separation between positive and negative answers, the difference in nutritional value is at least 20\% of the positive answer's value. %

\noindent \textbf{Nutrition Change. } We use ``From \texttt{$<$TIME HH:SS:MM video 1$>$} to \texttt{$<$TIME HH:SS:MM video 1$>$}, what changed in the nutrition values of the dish with recipe \texttt{$<$recipe name$>$}?''. The input is a video containing a recipe (\texttt{$<$recipe name$>$}) and one or more adding segments. \texttt{$<$TIME HH:SS:MM video 1$>$} placeholders are replaced with the annotated start-end times of an ingredient-adding segment. Answers are in the format ``calories changed by \texttt{$<$a$>$}, fat changed by \texttt{$<$b$>$}, carbs changed by \texttt{$<$c$>$}, protein changed by \texttt{$<$d$>$}''. For the positive  \texttt{$<$a/b/c/d$>$} are replaced with the correct ingredient nutritional values. For the negatives, the placeholders are multiples of the correct nutritional value. The random multiplier is sampled within $(0.10, 0.75)$ or $(1.25, 1.75)$.

\noindent\textbf{Video Nutrition Estimation. } Finally, we also use the prototype ``What is the ingredient with the highest \texttt{$<$nutrition$>$} in this recipe?'', where \texttt{$<$nutrition$>$} is the same as above. For this question, the input is an untrimmed video containing one recipe. The positive answer is the ingredient with the highest calories/fat/carbs/protein in the recipe. Negative answers are sampled from ingredients that have a value-to-amount ratio similar to the positive. At least 2 negative answers ingredients come from the same recipe, while the rest can be sampled from other recipes when it is not possible to obtain enough negative answers from a single recipe. %

Final benchmark questions are randomly sampled from the set of possible questions uniformly across all participants to ensure diversity.

\subsubsection{Fine-Grained Action}
With these questions, we assess a model's ability to understand detailed fine-grained actions as described by the participants in our ground-truth narrations.

\noindent\textbf{Action Recognition. } We first formulate the question prototype ``Which of these sentences best describe the ongoing action(s) in the video 1?''. The input is a short video clip containing 1-3 actions. Positives contain the correct 1-3 action descriptions obtained from the narration transcriptions. We sample negatives by selecting other sequences of the same number of actions from other video clips in the dataset. We ensure difficult narrations by requiring negatives to have at least one verb and noun in common with the positive and prioritise those with more common verbs and nouns.

\noindent\textbf{How Recognition. } 
To assess the model's ability to understand how fine-grained actions are performed, we use the prototype ``What is the best description for how the person carried out the action \texttt{$<$verb noun$>$} in this video segment?". We randomly sample a narration annotated with start and end times. The input is a video clip of this action as given by the start-end time annotations and \texttt{$<$verb noun$>$} in the question is the main action extracted from the narration by our parsing. The positive answer is the manner of the target action as provided by the ground-truth narration also extracted by the parsing.
To sample negative answers, we first identify other narrations that share both the same verb and noun cluster (see Sec.~\ref{subsec:narrations_supp}) with the main action of the target narration. We then extract `how' from these narrations using our parsing. To avoid false positives, we filter out any `how' with verb and noun clusters which are subsets or supersets of the positive answer and manually filter the remaining questions.

\noindent\textbf{Why Recognition. } 
To assess the model's ability to understand why fine-grained actions are performed, we use ``What is the best description for why the person performed the action  \texttt{$<$verb noun$>$} in video 1?". The input is a video clip for a randomly sampled narration and \texttt{$<$verb noun$>$} is the main action parsed from the narration. The positive answer contains the reason for performing the target action as extracted from the narration by our parsing.
To sample negatives, we use `whys' extracted from other narrations with the same main action for at least 2 negatives. When there are not enough of these negatives we add whys from actions that share the same verb. To avoid false positives, we remove any negative `why' with verb and noun clusters which are subsets or supersets of the positive answer and manually filter the remaining questions. %

\noindent\textbf{Action Localisation. } 
To assess ability to localize fine-grained actions, we created the question prototype ``When did the action \texttt{$<$verb noun$>$} happen in the video 1 ?". The input is a 30 second to 15 minute video clip. Answers are in time period format (e.g., \texttt{$<$TIME HH:MM:SS.MS$>$} to \texttt{$<$TIME HH:MM:SS.MS$>$}). The positive answer corresponds to the time period when a randomly sampled target action occurs, while the negative answers are time periods when different actions occur. To make the questions challenging, at least two of the negative timestamps are from actions that share the same noun as the target action but differ in the verb. To ensure 4 negatives we add timestamps for actions that share the same verb but have a different noun, or those that share either noun or verb cluster.

\subsubsection{3D Perception} \label{sec:3d_perception}
The 3D perception questions assess the model's capability to understand 3D environment or 3D movements from only video. 

\noindent\textbf{Fixture Location.}
These questions test the ability to understand a fixture's relative location in 3D. We use the fixture mesh annotations for each kitchen to generate questions of type ``Given the direction I am looking at \texttt{$<$TIME HH:MM:SS video 1$>$}, where is the \texttt{$<$query fixture$>$} located?" We select the \texttt{$<$query fixture$>$} as one of the unique fixtures in the kitchen \eg fridge, hob, sink, microwave. 
For the positive answer, we calculate the angle between the direction where the camera is pointed and the direction of the \texttt{$<$query fixture$>$} from the camera location. We represent this angle as position of the hour hand on the clock \eg 2 o'clock, 7 o'clock. For the negatives, we randomly sample from other possible positions of the hour hand on the clock. We don't choose negatives with an absolute $\leq 1$ o'clock difference to the positive, in case it might lead to confusion, %
\eg if the positive is 2 'o clock, then the negatives can be sampled from 4 'o clock to 12 'o clock. For the timestamp in the question, we ignore frames that have a tilt greater than $45^{\circ}$ in the camera because it would lead to incorrect calculation of the positive.  %

\noindent\textbf{Object Location.}
Here we test the model's ability to understand object location as objects move %
with the prototype ``Where did I take/put the object identified by \texttt{$<$BBOX Y1 X1 Y2 X2$>$} at \texttt{$<$TIME HH:MM:SS video 1>} from before/after putting/taking it at \texttt{<TIME HH:MM:SS>}?" The bounding box and timestamp in the question is directly taken from the annotations of moving objects in 2D (Sec.~\ref{sec:digital_twin_supp}). If the question asks where the object is put down, the bounding box and timestamp are when the object is picked up before that. If the question asks where the object is taken from, the box and timestamp are from when the object is put down after that. %
For the positives, we use the name of the correct fixture associated with the object bounding boxes at the time of pick up/put down. For the negatives, we sample different fixtures that other moving objects have been associated with in the video. We use relative location to another unique fixture \eg sink to identify fixtures without a unique name \eg the cupboard top left of the sink. 

\noindent\textbf{Object Contents Retrieval.}
Here we test the model's ability to understand 3D locations of fixtures as well as recognise objects that were put/taken from the fixture. We use the fixture meshes to generate questions of type ``Which of these objects did the person take/put from/in/on the item indicated by bounding box \texttt{$<$BBOX Y1 X1 Y2 X2$>$} in \texttt{$<$TIME HH:MM:SS video 1$>$}?" We randomly sample a fixture in the kitchen and project the 3D vertices of its mesh onto the image plane. We use the maximum and minimum of the projected vertices to find the 2D bounding box. We repeat this for all frames in the video and search for frames where the sampled fixture is occluded less than 30\% and the projected bounding box lies fully within the image. We then choose one of the identified frames and the corresponding bounding box for the question. We use the long-term object track and fixture transition annotations \cref{sec:digital_twin_supp} to identify objects that were either taken from or put on/in the fixture in question. We treat these objects as positive objects and sample few of them as our positive answer. For negatives, we sample objects that were taken from or put on the same fixture in different videos but do not overlap with the positives. We also sample objects that were assigned to other fixtures as negatives. 

\noindent\textbf{Fixture Interaction Counting.} 
Here we assess the model's ability to remember the interactions between participants and fixtures with the question %
``How many times did I open/close the item indicated by bounding box \texttt{<BBOX Y1 X1 Y2 X2>} at \texttt{<TIME HH:MM:SS video 1>}?" The timestamp and bounding box for the query fixture is obtained following the same procedure as for \textbf{Object Retrieval} questions. For the positive, we use the correct number of times the fixture was opened or closed. We obtain this by first using the parsed narrations to identify phrases relevant to fixture interactions, \ie where the verb includes `open' or `close' and the noun includes one of the fixtures. For relevant narrations, we use a 1s temporal window before the corresponding narration timestamp to find the fixture with the highest cumulative gaze intersection. This identifies the exact fixture that is being interacted with at that time.  The negatives are sampled randomly from $\pm 4$ of the positive.

\subsubsection{Object Motion}
Here, we test the model's ability to correctly reason over various moving objects given a full video as input. We use the long-term object tracking annotations mentioned in Sec. \ref{sec:digital_twin_supp} for questions in this category.

\noindent\textbf{Object Movement Itinerary.} %
We design the question as ``Where was the object \texttt{<BBOX X1 Y1 X2 Y2>} seen at time \texttt{<TIME HH:MM:SS video 1>} moved from/to throughout the video?". %
We provide bounding boxes around the object instead of giving the object name as text. %
As positives, we selected the correct order of the fixture names involved as either source or destination in all trajectories of the highlighted object in the query video \eg ``from fridge to sink, then from sink to microwave". As negatives, we select the fixtures associated with the itineraries of other objects that are moving in the video. For fixtures that do not have a unique name \eg cupboards, we follow the same procedure as \textbf{Object Location} and generate descriptions of the relative position.

\noindent\textbf{Object Movement Counting.} We begin by testing the ability to count how many times an object moves with the prototype ``How many times did the object \texttt{<BBOX Y1 X1 Y2 X2>} seen at \texttt{<TIME HH:MM:SS video 1>} change locations in the video?" The input is the full video, and the object is specified by its bounding box at a particular timestamp from the moving objects in 2D annotations (Sec. \ref{sec:digital_twin_supp}).
Positive answers are derived by counting distinct movements using the long-term object tracking annotations in Sec. \ref{sec:digital_twin_supp}.  To mitigate rapid movements and erroneous annotations, we retain only objects satisfying two criteria: (1) consecutive tracks must be separated by at least 1.1 seconds, and (2) the spatial displacement between the end of one track and the start of the next must not exceed 20 cm.  For negative answers, we randomly sample values around the positive answer. To avoid text-based shortcuts, the sampling window is varied randomly from [1,8], and the maximum negative value is restricted to 8. We additionally enforce that the negative values for cases having more than 2 movements should be at least $\pm2$ value apart from the ground truth answer.

\noindent\textbf{Stationary Object Localization.} Here, we evaluate the capability to perform the reverse task: determining when an object remains static in the video. The question is: "After the object \texttt{<BBOX Y1 X1 Y2 X2>} seen at \texttt{<TIME HH:MM:SS
video 1>} is first moved, from which of the following starting times does the object remain static for more than \texttt{<X>} seconds?". Similar to the previous task, the input is the entire video, and the object is specified by its bounding box at a given timestamp. The static duration is determined using long-term object tracking annotations (Sec. \ref{sec:digital_twin_supp}).
We find $t_1$ and $t_2$, the two longest static durations of an object. A value is then randomly sampled from the range $(t_2+5, t_1)$ to fill the placeholder \texttt{<X>}, which specifies the required static duration for the question. %
We also use the criteria mentioned in the previous question to filter objects with rapid movements and spatial inconsistency across tasks.
For the positive answer, a timestamp is randomly selected from the interval starting at the beginning of the maximum static duration and ending at a point determined by subtracting the previously chosen static duration from its endpoint. Negative answers are randomly sampled from the video from the start of the video to the last movement of the object, ensuring no overlap with the positive. All answers are provided as timestamps formatted as \texttt{<TIME HH:MM:SS video 1>}.

\subsubsection{Gaze}
The Gaze questions evaluate the model's ability to understand the subject of the camera wearer's visual attention, which can also be a signifier for future interactions. 

\noindent\textbf{Gaze Estimation.} We assess the model's capabilities to understand where the camera wearer is fixating their gaze within a given video clip. 
We formulate this question as ``What is the person looking at in this video segment?''. The input is a trimmed video clip where the camera wearer fixates on some large object, \ie a landmark, for at least 0.5 seconds. 
Positives are the name of the landmark being fixated on, negatives were randomly sampled from all other landmarks visible in the video segment.
To determine if a landmark is visible, we consider all 8 vertices of the 3D bounding box, as well as the centre of the object. 
An object is considered visible if, in any frame, 5 of the 9 points pass three checks: 1) the point projects onto the camera plane, 2) the point is in front of the camera and 3) the ray from the point to the camera centre does not intersect with another landmark's bounding box \ie it is not occluded. Common objects are given unique names according to relative positions similar to \textbf{Object Location}.
Overall, there are 1220 possible questions, of which we randomly sample 1000.

\noindent\textbf{Interaction Anticipation.} We also evaluate how effectively a model can anticipate the next object to be interacted with. 
We format the question as ``What object will the person interact with next, ignoring ongoing interactions?'' and provide a 10 second video segment, concluding 0.3 seconds after an object is primed for pick-up using eye gaze.
For these questions, the positive answer is the object which has just been primed, whilst the negatives are randomly sampled from all other objects that have been moved within the 2 minute video segment. We generate 1110 total questions and randomly select 1000.

\subsection{Additional VQA Experiments}
\label{sec:vqa_experiment_supp}

\begin{table}[]
    \centering
    \setlength{\tabcolsep}{1pt}
    \resizebox{\linewidth}{!}{\begin{tabular}{lcccccccc}
    \toprule
    Input & \recipe{Recipe} & \ingredient{Ingredient} & \nutrition{Nutrition} & \fine{Action} & \threed{3D} & \object{Motion} & \gaze{Gaze} & Avg.\\
\midrule
\rowcolor{LightGrey} \multicolumn{9}{l}{\textbf{Llama 3.2}} \\
 A only & 26.8 & 23.8 & 14.0 & 20.2 & 14.9 & 15.4 & 17.8 & 19.0\\
 Q + A & 33.5 & 25.0 & 36.7 & 23.3 & 22.3 & 25.3 & 19.5 & 26.5\\
 GT Narrations + Q + A & 70.8 & 46.3 & 34.0 & 62.5 & 42.9 & 28.7 & 29.4 & 45.0 \\
\rowcolor{LightGrey} \multicolumn{9}{l}{\textbf{Gemini Pro}} \\
 A only & 29.6 & 21.0 & 17.7 & 19.2 & 18.9 & 16.3 & 18.0 & 20.1\\
 Q + A & 38.0 & 26.8 & 30.0 & 22.1 & 21.5 & 27.7 & 20.5 & 26.7\\
 GT Actions + Q + A & 79.0 & 54.8 & 36.3 & 31.3 & 42.5 & 32.8 & 25.5 & 43.2\\
 GT Narrations + Q + A & 82.6
& 57.5 & 36.7 & 63.6 & 47.6 & 38.5 & 29.0 & 50.8\\
 Video + Q + A & 60.5
& 46.2 & 34.7 & 39.6 & 32.5 & 20.8 & 28.7 & 38.5\\
       \bottomrule
   \end{tabular}}
   \caption{\textbf{VQA Input Ablation} Our benchmark cannot be solved by analysing Q+A pairs or external knowledge and is a challenge for state-of-the-art closed and open source video VLM models.}
    \label{tab:result_model_inputs}
\end{table}

\noindent\textbf{VQA Input Ablation}. To explore how much information is contained in different inputs (video, ground-truth narrations, questions and answers) we ablate them in Tab. \ref{tab:result_model_inputs}. For both Llama and Gemini, providing only the possible answers gives random performance. 
Recipe and ingredient prototypes are the exception, likely due to textual recipes within the training data. Giving the question improves performance slightly as question contain contextual information, \eg \fine{Why Recognition} provides the action corresponding to the possible reasons why.  

Making predictions from the full ground-truth (GT) narrations is better than using  GT action clusters for all categories demonstrating the benefit of more detailed short-term understanding. Using GT narrations instead of visual input gives better performance for all question categories, highlighting the challenge of visual understanding in \DName. 
Having GT narrations helps most with recipe and fine-grained action and object motion questions but far from solves these categories \eg fine-grained action is 63.6\%. Regardless, nutrition, 3D perception, object motion, and gaze still have low performance with GT narrations demonstrating their difficulty.

\begin{figure*}
    \centering
    \includegraphics[width=\textwidth]{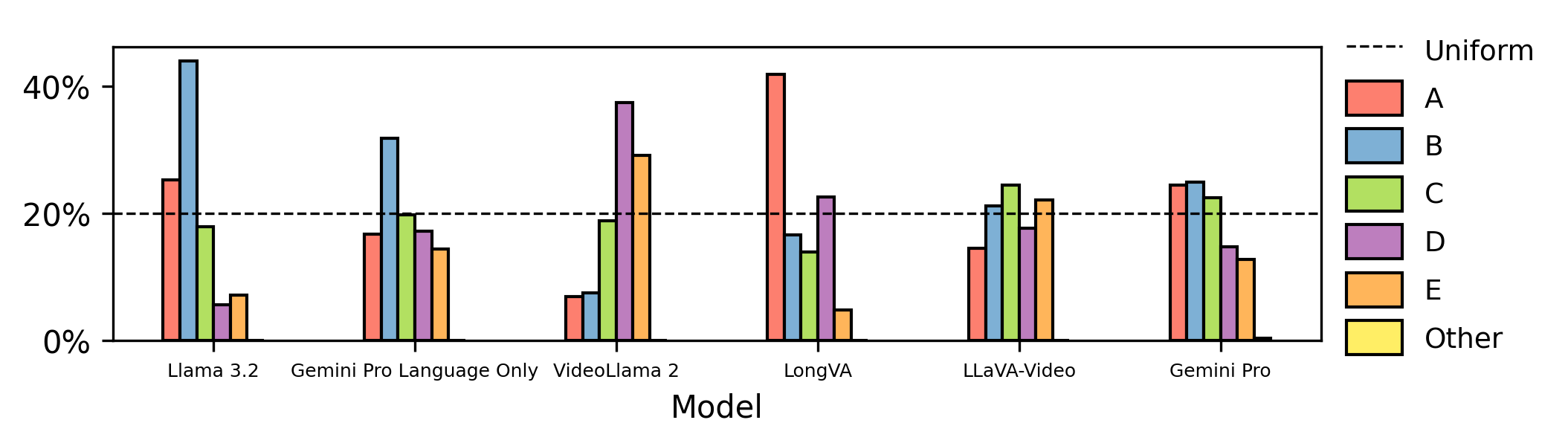}
    \caption{\textbf{Prediction Bias of Models}. Most models have a  bias in answer, although it is different for each model.}
    \label{fig:prediction_dist}
\end{figure*}

\noindent\textbf{Prediction Bias of Models}. Fig.~\ref{fig:prediction_dist} shows the distribution of output predictions for the 5 models tested. All models were able constrain their answers to the 5 options in almost all cases. 
Some models have clear significant bias (e.g. Llama 3.2 for option B and LongVA for option A).

\noindent\textbf{Per Prototype Results} In Table~\ref{tab:full_res} we show the numerical values for the per-prototype results shown in Fig.~\ref{fig:results_per_prototype}. 

\subsection{VQA Model Details}
\label{sec:vqa_bench_models_supp}
\noindent\textbf{Gemini Pro \cite{team2024gemini}.} 
Version ``gemini-1.5-pro-001'' was used as it is the latest version which supports context caching - necessary for asking repeated questions about long videos to keep costs tractable. Gemini processes all videos at 1fps regardless of length, so we passed all videos at 768 x 768 at their default speed up to 6000s long, after which we re-encoded the videos at a higher speed to fit in the model's 2M context window. In these cases we scaled all timestamps in questions to match this new speed to maintain the 1fps/time dependency noted in the documentation.
The following prompt was used:

\emph{You are an expert video analyzer, and your job is to answer the multiple choice question by giving only the letter identifying the answer. Do not give any other information. For example, acceptable answers are `A' or `B' or `C' etc. You must give an answer, even if you are not sure. Bounding boxes are in the format (ymin, xmin, ymax, xmax) relative to an image size of 1000x1000.}

0.2\% of questions were blocked/refused by the API due to safety reasons. These were counted as failure cases.

\noindent\textbf{Llama.} We used version ``Llama-3.2-90B-Vision-Instruct'' due to it being the largest and most up-to-date model we could run inference on. The model could run a majority of the questions, though occasionally when including the narrations, we had to evenly sample the input to be a maximum of 120K input tokens. The system prompt matched Gemini Pro.

\noindent\textbf{LongVA \cite{zhang2024long}.} Version ``LongVA-7B-DPO'' was used. We used the default resolution of 384x284. Videos were processed at 8fps up to a maximum context window of 400 frames (as large as would fit in VRAM with the model). After this, frames were uniformly sampled. The same prompt was used as for Gemini Pro. 

\noindent\textbf{VideoLLaMA 2 \cite{cheng2024videollama}.} Version ``VideoLLaMA2-7B-16F'' was used. Videos were processed at 336x336 resolution with timestamps scaled similarly to Gemini Pro, and with the maximum supported context window of 16 frames. The same prompt was used as for Gemini Pro. In the case of multi-video input, if input exceeded the frame limit and one or more videos were not used, the prompt included: 
\emph{\texttt{N} input videos have been concatenated together and the remaining videos have been truncated due to input length limit.}

\noindent\textbf{LLaVA-Video~\cite{zhang2024videoinstructiontuningsynthetic}.} Version ``LLaVA-Video-7B-Qwen2'' was used. Videos were processed the same as LongVA, but with the maximum number frames set to 128.

\noindent\textbf{Sample Human Baseline}. We also provide a sample human baseline to further assess the gap in understanding between humans and SOTA video-language models. For this we sampled 20 questions per question prototype, 600 questions in total, split between 3 participants.

\subsection{Long-Term VOS Benchmark}
\label{sec:long_term_vos_supp}

\begin{table}[]
    \centering
    \resizebox{\linewidth}{!}{\begin{tabular}{lrrrrr}
    \toprule
    \multirow{2}{*}{Dataset}        & \multirow{2}{*}{Sequences} & Avg. & Total & \multirow{2}{*}{Objs/Seq} & Annotated \\
    & & Duration & Duration & & Frames\\
    \toprule
    DAVIS~\cite{Pont-Tuset_arXiv_2017}  & 30        & 2.8s          & 83.3s         &  2.0        & 1,999        \\
    YouTube-VOS~\cite{vos2019} & 507 & 21.9s & 11,097.0s & 2.1 & 13,710 \\
    Ours & 1000 & 561.0s & 561,034.1s & 6.3 & 20,548 \\
    \bottomrule
\end{tabular}}
\caption{Our HD-EPIC long term VOS benchmark compared to the validation sets of DAVIS and YouTubeVOS, where duration is measured in seconds and calculated using the first and last frame indices of the annotated frames.}
\label{tab:long_term_vos_stats_supp}
\end{table}

The details of our long-term VOS benchmark are compared against the validation sets of two popular VOS benchmarks, DAVIS 2017 \cite{Pont-Tuset_arXiv_2017} and YouTube-VOS 2019 \cite{vos2019}. 
Our benchmark includes 1000 sequences, with an average duration of 561.0 seconds. This is calculated using the first and last frame indices of the annotated frames. When calculating durations for the DAVIS and YouTube-VOS we assume FPS values of 24 and 6, respectively, as stated in the original papers \cite{perazzi2016benchmark, xu2018youtube}. It is clear that our benchmark provides a much longer temporal duration for video object segmentation evaluation. Furthermore, the table shows that the average number of objects in each sequence is nearly tripled compared to the prior benchmarks.

We evaluate three models on our long-term VOS benchmark (Static, SAM2 \cite{ravi2024sam} and Cutie \cite{cheng2024putting}) and describe the results in the table of Fig. \ref{fig:vos_benchmark}. For SAM2 and Cutie, we insert each object's first appearance frame into the working memory of the model and use the remaining frames as evaluation frames. For the method labelled Static, we copy the mask from the first appearance to all evaluation frames, to act as a naive baseline. Following \cite{perazzi2016benchmark}, we use jaccard index $\mathcal{J}$, contour accuracy $\mathcal{F}$ and their average $\mathcal{J}\&\mathcal{F}$.

Due to the large number of frames in our dataset, we only pass the memory and evaluation frames into the model. We acknowledge that this is limited and the results could be improved further by sampling more frames.

\subsection{Recognition Benchmarks}
\label{sec:benchmark-recognition_supp}
Both action and sound recognition remain fundamental downstream tasks for video models.
We thus evaluate strong video models for both tasks in this section. 
their performance in the following. Note that for all works we follow the data (frames/audio) preprocessing as described in the original papers.

\noindent \textbf{Action Recognition.} We assess 5  action recognition methods~\cite{Feichtenhofer_2019_ICCV,patrick2021keeping,girdhar2022omnivore,tong2022videomae,Chalk_2024_CVPR}, using publicly available weights fine-tuned on EPIC-KITCHENS-100.
We detail the models and 
For VideoMAE-L, we used the finetuned model from~\cite{Chalk_2024_CVPR}.
As customary, we perform test augmentations over 10 clips during inference. %
The exception of this is TIM which also includes the audio modality, and averages the predictions are averaged across all context windows in which the action appears.
We also include a chance baseline, where we randomly shuffle the ground-truth labels and compute the accuracy, giving a lower-bound for the action recognition challenge.

\newcommand*\rot{\rotatebox{90}}
\afterpage{%
    \clearpage%
    \thispagestyle{empty}%
    \begin{landscape}%
        \centering %
    \centering
    \setlength{\tabcolsep}{3pt}
    \resizebox{\linewidth}{!}{
    \begin{tabular}{lrrrrrrrrrrrrrrrrrrrrrrrrrrrrrr}
    \toprule
         & \rot{\recipe{Recipe Recognition}} & \rot{\recipe{Multi-Recipe Recognition}} &  \rot{\recipe{Multi-Step Localisation}} & \rot{\recipe{Step Localisation}} & \rot{\recipe{Prep Localisation}} & \rot{\recipe{Step Recognition}} & \rot{\recipe{Rough Step Localisation}} & \rot{\recipe{Following Activity Recognition}} & \rot{\ingredient{Ingredient Retrieval}} & \rot{\ingredient{Ingredient Weight}} & \rot{\ingredient{Ingredients Order}} & \rot{\ingredient{Ingredient Adding Localisation}} & \rot{\ingredient{Ingredient Recognition}} & \rot{\ingredient{Exact Ingredient Recognition}} & \rot{\nutrition{Image Nutrition Estimation}} & \rot{\nutrition{Nutrition Change}} & \rot{\nutrition{Video Nutrition Estimation}} & \rot{\fine{Action Recognition}} & \rot{\fine{How Recognition}} & \rot{\fine{Why Recognition}} & \rot{\fine{Action Localisation}} & \rot{\threed{Fixture Location}} & \rot{\threed{Object Location}} & \rot{\threed{Object Contents Retrieval}} & \rot{\threed{Fixture Interaction Counting}} & \rot{\object{Object Movement Itinerary}} & \rot{\object{Object Movement Counting}} & \rot{\object{Stationary Object Localisation}} & \rot{\gaze{Gaze Estimation}} & \rot{\gaze{Interaction Anticipation}}\\
         \midrule
         \rowcolor{LightGrey} \multicolumn{31}{l}{\textbf{Blind - Language Only}} \\
 Llama 3.2 & 38.0 & 24.0 & 44.0 & 16.0 & 23.0 & 23.0 & 22.0 & \textbf{78.0} & 18.0 & 12.0 & 32.0 & 28.0 & \textbf{40.0} & 20.0 & 22.0 & 20.0 & \textbf{68.0} & 20.1 & 30.8 & 21.8 & 20.6 & 21.4 & 25.4 & 11.5 & 31.0 & 8.6 & 35.5 & 32.5 & 17.3 & \textbf{21.7} \\
  Gemini Pro & \textbf{54.0}& 54.0&42.0&20.0&21.0&15.0&22.0&76.0&18.0&28.0&38.0&21.0&28.0&28.0&26.0&14.0&50.0&21.3&24.6&21.6&20.7&20.4&33.4&9.0&23.0&\textbf{31.2}&18.5&\textbf{33.5}&21.2&19.8\\
\rowcolor{LightGrey} \multicolumn{31}{l}{\textbf{Video-Language}} \\
 VideoLlama 2 & 22.0&
52.0&
18.0&
38.0&
13.0&
18.0&
21.0&
64.0&
19.0&
30.0&
20.0&
27.0&
26.0&
\textbf{32.0}&
24.0&
20.0&
54.0&
30.9&
25.2&
32.2&
20.7&
18.8&
31.0&
35.5&
17.7&
11.0&
\textbf{44.0}&
30.5&
30.0&
12.4
\\
  LongVA & 14.0&
44.0&
36.0&
18.0&
18.0&
26.0&
19.0&
62.0&
25.0&
24.0&
44.0&
42.0&
30.0&
20.0&
25.0&
22.0&
54.0&
36.9&
28.4&
37.0&
20.5&
\textbf{26.6}&
\textbf{41.2}&
31.5&
32.3&
10.2&
34.5&
23.5&
36.0&
13.0
\\
  LLaVA-Video &  28.0&
68.0&
44.0&
20.0&
21.0&
23.0&
24.0&
62.0&
22.0&
36.0&
38.0&
41.0&
36.0&
28.0&
\textbf{28.0}&
\textbf{26.0}&
62.0&
\textbf{58.6}&
\textbf{41.4}&
\textbf{51.2}&
20.9&
21.8&
30.6&
40.5&
16.3&
9.8&
20.0&
27.0&
\textbf{47.5}&
11.1
\\
  Gemini Pro & 42.0&
\textbf{76.0}&
\textbf{88.0}&
\textbf{70.0}&
\textbf{35.0}&
\textbf{45.0}&
\textbf{74.0}&
54.0&
\textbf{49.0}&
\textbf{46.0}&
\textbf{56.0}&
\textbf{62.0}&
36.0&
28.0&
26.0&
16.0&
62.0&
49.3&
35.6&
43.2&
\textbf{30.3}&
20.8&
32.4&
\textbf{41.5}&
\textbf{35.3}&
18.0&
13.0&
31.5&
36.5&
20.8
\\
         \bottomrule
    \end{tabular}}
        \captionof{table}{Model results per question prototype}%
            \label{tab:full_res}
    \end{landscape}
    \clearpage%
    }

\end{document}